\documentclass[10pt]{article} %
\usepackage[accepted]{tmlr}

\usepackage{amsmath,amsfonts,bm}

\def\eqref#1{equation~\ref{#1}}

\def\1{\bm{1}}

\DeclareMathAlphabet{\mathsfit}{\encodingdefault}{\sfdefault}{m}{sl}
\SetMathAlphabet{\mathsfit}{bold}{\encodingdefault}{\sfdefault}{bx}{n}

\usepackage[table]{xcolor}
\usepackage{xcolor}
\usepackage{tcolorbox}
\usepackage{hyperref}
\usepackage{url}
\usepackage{amsmath}
\usepackage{xurl}
\usepackage{booktabs}
\usepackage{array,multirow}   
\usepackage{amssymb}
\usepackage{amsfonts}
\usepackage{caption}
\usepackage{subcaption}
\usepackage{here}
\usepackage{tabularx}
\usepackage{enumitem}
\usepackage{hhline}
\usepackage{bm}
\usepackage{listings}
\usepackage{tikz}
\usepackage{pifont}
\usetikzlibrary{patterns} %
\usepackage{afterpage}
\usepackage{graphicx}
\usepackage{wrapfig}

\title{Adapting Chat Language Models\\ Using Only Target Unlabeled Language Data}

\author{\name Atsuki Yamaguchi \email ayamaguchi1@sheffield.ac.uk \\
      \addr University of Sheffield\\
      \AND
      \name Terufumi Morishita \email \\
      \addr Hitachi, Ltd.
      \AND
      \name Aline Villavicencio \email a.villavicencio@sheffield.ac.uk \\
      \addr University of Exeter\\
      University of Sheffield\\
      \AND
      \name Nikolaos Aletras \email n.aletras@sheffield.ac.uk \\
      \addr University of Sheffield\\
}

\def\month{09}  %
\def\year{2025} %
\def\openreview{\url{https://openreview.net/forum?id=6IdoIKowfe}} %

\begin{document}

\maketitle

\begin{abstract}
Vocabulary expansion (VE) is the de-facto approach to language adaptation of large language models (LLMs) by adding new tokens and continuing pre-training on target data. While this is effective for base models trained on unlabeled data, it poses challenges for \textit{chat} models trained to follow instructions through labeled conversation data. Directly adapting the latter with VE on target unlabeled data may result in forgetting chat abilities.
While ideal, target chat data is often unavailable or costly to create for low-resource languages, and machine-translated alternatives are not always effective.
To address this issue, previous work proposed using a base and chat model from the same family. This method first adapts the base LLM with VE on target unlabeled data and then converts it to a chat model by adding a chat vector (CV) derived from the weight difference between the source base and chat models.
We propose ElChat, a new language adaptation method for chat LLMs that adapts a chat model directly on target unlabeled data, \textit{without a base model}. It elicits chat abilities by injecting information from the source chat model.
ElChat offers more robust and competitive target language and safety performance while achieving superior English, chat, and instruction-following abilities compared to CV.\footnote{Our code is available on \href{https://github.com/gucci-j/chat-cve}{GitHub}. The adapted models are available on \href{https://huggingface.co/collections/atsuki-yamaguchi/elchat-680bb4831dd248e734015d27}{Hugging Face Hub}.}
\end{abstract}

\section{Introduction}
Vocabulary expansion (VE) is the de-facto approach to adapting large language models (LLMs) in a target language~\citep{cui2024efficienteffectivetextencoding,fujii2024continual,choi-etal-2024-optimizing}. It typically consists of two main steps: (i) new tokens are added to the model vocabulary by expanding the input embedding and output head matrices; and (ii) continual pre-training on target data to learn the input and output embeddings of the new tokens~\cite[\textit{inter alia}]{cui2024efficienteffectivetextencoding,fujii2024continual,choi-etal-2024-optimizing,tejaswi-etal-2024-exploring,mundra-etal-2024-empirical}. VE is important because LLMs including chat models often perform poorly in languages underrepresented in the training data~\citep{geng2025why,huang-etal-2024-chat}. Moreover, target language tokenization suffers from overfragmentation due to the heavy reliance on data and vocabulary from particular languages (e.g., English), resulting to more inference steps especially in low-resource languages~\citep{ahia-etal-2023-languages,petrov2023language,ali-etal-2024-tokenizer}.
Consequently, VE is a necessary step to mitigate this issue and achieve crucial inference speedups.
While the vocabularies of frontier LLMs are often large, e.g., 152K for Qwen2.5~\citep{yang2024qwen2technicalreport} and 128K for Llama 3.1~\citep{dubey2024llama3herdmodels}, they still suffer from this overfragmentation in underrepresented languages. This means such languages often require substantially more inference steps than their high-resource counterparts.
For example, processing a text sequence in Amharic requires 3.48x more inference steps using the default Qwen2.5 without VE.

While VE is effective for \textit{base} models trained on unlabeled data, its application poses significant challenges when the LLM at hand is a \textit{chat} model trained to follow instructions through labeled conversation data.
Ideally, we need access to target chat data to effectively adapt chat models. However, this is often unavailable or costly to create for low-resource languages, including the acquisition of human feedback~\citep{huang-etal-2024-chat}. 
Alternatively, machine-translated chat data are not consistently effective~\citep{tao-etal-2024-unlocking}. 

To address this issue, \citet{huang-etal-2024-chat} proposed chat vector (CV), a method to obtain a chat model in the target language with access to target unlabeled data only. CV first adapts the base LLM with VE on target unlabeled data and then converts it to a chat model by adding a chat vector derived from the weight difference between the source base and chat models. However, this requires access to base and chat models from the same family that might not always be available, hindering its applicability.
For example, the Phi-3~\citep{abdin2024phi3technicalreporthighly} and Phi-4 ~\citep{abdin2024phi4technicalreport} do not provide base models due to safety reasons.\footnote{\url{https://huggingface.co/microsoft/phi-4/discussions/4}}
Similarly, Velvet\footnote{\url{https://huggingface.co/Almawave/Velvet-14B}}, EXAONE-3.5\footnote{\url{https://huggingface.co/collections/LGAI-EXAONE/exaone-35-674d0e1bb3dcd2ab6f39dbb4}}, and Trillion~\citep{han2025trillion7btechnicalreport} models are available only as a chat model.
Crucially, it is completely to the discretion of developers to decide whether they publish both base and chat variants.

\begin{wrapfigure}{r}{0.48\textwidth}
\vspace{-5pt}
\centering
\includegraphics[width=\linewidth]{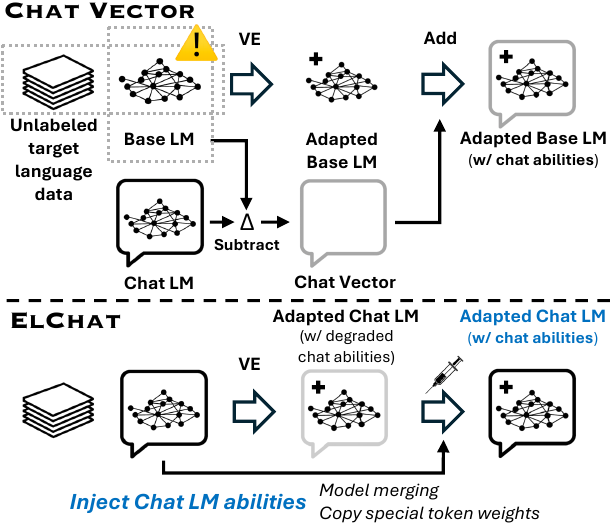}
\caption{
Chat LLM language adaptation with Chat Vector~\citep{huang-etal-2024-chat} and ElChat (ours).
Note that a \textit{base} LM in this paper refers to an LM pre-trained on unlabeled data without any further post-training.
A \textit{chat} LM, on the other hand, is a base model that has been further supervised fine-tuned on labeled conversational data, enabling it to follow instructions.
}
\label{fig:motivation}
\end{wrapfigure}

In this paper, we propose ElChat, a new language adaptation method for chat LLMs that adapts a chat model directly on target unlabeled data, eliminating the need for a base model (Figure \ref{fig:motivation}).
We hypothesize that direct adaptation of a source chat model with VE on target unlabeled data negatively impacts its chat and instruction-following abilities by altering its parametric knowledge. However, we posit that these can still be recovered. For this purpose, ElChat leverages information from the source chat model to elicit chat abilities through two key mechanisms.
First, we employ model merging to integrate distinct parametric knowledge from the source and target models~\citep{pmlr-v162-wortsman22a,yadav2023tiesmerging,yu2024language,goddard-etal-2024-arcees}.
We hypothesize that model merging helps restore the chat and instruction-following abilities of the source model while preserving the target language performance achieved by the target model.
Second, we reuse the weights of special tokens from the source model.
For example, tokens that mark the start of a conversation turn should be crucial for activating the instruction-following ability as they are used to structure raw input into chat format. However, direct adaptation on target unlabeled data may degrade their functionality as they are modified during VE. To mitigate this, we copy these token weights directly from the source model to the target model.

We investigate the efficacy of ElChat by experimenting with two popular chat models across seven typologically diverse languages.
Our evaluation includes safety, chat, and instruction-following performance.
Additionally, we also assess target and source language task performance and target language inference speed.
Our key contributions are as follows:

\begin{itemize}
    \item We propose ElChat that adapts a chat model directly on target unlabeled data, eliminating the need for (i) a base model and (ii) target chat data.

    \item ElChat achieves better chat and instruction-following abilities and source language performance than CV. It is also competitive and more robust (i.e., consistently outperforming the source chat model) in the target language and safety tasks compared to CV (\S\ref{subsec:task_target},  \S\ref{subsec:task_english}, \S\ref{subsec:task_chat}).

    \item Despite model modifications, ElChat achieves comparable target inference speedups across models and tasks, matching the performance of the adapted VE and CV models (\S\ref{sec:efficiency-result}).

\end{itemize}

\section{Related Work} \label{sec:related}

\subsection{Cross-lingual Vocabulary Adaptation}

The most popular approach to adapting LLMs to a target language is by expanding their vocabulary (VE) with tokens from the target language~\citep{balachandran2023tamilllamanewtamillanguage,larcher2023cabritaclosinggapforeign,pipatanakul2023typhoonthailargelanguage,lin2024mala500massivelanguageadaptation,cui2024efficienteffectivetextencoding,kim2024efficienteffectivevocabularyexpansion,fujii2024continual,choi-etal-2024-optimizing,nguyen-etal-2024-seallms,tejaswi-etal-2024-exploring,mundra-etal-2024-empirical}.

Other methods to language adaptation include full or partial vocabulary replacement with a new target vocabulary~\citep{ostendorff2023efficientlanguagemodeltraining,csaki2023efficientlyadaptingpretrainedlanguage,da-dalt-etal-2024-flor,remy2024transtokenization,yamaguchi-etal-2024-empirical,dobler2024language,cahyawijaya-etal-2024-cendol},  hypernetwork for tokenizer transfer~\citep{minixhofer2024zeroshot}, and adapters for vocabulary alignment~\citep{han2024adaptersalteringllmvocabularies}.
Our work focuses on VE as it has been widely used recently for mostly base LLM adaptation in languages such as Chinese, Japanese, Korean, and Persian~\citep[\textit{inter alia.}]{cui2024efficienteffectivetextencoding,fujii2024continual,choi-etal-2024-optimizing,mahdizadeh-sani-etal-2025-extending}.

\subsection{Language Adaptation of Chat Models}

Recent work has proposed developing chat models in a target language from source base models. For example, \citet{toraman2024llamaturkadaptingopensourcegenerative} and \citet{zhao2024llamaenglishempiricalstudy} apply VE to \textit{base} models using target language chat data, consisting of 52k samples and 500M tokens, respectively.
\citet{bandarkar2024layerswappingzeroshotcrosslingual} also adapt base models using 30-40k target language chat data. Their approach also adapts a task-specific (i.e., math) model on 200k English math samples, followed by merging the two models to enhance math performance in the target language.
\citet{alexandrov-etal-2024-mitigating} iteratively merge models trained on subsets of available target language data to effectively mitigate catastrophic forgetting.
Their method first adapts base models with continual pre-training (CPT) on target unlabeled language data, followed by instruction tuning on target language chat data samples.
However, this approach requires substantial target language data (at least 50B tokens of unlabeled data for CPT, and 78K samples of target language chat data) to ensure that each subset contains sufficient information for effective adaptation.
A different approach, proposed by \citet{tao-etal-2024-unlocking}, involves merging two base models: (1) one supervised fine-tuned on 162k English data samples, and (2) another trained on at least eight billion tokens of target unlabeled language data. However, it still relies on the availability of a base model.
\citet{geng2025why} propose adapting source chat models directly through a multi-stage training approach.
This method involves target unlabeled data and transfer fine-tuning (i.e., supervised fine-tuning tasks using translated target chat data.)

The main limitation with this line of work is that it requires access to target chat data (real or translated), typically in large volumes.
Chat data is often unavailable or costly to produce for low-resource languages, while machine-translated chat data is not always effective for adaptation~\citep{tao-etal-2024-unlocking}. For example, Burmese, one of our experimental languages, consists only of 472 manually annotated instruction samples in the Aya Dataset~\citep{singh-etal-2024-aya}. This is insufficient for direct application of VE, as its typical data requirements are in the order of millions of tokens~\citep{tejaswi-etal-2024-exploring}, making these methods not applicable in such settings.

\citet{huang-etal-2024-chat} assumes a language adaptation setting where there is no access to target chat data (real or translated). This is a more realistic scenario for low-resource languages. For example, Burmese has only 172k target unlabeled language data in MADLAD-400~\citep{kudugunta2023madlad} available for adaptation.
Their proposed CV method obtains a target chat model using a source base and a source chat model. This approach to chat LLM adaptation is the closest to our work.

\section{ElChat: Eliciting Chat and Instruction-following Abilities} \label{sec:elchat}

\begin{figure*}[t]
\centering
\includegraphics[width=0.95\textwidth]{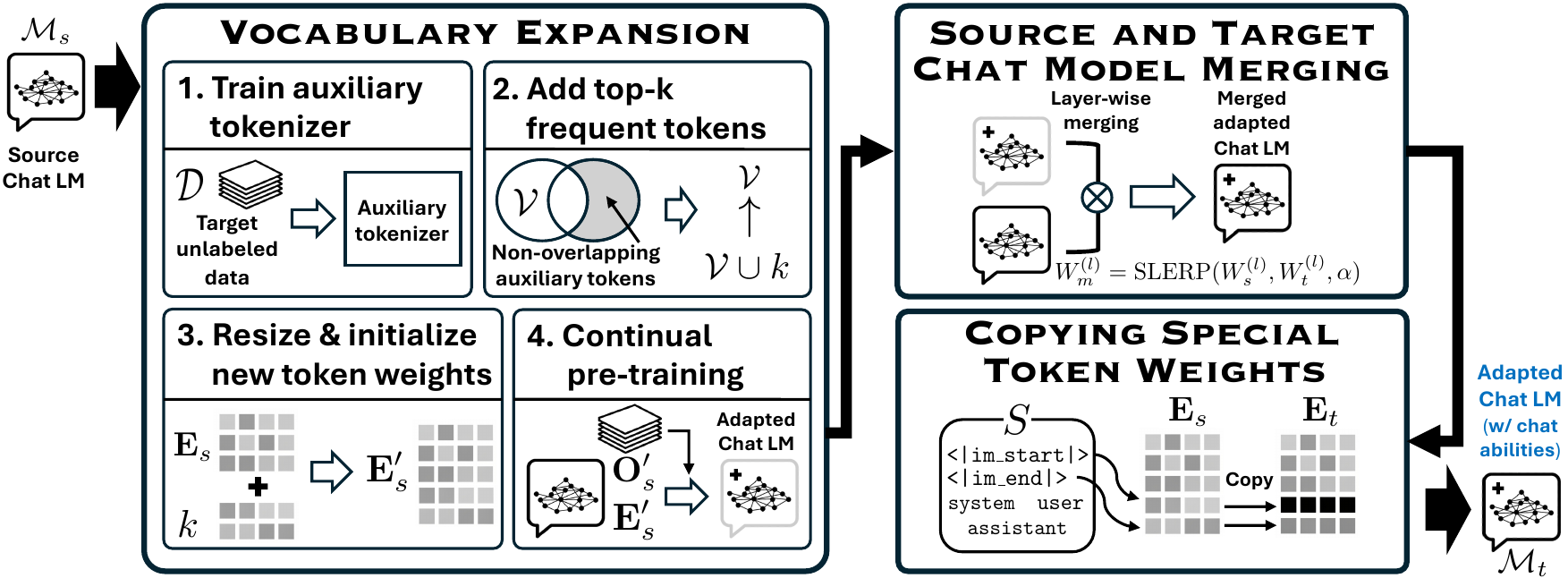}
\caption{
ElChat: A three-step adaptation process overview.
}
\label{fig:methodology}
\end{figure*}

Similar to \citet{huang-etal-2024-chat}, we aim to adapt a chat LLM to a target language, assuming that we only have access to target unlabeled data and no target chat data (real or translated) at all.
Unlike \citet{huang-etal-2024-chat}, our goal is to remove the dependence on a base model that might not always be available and adapt a chat model directly on target unlabeled data with VE.

To achieve this, we introduce ElChat that consists of three steps: (i) VE on the source chat model using target unlabeled  data to obtain an initial adapted target chat model; (ii) merging the source and target models; (iii) injecting information from the source to the target by copying tokens that are specific to chat and instruction-following capabilities. Figure \ref{fig:methodology} visualizes these steps.
While the individual components of these are established techniques, the key novelty of ElChat resides in its strategic combination to overcome the dependence on the base model.

We expect that modifying the parametric knowledge of the source model by training on target unlabeled data in step (i) will improve its target language skills by updating specific areas of the network but negatively impact its chat and instruction-following abilities due to catastrophic forgetting.
To remedy this, we hypothesize that we can elicit the latter through (ii) model merging~\citep{pmlr-v162-wortsman22a,yadav2023tiesmerging,yu2024language,goddard-etal-2024-arcees}. It allows the integration of distinct parametric knowledge from the source and target models without incurring any additional training costs, i.e., chat abilities from the source and target language knowledge from the target. Finally, in step (3), we restore the representation of special tokens used in the chat template.
We assume this will guide the target model in effectively responding to user instructions by transferring this information from the source.

\paragraph{Vocabulary Expansion (VE).}
Given the sole availability of target unlabeled data, we apply VE on the source chat LLM by expanding its input and output head matrices with new tokens, followed by CPT on the target data. This process follows the standard protocol adopted in \cite{tejaswi-etal-2024-exploring} and \cite{mundra-etal-2024-empirical} for resizing and initializing these matrices.

More specifically, given target unlabeled data $\mathcal{D}$ and a source chat model $\mathcal{M}_s$ with initial vocabulary $\mathcal{V}$:
\begin{enumerate}
    \item An auxiliary target language tokenizer is first trained on $\mathcal{D}$. This tokenizer utilizes the same underlying tokenization function (e.g., byte-level BPE in this paper) as $\mathcal{M}_s$.

    \item From this auxiliary tokenizer, we then select the top-$k$ most frequent tokens (e.g., $k = 10$K by default, as in \cite{tejaswi-etal-2024-exploring}) that do not overlap with $\mathcal{V}$ but are present in the auxiliary tokenizer's vocabulary. These selected $k$ tokens are then added to $\mathcal{V}$.

    \item To accommodate these $k$ new tokens in $\mathcal{M}_s$, its input $\mathbf{E}_s\in\mathbb{R}^{|\mathcal{V}| \times H}$ and output head $\mathbf{O}_s\in\mathbb{R}^{H \times |\mathcal{V}|}$ matrices are resized.
    They become $\mathbf{E}_s'\in\mathbb{R}^{(|\mathcal{V}|+k) \times H}$ and $\mathbf{O}_s'\in\mathbb{R}^{H \times (|\mathcal{V}|+k)}$ respectively, where $H$ denotes the hidden dimensionality of $\mathcal{M}_s$. The weights corresponding to the newly added tokens are then initialized using mean initialization~\citep{yao-etal-2021-adapt}, a popular and simple, yet effective method in VE~\citep{fujii2024continual,tejaswi-etal-2024-exploring,mundra-etal-2024-empirical}. The weight of each newly added token is initialized as the average embedding or language modeling head weight of their corresponding source tokens, obtained using the corresponding source tokenizer.
    
\end{enumerate}

After the above procedure, $\mathcal{M}_s$ undergoes continual pre-training on $\mathcal{D}$ using a causal language modeling objective. A key difference between chat and base models is that the former uses a chat template. This includes specific roles (e.g., user, system, or assistant) and has a placeholder for message text in the prompt (see Appendix \ref{appendix:setup} for details). For example, a chat template might structure input as follows:
\begin{verbatim}
<|im_start|>system
You are a helpful assistant.<|im_end|>
<|im_start|>user
What is the capital of the US?<|im_end|>
<|im_start|>assistant
\end{verbatim}

During CPT, we remove the default chat template of the model to support unlabeled data because the unlabeled data typically lacks these explicit role annotations.
During inference, we append it to task-specific prompt templates (see Table \ref{tab:prompt} for task-specific prompts).

\paragraph{Source and Target Chat Model Merging (Merge).}
After VE, we merge the source and target chat models.
We employ a popular merging method: spherical linear interpolation (SLERP)~\citep{goddard-etal-2024-arcees} to merge each layer of the source and adapted models.\footnote{We also test linear merging~\citep{pmlr-v162-wortsman22a} yielding similar results (see Appendix \ref{appendix:results}). This suggests the robustness of our merging approach regardless of the specific interpolation method. Nonetheless, we primarily utilize SLERP to streamline our experiments due to its superior fine-grained control for merging, even though it may not be empirically superior across all tasks~\citep{goddard-etal-2024-arcees}.}
This process excludes the embedding and language modeling head from merging because the source and target models use different vocabularies.

Given the source chat model $\mathcal{M}_s$ and the target-adapted chat model $\mathcal{M}_t$ (which has undergone VE), the weights of each corresponding layer $l$ (e.g., self-attention and feed-forward layers) are merged. For each such layer $l$, let $\mathbf{W}_s^{(l)}$ denote the weight matrix from $\mathcal{M}_s$, and $\mathbf{W}_t^{(l)}$ denote the corresponding weight matrix from $\mathcal{M}_t$. The merged weight matrix, denoted $\mathbf{W}_m^{(l)}$, is then computed using SLERP:
$$\mathbf{W}_m^{(l)} = \text{SLERP}(\mathbf{W}_s^{(l)}, \mathbf{W}_t^{(l)}, \alpha)$$
\noindent where $\alpha \in [0, 1]$ is an interpolation coefficient. This $\alpha$ serves as a hyperparameter controlling the balance between $\mathcal{M}_s$ and $\mathcal{M}_t$ learned representations, where $\alpha=0$ corresponds to using only the weights of $\mathcal{M}_s$ and $\alpha=1$ to using only those of $\mathcal{M}_t$.

\paragraph{Copying Special Token and Language Modeling Head Weights (Copy).}
Special tokens used in a chat template (e.g., \texttt{<im\_start>} in Qwen2.5 to represent the start of a turn) should be critical in supporting chat and instruction-following abilities of a model (see Appendix \ref{appendix:setup} for a full list of special tokens).
Although the embedding and language modeling heads are excluded from merging due to vocabulary differences, leaving them unchanged may not be optimal for eliciting the chat and instruction-following abilities of the adapted chat model.
Hence, we copy all the special token weights from the source model to the adapted model.

Specifically, let $S$ be the set of token IDs corresponding to the special tokens defined by the chat template of $\mathcal{M}_s$.
The input embedding matrix and the language modeling head of $\mathcal{M}_t$ are updated as follows:

For each special token ID $x \in S$, the embedding vector for token $x$ in $\mathcal{M}_t$, $\mathbf{E}_t \in \mathbb{R}^{(|\mathcal{V}|+k) \times H}$, is updated by copying the corresponding vector from $\mathcal{M}_s$'s input embedding matrix ($\mathbf{E}_s$): $\mathbf{E}_t[x, :] \leftarrow \mathbf{E}_s[x, :]$, where $\mathbf{E}[i, :]$ denotes the embedding vector for token ID $i$.

Similarly, the language modeling head weights corresponding to token $x$ in $\mathcal{M}_t$, $\mathbf{O}_t \in \mathbb{R}^{H \times (|\mathcal{V}|+k)}$, are updated by copying the corresponding vector from $\mathcal{M}_s$: $\mathbf{O}_t[:, x] \leftarrow \mathbf{O}_s[:, x]$, where $\mathbf{O}[:, j]$ denotes the column of weights for token ID $j$ in the language modeling head.

\section{Experimental Setup} \label{sec:setup}

This section describes the experimental setup in this paper. More details are listed in Appendix \ref{appendix:setup}.

\subsection{Source Models}\label{subsec:models}
We use two popular chat models as source: Qwen2.5 7B~\citep{yang2024qwen2technicalreport}; and Llama 3.1 8B~\citep{dubey2024llama3herdmodels}, across experiments.
Additionally, to ensure consistency in our analysis, we also incorporate the state-of-the-art chat model Qwen3 14B~\citep{yang2025qwen3technicalreport}. The corresponding results and analysis for Qwen3 are presented in Appendix \ref{appendix:qwen3-result}.

\subsection{Target Languages and Adaptation Data}
We experiment with the following seven typologically diverse languages, assuming that they are likely to be underrepresented compared to English in the pre-training data of the source models, or entirely absent: Amharic (Afroasiatic), Bengali (Indo-European), Burmese (Sino-Tibetan), Gujarati (Indo-European), Sinhala (Indo-European), Tamil (Dravidian), and Telugu (Dravidian).
The ratio of training data in each model for each source base and chat model has not been explicitly disclosed (Appendix \ref{appendix:language}).
Note that we do not consider Latin script target languages as they are less likely to suffer from overfragmentation and usually benefit less from VE~\citep{yamaguchi2024effectivelyexpandvocabularyllms,tejaswi-etal-2024-exploring}.
Due to computational constraints, we use Qwen3 for Amharic, Bengali, and Telugu only.\footnote{Amharic is selected because it shows the most significant speedup gains in target language generative tasks with Qwen2.5 and Llama 3.1 (Tables \ref{tab:speed_performance_qwen25} and \ref{tab:speed_performance_llama31}). Bengali and Telugu are chosen as they are the only languages covered by \textsc{mgsm}.}

For the CPT part of VE, we use MADLAD-400~\citep{kudugunta2023madlad}, which consists of highly-filtered document-level samples sourced from CommonCrawl, and randomly sample 250K language-specific documents for each language as the target unlabeled data.\footnote{For languages with less than 250K documents (i.e., Amharic and Burmese), we use the full articles.}

\subsection{Continual Pre-training} \label{subsec:init}

Following \citet{remy2024transtokenization}, we train the embedding, LM head, and the top and bottom two layers of a source model.
This approach aims to calibrate only the parts closely related to the encoding and decoding of the target language~\citep{wendler-etal-2024-llamas,tang-etal-2024-language,zhao2024how}, minimizing changes to the source model while allowing cost-effective tuning.

\subsection{Baselines} \label{subsec:baselines}
We compare ElChat against the following baselines:

\begin{itemize}
    \item Off-the-shelf base (\textbf{Base}) and chat (\textbf{Chat}) models without target language adaptation.
    \item Base and Chat models adapted using standard VE, denoted by \textbf{Base+VE} and \textbf{Chat+VE} respectively. Note that the latter uses a chat template in inference (see \S\ref{sec:elchat}).
    \item \textbf{CV} proposed by \citet{huang-etal-2024-chat}, augmenting Base+VE with chat vector using Base and Chat.
\end{itemize}

\noindent
For reference, we also experiment with adapting Chat and Base using only CPT on the same target language data without VE (i.e., no inference speedup in a target language).
We provide the results and analysis of these CPT-only models in Appendix \ref{appendix:cpt-result}.

\subsection{Evaluation Tasks}
We evaluate the efficacy of ElChat in safety, chat, and instruction-following performance, and target and source language performance.

\paragraph{Safety, Chat, and Instruction-Following.}
Following \citet{cahyawijaya-etal-2024-cendol}, we conduct safety evaluation on target language translated data including \textsc{TruthfulQA}~\citep{lin-etal-2022-truthfulqa}, \textsc{ToxicGen}~\citep{hartvigsen-etal-2022-toxigen}, and \textsc{ImplicitHate}~\citep{elsherief-etal-2021-latent}.
We also measure chat and instruction-following abilities in the source language (English) using \textsc{IFEval}~\citep{zhou2023instructionfollowingevaluationlargelanguage}, \textsc{GSM8K}~\citep{cobbe2021trainingverifierssolvemath} as multi-turn few-shot, and \textsc{MT-Bench}~\citep{zheng2023judging}.
Furthermore, we measure the performance on English \textsc{AlpacaEval} v2.0~\citep{alpaca_eval,dubois2024length} for additional analysis.

Target language evaluation is challenging for instruction-following and chat tasks due to the limited data availability.
LLM-as-a-Judge~\citep{zheng2023judging} is also unstable according to \citet{azime-etal-2024-walia} in low-resource languages.
Hence, we use multi-turn \textsc{MGSM}~\citep{shi2023language} for target language evaluation as it consists of manually translated high-quality data.

\paragraph{Target Language.}
We use both generative and discriminative target language tasks.
For generative tasks, we use summarization (\textsc{sum}) using XL-SUM~\citep{hasan-etal-2021-xl} and English-to-target machine translation (\textsc{mt}) using FLORES-200~\citep{nllb-22}.
For a discriminative task, we employ multiple-choice text classification (\textsc{mc}) using Belebele~\citep{bandarkar-etal-2024-belebele} and Global MMLU (\textsc{gmmlu})~\citep{singh2024globalmmluunderstandingaddressing} as general target language understanding benchmarks.

\paragraph{Source Language (English).}
We assess the extent to which the adapted models retain their general task-solving abilities in English \textsc{sum}, target-to-English \textsc{mt}, and English \textsc{mc} using the same datasets as those employed for target languages.
We also use \textsc{mmlu}~\citep{hendrycks2021measuring} as an English language understanding benchmark and English \textsc{bbh}~\citep{srivastava2023beyond,suzgun-etal-2023-challenging} as a stress-test benchmark.

Following \citet{ahia-etal-2023-languages}, we use 500 random samples for generative tasks: \textsc{sum} and \textsc{mt}. The rest use the full test sets for evaluation.

\subsection{Evaluation Metrics} \label{subsec:metric}
\paragraph{Task Performance.}
We report the standard metrics for each task: accuracy for \textsc{mc}, \textsc{gmmlu}, \textsc{mmlu}, \textsc{bbh}, \textsc{TruthfulQA}, and \textsc{IFEval} (strict prompt), and exact match for \textsc{GSM8K} and \textsc{MGSM}.
For \textsc{MT-Bench}, we use the mean score over two turns across all questions. Adhering to the standard protocol in LightEval~\citep{lighteval}, each score is determined using \href{https://huggingface.co/flowaicom/Flow-Judge-v0.1}{\texttt{Flow-Judge-v0.1}} and follows a Likert-5 scale.
For \textsc{AlpacaEval}, we use a win-rate over GPT-4 (1106 Preview) measured by GPT-4.1 nano (2025-04-14).\footnote{\url{https://openai.com/index/gpt-4-1/}}
For \textsc{sum} and \textsc{mt}, we primarily use chrF~\citep{popovic-2015-chrf}.\footnote{Although chrF has been a widely used metric for \textsc{sum} and \textsc{mt} \citep[\textit{inter alia}]{ebrahimi-etal-2023-findings,remy2024transtokenization}, we also show ROUGE-L~\citep{lin-2004-rouge} for \textsc{sum} and chrF++~\citep{popovic-2017-chrf} for \textsc{mt} in Appendix \ref{appendix:results}.}
For \textsc{ToxicGen} and \textsc{ImplicitHate}, we use safety score, which is the percentage of likeliness of the model producing benign over harmful sentences, following \citet{cahyawijaya-etal-2024-cendol}.

We report average zero- and three-shot performance across three different runs for \textsc{sum} and \textsc{mt}, respectively.
For the remaining tasks, we report single-run zero-shot performance for \textsc{IFEval}, \textsc{MT-Bench}, \textsc{ToxicGen} and \textsc{ImplicitHate}, three-shot performance for \textsc{mc}, \textsc{TruthfulQA}, five-shot performance for \textsc{gmmlu}, \textsc{mmlu}, \textsc{GSM8K}, and \textsc{MGSM} as these tasks are deterministically evaluated with temperature set to zero.

\paragraph{Inference Efficiency.}
VE offers inference speedups in a target language compared to source models~\citep{tejaswi-etal-2024-exploring,mundra-etal-2024-empirical,yamaguchi2024effectivelyexpandvocabularyllms}.
To quantify this, we measure the number of tokens generated per second (tokens/s) \citep{hong-etal-2024-accelerating}.

\section{Task Performance}

\subsection{Safety, Chat, and Instruction-following} \label{subsec:task_chat}

\begin{wrapfigure}{r}{0.5\textwidth}
    \centering
    \vspace{-13pt}
    \begin{subfigure}[b]{\linewidth}
        \centering
        \includegraphics[width=\textwidth]{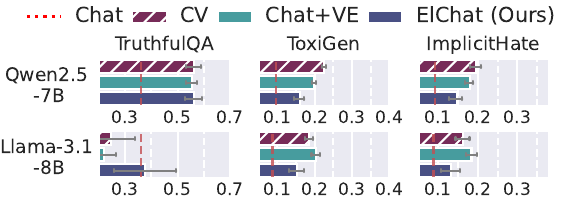}
        \caption{Target Language}
    \end{subfigure}
    \begin{subfigure}[b]{\linewidth}
        \vspace{5pt}
        \centering
        \includegraphics[width=\textwidth]{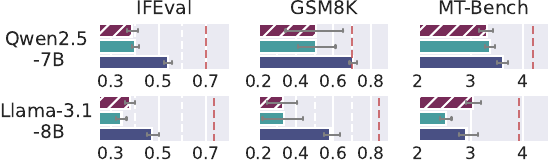}
        \caption{English (Source)}
    \end{subfigure}
\caption{Aggregated mean performance ($\uparrow$) across seven target languages for each model on safety, chat, and instruction-following tasks.
Full results are in the Appendix (Tables \ref{tab:chat_performance_qwen25} and \ref{tab:chat_performance_llama31}).}
\label{fig:chat}
\end{wrapfigure}

\paragraph{Safety.}

Figure \ref{fig:chat} (a) shows the aggregated mean performance in safety tasks across the seven target languages.
We first observe that ElChat outperforms CV in \textsc{TruthfulQA} for both Qwen2.5 and Llama 3.1.
In particular, it substantially helps Llama 3.1, achieving 13-point gains over CV on average.
We speculate that CV may be less effective for classification tasks as reflected in its performance on other discriminative target language tasks (\S\ref{subsec:task_target}).

In contrast, CV often surpasses ElChat in \textsc{ToxiGen} and \textsc{ImplicitHate} for both models, with differences of up to 6 points, e.g., in \textsc{ToxiGen} with Qwen2.5.
This is primarily due to the use of Merge instead of Copy in ElChat (see Table \ref{tab:chat_performance_llama31}).  Specifically, ElChat without Copy follows similar trends to ElChat, while ElChat without Merge exhibits similar trends to Chat+VE and CV. 
This is intuitive, as merging a model with the lowest target safety performance (i.e., Chat) with Chat+VE can degrade safety performance.

We finally find that ElChat, CV, and Chat+VE outperform the Chat baseline across tasks and models, with gains ranging from 1.4 (\textsc{TruthfulQA} with Llama 3.1 using ElChat) to 20 points (\textsc{TruthfulQA} with Qwen2.5 using ElChat).
The only exceptions are CV and Chat+VE in \textsc{TruthfulQA}, where they underperform Chat by 11 and 14 points, respectively. 
This instability highlights the advantage of ElChat, as it consistently enhances safety performance over Chat across tasks.

\paragraph{Chat and Instruction-following.}
Figure \ref{fig:chat} (b) shows the aggregated mean performance across chat and instruction-following tasks in English (source).
We first analyze the extent to which ElChat impacts performance on chat and instruction-following tasks compared to the Chat baseline.
Note that some performance degradation is inevitable, as adapting Chat to \textit{target} unlabeled data inherently affects the \textit{source} chat and instruction-following abilities (\S\ref{sec:elchat}). 

As anticipated, we find that ElChat exhibits performance degradation across tasks and models compared to the Chat baseline.
However, the extent of this degradation varies depending on the model, task, and adaptation approach.
For example, ElChat achieves comparable performance on \textsc{GSM8K} but experiences reductions of 16 and 0.57 points on \textsc{IFEval} and \textsc{MT-Bench}, respectively.
Despite these drops, ElChat successfully improves instruction-following performance compared to the respective adapted model, Chat+VE.
It demonstrates improvements of 14 and 13 points over Chat+VE for Qwen2.5 and Llama 3.1, respectively.
These results indicate that ElChat can inject instruction-following capabilities into the adapted model.

\begin{wraptable}{r}{0.3\textwidth}
\centering
\vspace{-2em}
\small
\caption{\textsc{MGSM} performance in Bengali (bn) and Telugu (te) by model.}
\begin{subtable}{\linewidth} %
    \centering
    \caption{Qwen2.5 7B}
    \begin{tabular}{lcc}
        \toprule
        \textbf{Model} & \multicolumn{2}{c}{\textbf{\textsc{EM}}} \\
        & bn & te\\
        \midrule
        \rowcolor{gray!25}
        Chat & .23 & .06\\
        CV & \textbf{.60} & .27\\
        Chat+VE & .39 & .27\\ \midrule
        ElChat (Ours) & .46 & \textbf{.35}\\
        \bottomrule
    \end{tabular}
\end{subtable}

\vspace{1em} %

\begin{subtable}{\linewidth} %
    \centering
    \caption{Llama 3.1 8B}
    \begin{tabular}{lcc}
        \toprule
        \textbf{Model} & \multicolumn{2}{c}{\textbf{\textsc{EM}}} \\
        & bn & te\\
        \midrule
        \rowcolor{gray!25}
        Chat & .30 & .12\\
        CV & .31 & .24\\
        Chat+VE & .26 & .28\\ \midrule
        ElChat (Ours) & \textbf{.51} & \textbf{.41}\\
        \bottomrule
    \end{tabular}
\end{subtable}
\label{tab:mgsm}
\vspace{-5em}
\end{wraptable}

We next observe that ElChat generally outperforms CV across tasks and models in five out of six cases, with performance differences ranging from 9.5 points (\textsc{IFEval} with Llama 3.1) to 25 points (\textsc{GSM8K} with Llama 3.1).
A similar trend is observed on \textsc{AlpacaEval} (Table \ref{tab:alpacaeval} in the Appendix), where ElChat substantially outperforms CV with up to a 16-point gain in Qwen2.5, while still performing competitively with only a 0.95-point drop in Llama 3.1.
These results suggest that ElChat is more effective than CV in enhancing both chat and instruction-following abilities.

Finally, ElChat’s performance advantage extends to target language tasks. Table \ref{tab:mgsm} shows model performance in \textsc{MGSM} (covering Bengali and Telugu). Overall, ElChat surpasses both CV and the original chat model, Chat, in three and four out of four cases, respectively.
Notably, ElChat significantly improves Telugu performance by 29 points with both Qwen2.5 and Llama 3.1. Additionally, while CV enhances Bengali performance by just 1 point over Chat in Llama 3.1, ElChat achieves a substantial 21-point gain. 
These results further support the superiority of ElChat over CV.

\subsection{Target Language} \label{subsec:task_target}

\begin{figure*}[t]
\centering
\includegraphics[width=\textwidth]{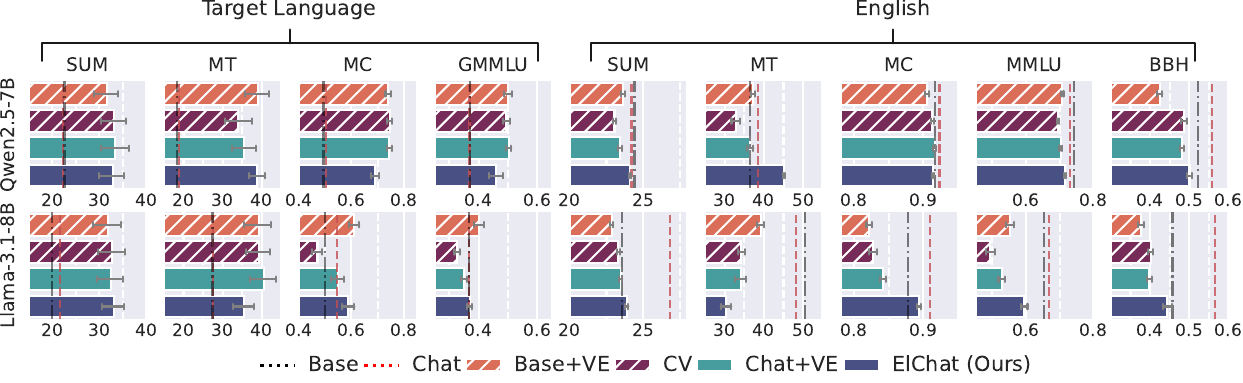}
\caption{Aggregated mean performance across seven target languages for each model (error bars indicate 95\% confidence interval).
Full results are in Tables \ref{tab:task_performance_qwen25} and \ref{tab:task_performance_llama31}.
}
\label{fig:performance}
\end{figure*}

Figure \ref{fig:performance} (left) shows the aggregated mean performance across seven languages for all source and adapted target models in target language.

We note that ElChat consistently outperforms its source chat model (Chat) across all models and tasks.
This improvement is particularly notable in generative tasks (i.e., \textsc{sum} and \textsc{mt}), with gains ranging from 8 points (Llama 3.1 on \textsc{mt}) to 20 points (Qwen2.5 on \textsc{mt}).
While ElChat generally maintains competitive performance (within 3 points) compared to Chat+VE, it exhibits slightly reduced performance (up to 5.9 points with Qwen2.5 on \textsc{mc}) in discriminative tasks (i.e., \textsc{mc} and \textsc{gmmlu}) and \textsc{mt} with Llama 3.1.
These results suggest that ElChat can overall preserve the target language performance, while the model modifications do not substantially degrade performance.

We further observe that ElChat demonstrates competitive performance with CV, with each method outperforming the other in half of the evaluated cases.
Specifically, CV generally outperforms ElChat in Qwen2.5, except for the \textsc{mt} task, whereas ElChat typically achieves better performance than CV in Llama 3.1, excluding \textsc{mt}.
However, CV notably underperforms both the source base and chat models in the two discriminative tasks (i.e., \textsc{mc} and \textsc{gmmlu}) with Llama 3.1.
Thus, although ElChat and CV achieve similar overall performance, our method is more likely to yield improvements in target language tasks.

\subsection{Source Language (English)} \label{subsec:task_english}

Figure \ref{fig:performance} (right) shows the aggregated mean performance across seven languages for all source and adapted target models in English tasks.
ElChat generally underperforms the source Chat baseline.
However, the degree of performance degradation varies considerably depending on the task and model similar to \S\ref{subsec:task_chat}.
For instance, with Llama 3.1, ElChat exhibits substantial performance drops of 7, 12, and 18 points on \textsc{mmlu}, \textsc{bbh} and \textsc{mt}, respectively. 
In contrast, the performance degradation observed with Qwen2.5 is less pronounced, with a maximum decrease of 5.7 points on \textsc{bbh}.
Interestingly, ElChat even demonstrates a 6.7-point improvement on \textsc{mt} with Qwen2.5.
This improvement likely stems from two key factors: (1) ElChat's effective utilization of source tokens (95\%) during generation, compared to other approaches (i.e., Base+VE, CV, and  Chat+VE) that achieve at most 71\% source token utilization; and (2) its successful early stopping, generating a similar number of tokens (33) as Chat (see Figures \ref{fig:ratio} and \ref{fig:tokens} in the Appendix).

Comparing ElChat and Chat+VE, we find that ElChat generally yields better or comparable performance across models and tasks, with the exception of \textsc{mt} with Llama 3.1.
This improvement suggests that ElChat can generally alleviate catastrophic forgetting not only in chat and instruction-following tasks (\S\ref{subsec:task_chat}) but also in source language tasks by injecting the source chat information into Chat+VE.

Conversely, CV exhibits poor performance on English tasks, typically falling short of ElChat in eight out of ten cases.  Moreover, the benefits of catastrophic forgetting mitigation are not consistently observed with CV, as it only outperforms its adapted base model, Base+VE, in five out of ten cases.
This result somewhat contrasts with \citet{huang-etal-2024-chat}, who highlight the benefits of using CV to mitigate catastrophic forgetting and improve knowledge retention and language ability.
We speculate that modifying models through simple arithmetic operations, as in CV, may be less robust than our method.
These results suggest that ElChat more effectively integrates target language abilities while mitigating performance degradation across chat, instruction-following, and source English tasks compared to CV.

\begin{figure}[t]
\centering
\includegraphics[width=\linewidth]{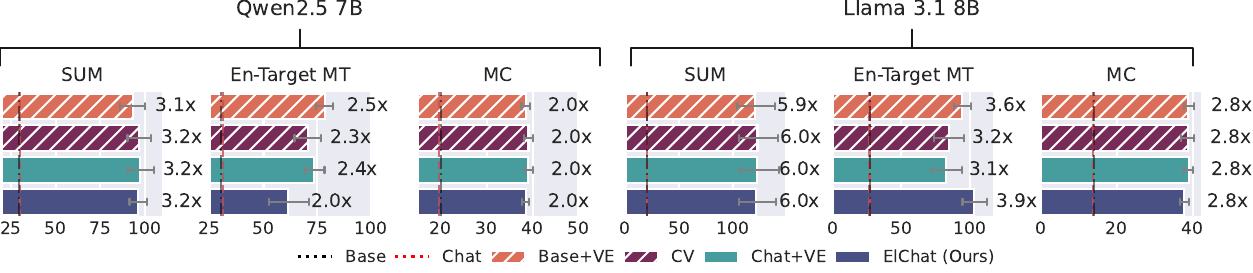}
\caption{Aggregated mean inference speedup (tokens/s) across seven target languages.
The value next to each bar represents the speedup ratio over Chat.
Full results are in the Appendix (Tables \ref{tab:speed_performance_qwen25} and \ref{tab:speed_performance_llama31}).
}
\label{fig:speed}
\end{figure}

\section{Inference Efficiency} \label{sec:efficiency-result}

VE offers inference speedups on target language tasks by reducing text overfragmentation (\S\ref{subsec:metric}).
Since ElChat and CV involve modifications to model weights, they may increase the generation of source tokens, potentially slowing inference compared to Chat+VE and Base+VE.
Furthermore, the inference efficiency of CV with VE has not been previously studied.
Therefore, we investigate the inference efficiency of both methods.
Figure \ref{fig:speed} shows the aggregated mean inference speedups across seven languages and three target tasks.

Overall, both ElChat and CV achieve comparable speedups to their respective adapted models, Chat+VE and Base+VE, demonstrating 2.0x to 6.0x speedups over the source chat model, Chat.
The largest drop is observed between ElChat and Chat+VE on the \textsc{mt} task with Qwen2.5 (0.4x). 
This likely happens because ElChat frequently terminates generation early (see Table \ref{fig:tokens}), resulting in fewer generated tokens (and thus a smaller numerator in the tokens/second calculation).
In summary, these results suggest that the model modifications introduced by ElChat and CV do not negatively impact the inference speedups provided by VE.

\section{Analysis} \label{sec:analysis}
To better understand the behavior of ElChat, we perform an ablation in chat and instruction-following tasks using Llama 3.1. We also conduct a qualitative analysis using \textsc{mgsm} to gain further insights.

\paragraph{Ablation.}

\begin{wraptable}{r}{0.6\textwidth}
    \centering
    
    \caption{Mean performance across languages for chat and instruction-following tasks using Llama 3.1.
    Ablation results in other tasks are available in Appendix \ref{appendix:results}.
    }
    \begin{tabular}{lcccc}
        \toprule
        \textbf{Model} & \textbf{\textsc{IFEval}} & \textbf{\textsc{GSM8K}} & \textbf{\textsc{MGSM}} & \textbf{\textsc{MT-Bench}}\\
        \midrule
        CV & .38\textsubscript{{\tiny.03}} & .33\textsubscript{{\tiny.11}} & .27\textsubscript{{\tiny.03}}  & 3.04\textsubscript{{\tiny.20}} \\
        \midrule
        ElChat & .47\textsubscript{{\tiny.04}} & .58\textsubscript{{\tiny.06}} & .46\textsubscript{{\tiny.05}}  & 2.92\textsubscript{{\tiny.27}} \\
        ~\textendash ~Merge & .34\textsubscript{{\tiny.05}} & .40\textsubscript{{\tiny.18}} & .37\textsubscript{{\tiny.06}}  & 2.53\textsubscript{{\tiny.22}} \\
        ~\textendash ~Copy & .39\textsubscript{{\tiny.02}} & .37\textsubscript{{\tiny.17}} & .32\textsubscript{{\tiny.04}}  & 2.65\textsubscript{{\tiny.23}} \\
        \bottomrule
    \end{tabular}
    \label{tab:ablation}
\end{wraptable}

Table \ref{tab:ablation} shows the importance of each ElChat component.
Removing either of the two components (\S\ref{sec:elchat}) significantly reduces performance across chat and instruction-following tasks, demonstrating their complementarity in eliciting ElChat's abilities. 
For instance, ablating the weight copying (Copy) component, leaving only VE and model merging, causes performance to drop by an average of 8 points on IFEval, 21 points on GSM8K, 14 points on MGSM, and 0.27 points on MT-Bench relative to ElChat.
This substantially undermines the advantage of ElChat over CV as it greatly lags behind CV in MT-Bench and shows marginally better performance than CV with only a few points (up to 4 points on GSM8K) differences in other tasks, unlike at least 9 points difference observed between ElChat and CV.
These results clearly demonstrate that while the individual components are established techniques, their synergistic effect within a chat model context is non-trivial, and their combination is essential for ElChat's strong performance.

\begin{figure*}[t]
    \centering
    \begin{subfigure}[b]{0.45\linewidth}
        \centering
        \includegraphics[trim=10 5 10 1, clip, width=\linewidth]{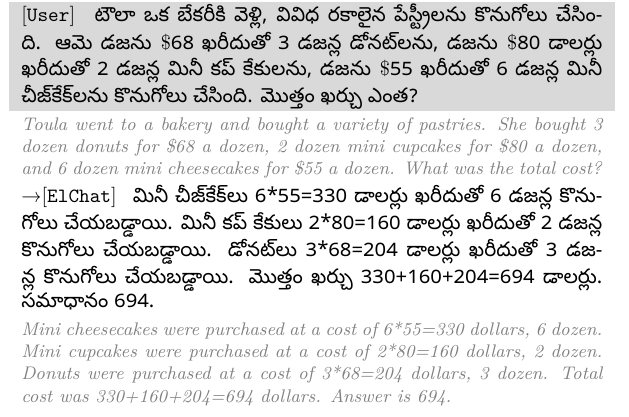}
        \caption{Correct answer - Correct reasoning}
    \end{subfigure}
    \hspace{1em}
    \begin{subfigure}[b]{0.45\linewidth}
        \centering
        \includegraphics[trim=10 5 10 1, clip, width=\linewidth]{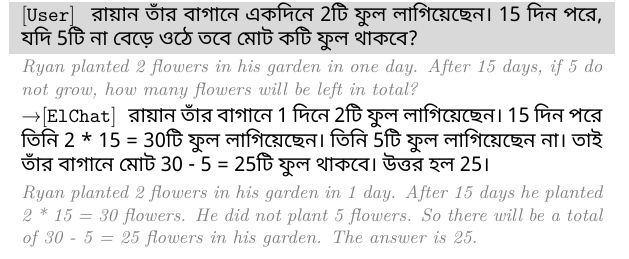}
        \caption{Correct answer - Wrong reasoning}
    \end{subfigure}

\caption{Output examples on \textsc{MGSM}. We use Google Translate for \textcolor{gray}{translation}. Few-shot demonstrations (conversation history) are omitted due to space constraints.}
\label{fig:case}
\end{figure*}

\paragraph{Qualitative Analysis.}

Figure \ref{fig:case} presents examples of ElChat's output on \textsc{MGSM}, highlighting both successes and challenges.
Case (a) showcases correct reasoning and answer generation in Telugu.
However, case (b) demonstrates that even when providing a correct answer in Bengali, ElChat can exhibit wrong reasoning. 
The misinterpretation of ``five flowers do not grow'' as ``Ryan did not plant five flowers'' suggests a potential limitation in understanding nuanced language.
Notably, in the same case, it correctly reasons in English that ``5 did not grow''.
Focusing on improving target language abilities further during VE while mitigating catastrophic forgetting of chat and instruction-following could address this issue.
For instance, making iterative model merging~\citep{alexandrov-etal-2024-mitigating} applicable to low-resource settings is a potential avenue for future investigation.

\section{Conclusion}
We introduced ElChat, a method for directly adapting \textit{chat} models with VE using unlabeled data, eliminating the need for a \textit{base} model and target chat data.
To mitigate potential catastrophic forgetting in the adapted chat models after VE, ElChat elicits chat abilities by injecting information from the source chat model without requiring further training.
Extensive experiments across safety, chat, and instruction-following, target language, and source language tasks demonstrated that ElChat outperforms the previous state-of-the-art CV approach in chat and instruction-following, and English tasks while being competitive and more robust in the target language and safety tasks.
These results highlight ElChat's superior abilities.

\section*{Limitations}

\paragraph{Continual Pre-training Methods.}
This paper uses a continual pre-training method proposed by \citet{remy2024transtokenization}, which tunes the top and bottom two layers of a model and its embedding and language modeling head, for efficient and effective target language adaptation.
Nonetheless, other continual pre-training methods exist, including adapter-based training (e.g., LoRA~\citep{hu2022lora}) and full fine-tuning.
It would be interesting to extensively investigate the effect of different training methods for future work, but this is beyond the scope of this paper.

\paragraph{Model Merging Methods.}
We experiment with linear and SLERP merging as representative model merging methods for simplicity.
More recent methods like TIES~\citep{yadav2023tiesmerging} and DARE-TIES~\citep{yu2024language} might perform even better in ElChat.
Given the resource constraints, we leave this investigation for future work.

\paragraph{Languages.}
This paper covers seven typologically diverse languages, following previous work on language adaptation that has also tested a similar number of languages.
For instance, \citet{minixhofer-etal-2022-wechsel} tested eight languages.
Note that \citet{huang-etal-2024-chat} used three languages (i.e., Chinese, Traditional Chinese, and Korean) to verify the effectiveness of CV.
Experimenting with more languages is an interesting avenue for future work but is out of the scope of this paper, given our limited computing capacity.

\paragraph{Chat and Instruction-following Evaluation.}
Our chat and instruction-following evaluation is mainly on English data except for \textsc{MGSM} due to the limited availability of manually curated language-specific evaluation resources.
\citet{azime-etal-2024-walia} has also noted the instability of using LLM-as-a-Judge in Amharic, which is also one of our experimental languages.
It would be an interesting avenue to explore more chat and instruction-following evaluation in target languages for future work. We hope our work inspires the development of extensive evaluation benchmarks in low-resource languages.

\section*{Ethical Considerations}
Although we conducted extensive experiments across diverse public datasets to validate the effectiveness of ElChat, these datasets do not fully represent all real-world scenarios. Therefore, any model derived from or based on this work should be used with caution.

While this work does not appear to raise immediate concerns, the deployment of adapted chat LMs, especially in under-monitored, low-resource language regions, warrants further analysis.
These models could inadvertently perpetuate harmful biases, compromise safety in ways not captured by current benchmarks, or be exploited for misinformation and other malicious purposes. Further research and responsible deployment strategies are crucial to address these potential risks.

\section*{Acknowledgments}
We would like to thank the Action Editor, Ruoyu Sun, the TMLR reviewers, Xi Wang, Sam Lewis-Lim, Maggie Mi, Huiyin Xue, and Xingwei Tan for their valuable feedback.
We acknowledge (1) IT Services at the University of Sheffield for the provision of services for high-performance computing; (2) the use of the University of Oxford Advanced Research Computing (ARC) facility; (3) EuroHPC Joint Undertaking for awarding us access to MeluXina at LuxProvide, Luxembourg; and (4) the use of resources provided by the Isambard-AI National AI Research Resource (AIRR). Isambard-AI is operated by the University of Bristol and is funded by the UK Government’s Department for Science, Innovation and Technology (DSIT) via UK Research and Innovation; and the Science and Technology Facilities Council [ST/AIRR/I-A-I/1023].
AY is supported by the Engineering and Physical Sciences Research Council (EPSRC)  [grant number EP/W524360/1] and the Japan Student Services Organization (JASSO) Student Exchange Support Program (Graduate Scholarship for Degree Seeking Students).

\bibliography{main,anthology}

\begin{thebibliography}{84}
\providecommand{\natexlab}[1]{#1}
\providecommand{\url}[1]{\texttt{#1}}
\expandafter\ifx\csname urlstyle\endcsname\relax
  \providecommand{\doi}[1]{doi: #1}\else
  \providecommand{\doi}{doi: \begingroup \urlstyle{rm}\Url}\fi

\bibitem[Abdin et~al.(2024{\natexlab{a}})Abdin, Aneja, Awadalla, Awadallah, Awan, Bach, Bahree, Bakhtiari, Bao, Behl, Benhaim, Bilenko, Bjorck, Bubeck, Cai, Cai, Chaudhary, Chen, Chen, Chen, Chen, Chen, Cheng, Chopra, Dai, Dixon, Eldan, Fragoso, Gao, Gao, Gao, Garg, Giorno, Goswami, Gunasekar, Haider, Hao, Hewett, Hu, Huynh, Iter, Jacobs, Javaheripi, Jin, Karampatziakis, Kauffmann, Khademi, Kim, Kim, Kurilenko, Lee, Lee, Li, Li, Liang, Liden, Lin, Lin, Liu, Liu, Liu, Liu, Liu, Luo, Madan, Mahmoudzadeh, Majercak, Mazzola, Mendes, Mitra, Modi, Nguyen, Norick, Patra, Perez-Becker, Portet, Pryzant, Qin, Radmilac, Ren, de~Rosa, Rosset, Roy, Ruwase, Saarikivi, Saied, Salim, Santacroce, Shah, Shang, Sharma, Shen, Shukla, Song, Tanaka, Tupini, Vaddamanu, Wang, Wang, Wang, Wang, Wang, Wang, Ward, Wen, Witte, Wu, Wu, Wyatt, Xiao, Xu, Xu, Xu, Xue, Yadav, Yang, Yang, Yang, Yang, Yu, Yuan, Zhang, Zhang, Zhang, Zhang, Zhang, Zhang, Zhang, and Zhou]{abdin2024phi3technicalreporthighly}
Marah Abdin, Jyoti Aneja, Hany Awadalla, Ahmed Awadallah, Ammar~Ahmad Awan, Nguyen Bach, Amit Bahree, Arash Bakhtiari, Jianmin Bao, Harkirat Behl, Alon Benhaim, Misha Bilenko, Johan Bjorck, Sébastien Bubeck, Martin Cai, Qin Cai, Vishrav Chaudhary, Dong Chen, Dongdong Chen, Weizhu Chen, Yen-Chun Chen, Yi-Ling Chen, Hao Cheng, Parul Chopra, Xiyang Dai, Matthew Dixon, Ronen Eldan, Victor Fragoso, Jianfeng Gao, Mei Gao, Min Gao, Amit Garg, Allie~Del Giorno, Abhishek Goswami, Suriya Gunasekar, Emman Haider, Junheng Hao, Russell~J. Hewett, Wenxiang Hu, Jamie Huynh, Dan Iter, Sam~Ade Jacobs, Mojan Javaheripi, Xin Jin, Nikos Karampatziakis, Piero Kauffmann, Mahoud Khademi, Dongwoo Kim, Young~Jin Kim, Lev Kurilenko, James~R. Lee, Yin~Tat Lee, Yuanzhi Li, Yunsheng Li, Chen Liang, Lars Liden, Xihui Lin, Zeqi Lin, Ce~Liu, Liyuan Liu, Mengchen Liu, Weishung Liu, Xiaodong Liu, Chong Luo, Piyush Madan, Ali Mahmoudzadeh, David Majercak, Matt Mazzola, Caio César~Teodoro Mendes, Arindam Mitra, Hardik Modi, Anh Nguyen,
  Brandon Norick, Barun Patra, Daniel Perez-Becker, Thomas Portet, Reid Pryzant, Heyang Qin, Marko Radmilac, Liliang Ren, Gustavo de~Rosa, Corby Rosset, Sambudha Roy, Olatunji Ruwase, Olli Saarikivi, Amin Saied, Adil Salim, Michael Santacroce, Shital Shah, Ning Shang, Hiteshi Sharma, Yelong Shen, Swadheen Shukla, Xia Song, Masahiro Tanaka, Andrea Tupini, Praneetha Vaddamanu, Chunyu Wang, Guanhua Wang, Lijuan Wang, Shuohang Wang, Xin Wang, Yu~Wang, Rachel Ward, Wen Wen, Philipp Witte, Haiping Wu, Xiaoxia Wu, Michael Wyatt, Bin Xiao, Can Xu, Jiahang Xu, Weijian Xu, Jilong Xue, Sonali Yadav, Fan Yang, Jianwei Yang, Yifan Yang, Ziyi Yang, Donghan Yu, Lu~Yuan, Chenruidong Zhang, Cyril Zhang, Jianwen Zhang, Li~Lyna Zhang, Yi~Zhang, Yue Zhang, Yunan Zhang, and Xiren Zhou.
\newblock Phi-3 technical report: A highly capable language model locally on your phone.
\newblock \emph{arXiv}, abs/2404.14219, 2024{\natexlab{a}}.
\newblock URL \url{https://arxiv.org/abs/2404.14219}.

\bibitem[Abdin et~al.(2024{\natexlab{b}})Abdin, Aneja, Behl, Bubeck, Eldan, Gunasekar, Harrison, Hewett, Javaheripi, Kauffmann, Lee, Lee, Li, Liu, Mendes, Nguyen, Price, de~Rosa, Saarikivi, Salim, Shah, Wang, Ward, Wu, Yu, Zhang, and Zhang]{abdin2024phi4technicalreport}
Marah Abdin, Jyoti Aneja, Harkirat Behl, Sébastien Bubeck, Ronen Eldan, Suriya Gunasekar, Michael Harrison, Russell~J. Hewett, Mojan Javaheripi, Piero Kauffmann, James~R. Lee, Yin~Tat Lee, Yuanzhi Li, Weishung Liu, Caio C.~T. Mendes, Anh Nguyen, Eric Price, Gustavo de~Rosa, Olli Saarikivi, Adil Salim, Shital Shah, Xin Wang, Rachel Ward, Yue Wu, Dingli Yu, Cyril Zhang, and Yi~Zhang.
\newblock Phi-4 technical report.
\newblock \emph{arXiv}, abs/2412.08905, 2024{\natexlab{b}}.
\newblock URL \url{https://arxiv.org/abs/2412.08905}.

\bibitem[Ahia et~al.(2023)Ahia, Kumar, Gonen, Kasai, Mortensen, Smith, and Tsvetkov]{ahia-etal-2023-languages}
Orevaoghene Ahia, Sachin Kumar, Hila Gonen, Jungo Kasai, David Mortensen, Noah Smith, and Yulia Tsvetkov.
\newblock Do all languages cost the same? tokenization in the era of commercial language models.
\newblock In Houda Bouamor, Juan Pino, and Kalika Bali (eds.), \emph{Proceedings of the 2023 Conference on Empirical Methods in Natural Language Processing}, pp.\  9904--9923, Singapore, December 2023. Association for Computational Linguistics.
\newblock \doi{10.18653/v1/2023.emnlp-main.614}.
\newblock URL \url{https://aclanthology.org/2023.emnlp-main.614/}.

\bibitem[Alexandrov et~al.(2024)Alexandrov, Raychev, M{\"u}ller, Zhang, Vechev, and Toutanova]{alexandrov-etal-2024-mitigating}
Anton Alexandrov, Veselin Raychev, Mark~Niklas M{\"u}ller, Ce~Zhang, Martin Vechev, and Kristina Toutanova.
\newblock Mitigating catastrophic forgetting in language transfer via model merging.
\newblock In Yaser Al-Onaizan, Mohit Bansal, and Yun-Nung Chen (eds.), \emph{Findings of the Association for Computational Linguistics: EMNLP 2024}, pp.\  17167--17186, Miami, Florida, USA, November 2024. Association for Computational Linguistics.
\newblock \doi{10.18653/v1/2024.findings-emnlp.1000}.
\newblock URL \url{https://aclanthology.org/2024.findings-emnlp.1000/}.

\bibitem[Ali et~al.(2024)Ali, Fromm, Thellmann, Rutmann, L{\"u}bbering, Leveling, Klug, Ebert, Doll, Buschhoff, Jain, Weber, Jurkschat, Abdelwahab, John, Ortiz~Suarez, Ostendorff, Weinbach, Sifa, Kesselheim, and Flores-Herr]{ali-etal-2024-tokenizer}
Mehdi Ali, Michael Fromm, Klaudia Thellmann, Richard Rutmann, Max L{\"u}bbering, Johannes Leveling, Katrin Klug, Jan Ebert, Niclas Doll, Jasper Buschhoff, Charvi Jain, Alexander Weber, Lena Jurkschat, Hammam Abdelwahab, Chelsea John, Pedro Ortiz~Suarez, Malte Ostendorff, Samuel Weinbach, Rafet Sifa, Stefan Kesselheim, and Nicolas Flores-Herr.
\newblock Tokenizer choice for {LLM} training: Negligible or crucial?
\newblock In Kevin Duh, Helena Gomez, and Steven Bethard (eds.), \emph{Findings of the Association for Computational Linguistics: NAACL 2024}, pp.\  3907--3924, Mexico City, Mexico, June 2024. Association for Computational Linguistics.
\newblock \doi{10.18653/v1/2024.findings-naacl.247}.
\newblock URL \url{https://aclanthology.org/2024.findings-naacl.247/}.

\bibitem[Ansel et~al.(2024)Ansel, Yang, He, Gimelshein, Jain, Voznesensky, Bao, Bell, Berard, Burovski, Chauhan, Chourdia, Constable, Desmaison, DeVito, Ellison, Feng, Gong, Gschwind, Hirsh, Huang, Kalambarkar, Kirsch, Lazos, Lezcano, Liang, Liang, Lu, Luk, Maher, Pan, Puhrsch, Reso, Saroufim, Siraichi, Suk, Zhang, Suo, Tillet, Zhao, Wang, Zhou, Zou, Wang, Mathews, Wen, Chanan, Wu, and Chintala]{10.1145/3620665.3640366}
Jason Ansel, Edward Yang, Horace He, Natalia Gimelshein, Animesh Jain, Michael Voznesensky, Bin Bao, Peter Bell, David Berard, Evgeni Burovski, Geeta Chauhan, Anjali Chourdia, Will Constable, Alban Desmaison, Zachary DeVito, Elias Ellison, Will Feng, Jiong Gong, Michael Gschwind, Brian Hirsh, Sherlock Huang, Kshiteej Kalambarkar, Laurent Kirsch, Michael Lazos, Mario Lezcano, Yanbo Liang, Jason Liang, Yinghai Lu, C.~K. Luk, Bert Maher, Yunjie Pan, Christian Puhrsch, Matthias Reso, Mark Saroufim, Marcos~Yukio Siraichi, Helen Suk, Shunting Zhang, Michael Suo, Phil Tillet, Xu~Zhao, Eikan Wang, Keren Zhou, Richard Zou, Xiaodong Wang, Ajit Mathews, William Wen, Gregory Chanan, Peng Wu, and Soumith Chintala.
\newblock Pytorch 2: Faster machine learning through dynamic python bytecode transformation and graph compilation.
\newblock In \emph{Proceedings of the 29th ACM International Conference on Architectural Support for Programming Languages and Operating Systems, Volume 2}, ASPLOS '24, pp.\  929–947, New York, NY, USA, 2024. Association for Computing Machinery.
\newblock ISBN 9798400703850.
\newblock \doi{10.1145/3620665.3640366}.
\newblock URL \url{https://doi.org/10.1145/3620665.3640366}.

\bibitem[Azime et~al.(2024)Azime, Tonja, Belay, Fuge, Wassie, Jada, Chanie, Sewunetie, and Yimam]{azime-etal-2024-walia}
Israel~Abebe Azime, Atnafu~Lambebo Tonja, Tadesse~Destaw Belay, Mitiku~Yohannes Fuge, Aman~Kassahun Wassie, Eyasu~Shiferaw Jada, Yonas Chanie, Walelign~Tewabe Sewunetie, and Seid~Muhie Yimam.
\newblock Walia-{LLM}: Enhancing {A}mharic-{LL}a{MA} by integrating task-specific and generative datasets.
\newblock In Yaser Al-Onaizan, Mohit Bansal, and Yun-Nung Chen (eds.), \emph{Findings of the Association for Computational Linguistics: EMNLP 2024}, pp.\  432--444, Miami, Florida, USA, November 2024. Association for Computational Linguistics.
\newblock \doi{10.18653/v1/2024.findings-emnlp.25}.
\newblock URL \url{https://aclanthology.org/2024.findings-emnlp.25/}.

\bibitem[Balachandran(2023)]{balachandran2023tamilllamanewtamillanguage}
Abhinand Balachandran.
\newblock {Tamil-Llama}: A new {Tamil} language model based on {L}lama 2.
\newblock \emph{arXiv}, abs/2311.05845, 2023.
\newblock URL \url{https://arxiv.org/abs/2311.05845}.

\bibitem[Bandarkar et~al.(2024)Bandarkar, Liang, Muller, Artetxe, Shukla, Husa, Goyal, Krishnan, Zettlemoyer, and Khabsa]{bandarkar-etal-2024-belebele}
Lucas Bandarkar, Davis Liang, Benjamin Muller, Mikel Artetxe, Satya~Narayan Shukla, Donald Husa, Naman Goyal, Abhinandan Krishnan, Luke Zettlemoyer, and Madian Khabsa.
\newblock The belebele benchmark: a parallel reading comprehension dataset in 122 language variants.
\newblock In Lun-Wei Ku, Andre Martins, and Vivek Srikumar (eds.), \emph{Proceedings of the 62nd Annual Meeting of the Association for Computational Linguistics (Volume 1: Long Papers)}, pp.\  749--775, Bangkok, Thailand, August 2024. Association for Computational Linguistics.
\newblock \doi{10.18653/v1/2024.acl-long.44}.
\newblock URL \url{https://aclanthology.org/2024.acl-long.44/}.

\bibitem[Bandarkar et~al.(2025)Bandarkar, Muller, Yuvraj, Hou, Singhal, Lv, and Liu]{bandarkar2024layerswappingzeroshotcrosslingual}
Lucas Bandarkar, Benjamin Muller, Pritish Yuvraj, Rui Hou, Nayan Singhal, Hongjiang Lv, and Bing Liu.
\newblock Layer swapping for zero-shot cross-lingual transfer in large language models.
\newblock In \emph{Proceedings of the Thirteenth International Conference on Learning Representations}, 2025.
\newblock URL \url{https://openreview.net/forum?id=vQhn4wrQ6j}.

\bibitem[Cahyawijaya et~al.(2024)Cahyawijaya, Lovenia, Koto, Putri, Cenggoro, Lee, Akbar, Dave, Nuurshadieq, Mahendra, Putri, Wilie, Winata, Aji, Purwarianti, and Fung]{cahyawijaya-etal-2024-cendol}
Samuel Cahyawijaya, Holy Lovenia, Fajri Koto, Rifki Putri, Wawan Cenggoro, Jhonson Lee, Salsabil Akbar, Emmanuel Dave, Nuurshadieq Nuurshadieq, Muhammad Mahendra, Rr~Putri, Bryan Wilie, Genta Winata, Alham Aji, Ayu Purwarianti, and Pascale Fung.
\newblock Cendol: Open instruction-tuned generative large language models for {I}ndonesian languages.
\newblock In Lun-Wei Ku, Andre Martins, and Vivek Srikumar (eds.), \emph{Proceedings of the 62nd Annual Meeting of the Association for Computational Linguistics (Volume 1: Long Papers)}, pp.\  14899--14914, Bangkok, Thailand, August 2024. Association for Computational Linguistics.
\newblock \doi{10.18653/v1/2024.acl-long.796}.
\newblock URL \url{https://aclanthology.org/2024.acl-long.796/}.

\bibitem[Choi et~al.(2024)Choi, Jeong, Park, Won, Lim, Kim, Kang, Yoon, Park, Lee, Lee, Hahm, Kim, and Lim]{choi-etal-2024-optimizing}
ChangSu Choi, Yongbin Jeong, Seoyoon Park, Inho Won, HyeonSeok Lim, SangMin Kim, Yejee Kang, Chanhyuk Yoon, Jaewan Park, Yiseul Lee, HyeJin Lee, Younggyun Hahm, Hansaem Kim, and KyungTae Lim.
\newblock Optimizing language augmentation for multilingual large language models: A case study on {K}orean.
\newblock In Nicoletta Calzolari, Min-Yen Kan, Veronique Hoste, Alessandro Lenci, Sakriani Sakti, and Nianwen Xue (eds.), \emph{Proceedings of the 2024 Joint International Conference on Computational Linguistics, Language Resources and Evaluation (LREC-COLING 2024)}, pp.\  12514--12526, Torino, Italia, May 2024. ELRA and ICCL.
\newblock URL \url{https://aclanthology.org/2024.lrec-main.1095/}.

\bibitem[Cobbe et~al.(2021)Cobbe, Kosaraju, Bavarian, Chen, Jun, Kaiser, Plappert, Tworek, Hilton, Nakano, Hesse, and Schulman]{cobbe2021trainingverifierssolvemath}
Karl Cobbe, Vineet Kosaraju, Mohammad Bavarian, Mark Chen, Heewoo Jun, Lukasz Kaiser, Matthias Plappert, Jerry Tworek, Jacob Hilton, Reiichiro Nakano, Christopher Hesse, and John Schulman.
\newblock Training verifiers to solve math word problems.
\newblock \emph{arXiv}, abs/2110.14168, 2021.
\newblock URL \url{https://arxiv.org/abs/2110.14168}.

\bibitem[Csaki et~al.(2023)Csaki, Pawakapan, Thakker, and Xu]{csaki2023efficientlyadaptingpretrainedlanguage}
Zoltan Csaki, Pian Pawakapan, Urmish Thakker, and Qiantong Xu.
\newblock Efficiently adapting pretrained language models to new languages.
\newblock \emph{arXiv}, abs/2311.05741, 2023.
\newblock URL \url{https://arxiv.org/abs/2311.05741}.

\bibitem[Cui et~al.(2024)Cui, Yang, and Yao]{cui2024efficienteffectivetextencoding}
Yiming Cui, Ziqing Yang, and Xin Yao.
\newblock Efficient and effective text encoding for {C}hinese {LLaMA} and {Alpaca}.
\newblock \emph{arXiv}, abs/2304.08177, 2024.
\newblock URL \url{https://arxiv.org/abs/2304.08177}.

\bibitem[Da~Dalt et~al.(2024)Da~Dalt, Llop, Baucells, Pamies, Xu, Gonzalez-Agirre, and Villegas]{da-dalt-etal-2024-flor}
Severino Da~Dalt, Joan Llop, Irene Baucells, Marc Pamies, Yishi Xu, Aitor Gonzalez-Agirre, and Marta Villegas.
\newblock {FLOR}: On the effectiveness of language adaptation.
\newblock In Nicoletta Calzolari, Min-Yen Kan, Veronique Hoste, Alessandro Lenci, Sakriani Sakti, and Nianwen Xue (eds.), \emph{Proceedings of the 2024 Joint International Conference on Computational Linguistics, Language Resources and Evaluation (LREC-COLING 2024)}, pp.\  7377--7388, Torino, Italia, May 2024. ELRA and ICCL.
\newblock URL \url{https://aclanthology.org/2024.lrec-main.650/}.

\bibitem[Dobler \& de~Melo(2024)Dobler and de~Melo]{dobler2024language}
Konstantin Dobler and Gerard de~Melo.
\newblock Language adaptation on a tight academic compute budget: Tokenizer swapping works and pure bfloat16 is enough.
\newblock In \emph{Proceedings of the 2nd Workshop on Advancing Neural Network Training: Computational Efficiency, Scalability, and Resource Optimization (WANT@ICML 2024)}, 2024.
\newblock URL \url{https://openreview.net/forum?id=VYfJaHeVod}.

\bibitem[Downey et~al.(2023)Downey, Blevins, Goldfine, and Steinert-Threlkeld]{downey-etal-2023-embedding}
C.m. Downey, Terra Blevins, Nora Goldfine, and Shane Steinert-Threlkeld.
\newblock Embedding structure matters: Comparing methods to adapt multilingual vocabularies to new languages.
\newblock In Duygu Ataman (ed.), \emph{Proceedings of the 3rd Workshop on Multi-lingual Representation Learning (MRL)}, pp.\  268--281, Singapore, December 2023. Association for Computational Linguistics.
\newblock \doi{10.18653/v1/2023.mrl-1.20}.
\newblock URL \url{https://aclanthology.org/2023.mrl-1.20/}.

\bibitem[Dubey et~al.(2024)Dubey, Jauhri, Pandey, Kadian, Al-Dahle, Letman, Mathur, Schelten, Yang, Fan, Goyal, Hartshorn, Yang, Mitra, Sravankumar, Korenev, Hinsvark, Rao, Zhang, Rodriguez, Gregerson, Spataru, Roziere, Biron, Tang, Chern, Caucheteux, Nayak, Bi, Marra, McConnell, Keller, Touret, Wu, Wong, Ferrer, Nikolaidis, Allonsius, Song, Pintz, Livshits, Esiobu, Choudhary, Mahajan, Garcia-Olano, Perino, Hupkes, Lakomkin, AlBadawy, Lobanova, Dinan, Smith, Radenovic, Zhang, Synnaeve, Lee, Anderson, Nail, Mialon, Pang, Cucurell, Nguyen, Korevaar, Xu, Touvron, Zarov, Ibarra, Kloumann, Misra, Evtimov, Copet, Lee, Geffert, Vranes, Park, Mahadeokar, Shah, van~der Linde, Billock, Hong, Lee, Fu, Chi, Huang, Liu, Wang, Yu, Bitton, Spisak, Park, Rocca, Johnstun, Saxe, Jia, Alwala, Upasani, Plawiak, Li, Heafield, Stone, El-Arini, Iyer, Malik, Chiu, Bhalla, Rantala-Yeary, van~der Maaten, Chen, Tan, Jenkins, Martin, Madaan, Malo, Blecher, Landzaat, de~Oliveira, Muzzi, Pasupuleti, Singh, Paluri, Kardas, Oldham, Rita,
  Pavlova, Kambadur, Lewis, Si, Singh, Hassan, Goyal, Torabi, Bashlykov, Bogoychev, Chatterji, Duchenne, Çelebi, Alrassy, Zhang, Li, Vasic, Weng, Bhargava, Dubal, Krishnan, Koura, Xu, He, Dong, Srinivasan, Ganapathy, Calderer, Cabral, Stojnic, Raileanu, Girdhar, Patel, Sauvestre, Polidoro, Sumbaly, Taylor, Silva, Hou, Wang, Hosseini, Chennabasappa, Singh, Bell, Kim, Edunov, Nie, Narang, Raparthy, Shen, Wan, Bhosale, Zhang, Vandenhende, Batra, Whitman, Sootla, Collot, Gururangan, Borodinsky, Herman, Fowler, Sheasha, Georgiou, Scialom, Speckbacher, Mihaylov, Xiao, Karn, Goswami, Gupta, Ramanathan, Kerkez, Gonguet, Do, Vogeti, Petrovic, Chu, Xiong, Fu, Meers, Martinet, Wang, Tan, Xie, Jia, Wang, Goldschlag, Gaur, Babaei, Wen, Song, Zhang, Li, Mao, Coudert, Yan, Chen, Papakipos, Singh, Grattafiori, Jain, Kelsey, Shajnfeld, Gangidi, Victoria, Goldstand, Menon, Sharma, Boesenberg, Vaughan, Baevski, Feinstein, Kallet, Sangani, Yunus, Lupu, Alvarado, Caples, Gu, Ho, Poulton, Ryan, Ramchandani, Franco, Saraf,
  Chowdhury, Gabriel, Bharambe, Eisenman, Yazdan, James, Maurer, Leonhardi, Huang, Loyd, Paola, Paranjape, Liu, Wu, Ni, Hancock, Wasti, Spence, Stojkovic, Gamido, Montalvo, Parker, Burton, Mejia, Wang, Kim, Zhou, Hu, Chu, Cai, Tindal, Feichtenhofer, Civin, Beaty, Kreymer, Li, Wyatt, Adkins, Xu, Testuggine, David, Parikh, Liskovich, Foss, Wang, Le, Holland, Dowling, Jamil, Montgomery, Presani, Hahn, Wood, Brinkman, Arcaute, Dunbar, Smothers, Sun, Kreuk, Tian, Ozgenel, Caggioni, Guzmán, Kanayet, Seide, Florez, Schwarz, Badeer, Swee, Halpern, Thattai, Herman, Sizov, Guangyi, Zhang, Lakshminarayanan, Shojanazeri, Zou, Wang, Zha, Habeeb, Rudolph, Suk, Aspegren, Goldman, Damlaj, Molybog, Tufanov, Veliche, Gat, Weissman, Geboski, Kohli, Asher, Gaya, Marcus, Tang, Chan, Zhen, Reizenstein, Teboul, Zhong, Jin, Yang, Cummings, Carvill, Shepard, McPhie, Torres, Ginsburg, Wang, Wu, U, Saxena, Prasad, Khandelwal, Zand, Matosich, Veeraraghavan, Michelena, Li, Huang, Chawla, Lakhotia, Huang, Chen, Garg, A, Silva, Bell,
  Zhang, Guo, Yu, Moshkovich, Wehrstedt, Khabsa, Avalani, Bhatt, Tsimpoukelli, Mankus, Hasson, Lennie, Reso, Groshev, Naumov, Lathi, Keneally, Seltzer, Valko, Restrepo, Patel, Vyatskov, Samvelyan, Clark, Macey, Wang, Hermoso, Metanat, Rastegari, Bansal, Santhanam, Parks, White, Bawa, Singhal, Egebo, Usunier, Laptev, Dong, Zhang, Cheng, Chernoguz, Hart, Salpekar, Kalinli, Kent, Parekh, Saab, Balaji, Rittner, Bontrager, Roux, Dollar, Zvyagina, Ratanchandani, Yuvraj, Liang, Alao, Rodriguez, Ayub, Murthy, Nayani, Mitra, Li, Hogan, Battey, Wang, Maheswari, Howes, Rinott, Bondu, Datta, Chugh, Hunt, Dhillon, Sidorov, Pan, Verma, Yamamoto, Ramaswamy, Lindsay, Lindsay, Feng, Lin, Zha, Shankar, Zhang, Zhang, Wang, Agarwal, Sajuyigbe, Chintala, Max, Chen, Kehoe, Satterfield, Govindaprasad, Gupta, Cho, Virk, Subramanian, Choudhury, Goldman, Remez, Glaser, Best, Kohler, Robinson, Li, Zhang, Matthews, Chou, Shaked, Vontimitta, Ajayi, Montanez, Mohan, Kumar, Mangla, Albiero, Ionescu, Poenaru, Mihailescu, Ivanov, Li, Wang,
  Jiang, Bouaziz, Constable, Tang, Wang, Wu, Wang, Xia, Wu, Gao, Chen, Hu, Jia, Qi, Li, Zhang, Zhang, Adi, Nam, Yu, Wang, Hao, Qian, He, Rait, DeVito, Rosnbrick, Wen, Yang, and Zhao]{dubey2024llama3herdmodels}
Abhimanyu Dubey, Abhinav Jauhri, Abhinav Pandey, Abhishek Kadian, Ahmad Al-Dahle, Aiesha Letman, Akhil Mathur, Alan Schelten, Amy Yang, Angela Fan, Anirudh Goyal, Anthony Hartshorn, Aobo Yang, Archi Mitra, Archie Sravankumar, Artem Korenev, Arthur Hinsvark, Arun Rao, Aston Zhang, Aurelien Rodriguez, Austen Gregerson, Ava Spataru, Baptiste Roziere, Bethany Biron, Binh Tang, Bobbie Chern, Charlotte Caucheteux, Chaya Nayak, Chloe Bi, Chris Marra, Chris McConnell, Christian Keller, Christophe Touret, Chunyang Wu, Corinne Wong, Cristian~Canton Ferrer, Cyrus Nikolaidis, Damien Allonsius, Daniel Song, Danielle Pintz, Danny Livshits, David Esiobu, Dhruv Choudhary, Dhruv Mahajan, Diego Garcia-Olano, Diego Perino, Dieuwke Hupkes, Egor Lakomkin, Ehab AlBadawy, Elina Lobanova, Emily Dinan, Eric~Michael Smith, Filip Radenovic, Frank Zhang, Gabriel Synnaeve, Gabrielle Lee, Georgia~Lewis Anderson, Graeme Nail, Gregoire Mialon, Guan Pang, Guillem Cucurell, Hailey Nguyen, Hannah Korevaar, Hu~Xu, Hugo Touvron, Iliyan Zarov,
  Imanol~Arrieta Ibarra, Isabel Kloumann, Ishan Misra, Ivan Evtimov, Jade Copet, Jaewon Lee, Jan Geffert, Jana Vranes, Jason Park, Jay Mahadeokar, Jeet Shah, Jelmer van~der Linde, Jennifer Billock, Jenny Hong, Jenya Lee, Jeremy Fu, Jianfeng Chi, Jianyu Huang, Jiawen Liu, Jie Wang, Jiecao Yu, Joanna Bitton, Joe Spisak, Jongsoo Park, Joseph Rocca, Joshua Johnstun, Joshua Saxe, Junteng Jia, Kalyan~Vasuden Alwala, Kartikeya Upasani, Kate Plawiak, Ke~Li, Kenneth Heafield, Kevin Stone, Khalid El-Arini, Krithika Iyer, Kshitiz Malik, Kuenley Chiu, Kunal Bhalla, Lauren Rantala-Yeary, Laurens van~der Maaten, Lawrence Chen, Liang Tan, Liz Jenkins, Louis Martin, Lovish Madaan, Lubo Malo, Lukas Blecher, Lukas Landzaat, Luke de~Oliveira, Madeline Muzzi, Mahesh Pasupuleti, Mannat Singh, Manohar Paluri, Marcin Kardas, Mathew Oldham, Mathieu Rita, Maya Pavlova, Melanie Kambadur, Mike Lewis, Min Si, Mitesh~Kumar Singh, Mona Hassan, Naman Goyal, Narjes Torabi, Nikolay Bashlykov, Nikolay Bogoychev, Niladri Chatterji, Olivier
  Duchenne, Onur Çelebi, Patrick Alrassy, Pengchuan Zhang, Pengwei Li, Petar Vasic, Peter Weng, Prajjwal Bhargava, Pratik Dubal, Praveen Krishnan, Punit~Singh Koura, Puxin Xu, Qing He, Qingxiao Dong, Ragavan Srinivasan, Raj Ganapathy, Ramon Calderer, Ricardo~Silveira Cabral, Robert Stojnic, Roberta Raileanu, Rohit Girdhar, Rohit Patel, Romain Sauvestre, Ronnie Polidoro, Roshan Sumbaly, Ross Taylor, Ruan Silva, Rui Hou, Rui Wang, Saghar Hosseini, Sahana Chennabasappa, Sanjay Singh, Sean Bell, Seohyun~Sonia Kim, Sergey Edunov, Shaoliang Nie, Sharan Narang, Sharath Raparthy, Sheng Shen, Shengye Wan, Shruti Bhosale, Shun Zhang, Simon Vandenhende, Soumya Batra, Spencer Whitman, Sten Sootla, Stephane Collot, Suchin Gururangan, Sydney Borodinsky, Tamar Herman, Tara Fowler, Tarek Sheasha, Thomas Georgiou, Thomas Scialom, Tobias Speckbacher, Todor Mihaylov, Tong Xiao, Ujjwal Karn, Vedanuj Goswami, Vibhor Gupta, Vignesh Ramanathan, Viktor Kerkez, Vincent Gonguet, Virginie Do, Vish Vogeti, Vladan Petrovic, Weiwei Chu,
  Wenhan Xiong, Wenyin Fu, Whitney Meers, Xavier Martinet, Xiaodong Wang, Xiaoqing~Ellen Tan, Xinfeng Xie, Xuchao Jia, Xuewei Wang, Yaelle Goldschlag, Yashesh Gaur, Yasmine Babaei, Yi~Wen, Yiwen Song, Yuchen Zhang, Yue Li, Yuning Mao, Zacharie~Delpierre Coudert, Zheng Yan, Zhengxing Chen, Zoe Papakipos, Aaditya Singh, Aaron Grattafiori, Abha Jain, Adam Kelsey, Adam Shajnfeld, Adithya Gangidi, Adolfo Victoria, Ahuva Goldstand, Ajay Menon, Ajay Sharma, Alex Boesenberg, Alex Vaughan, Alexei Baevski, Allie Feinstein, Amanda Kallet, Amit Sangani, Anam Yunus, Andrei Lupu, Andres Alvarado, Andrew Caples, Andrew Gu, Andrew Ho, Andrew Poulton, Andrew Ryan, Ankit Ramchandani, Annie Franco, Aparajita Saraf, Arkabandhu Chowdhury, Ashley Gabriel, Ashwin Bharambe, Assaf Eisenman, Azadeh Yazdan, Beau James, Ben Maurer, Benjamin Leonhardi, Bernie Huang, Beth Loyd, Beto~De Paola, Bhargavi Paranjape, Bing Liu, Bo~Wu, Boyu Ni, Braden Hancock, Bram Wasti, Brandon Spence, Brani Stojkovic, Brian Gamido, Britt Montalvo, Carl
  Parker, Carly Burton, Catalina Mejia, Changhan Wang, Changkyu Kim, Chao Zhou, Chester Hu, Ching-Hsiang Chu, Chris Cai, Chris Tindal, Christoph Feichtenhofer, Damon Civin, Dana Beaty, Daniel Kreymer, Daniel Li, Danny Wyatt, David Adkins, David Xu, Davide Testuggine, Delia David, Devi Parikh, Diana Liskovich, Didem Foss, Dingkang Wang, Duc Le, Dustin Holland, Edward Dowling, Eissa Jamil, Elaine Montgomery, Eleonora Presani, Emily Hahn, Emily Wood, Erik Brinkman, Esteban Arcaute, Evan Dunbar, Evan Smothers, Fei Sun, Felix Kreuk, Feng Tian, Firat Ozgenel, Francesco Caggioni, Francisco Guzmán, Frank Kanayet, Frank Seide, Gabriela~Medina Florez, Gabriella Schwarz, Gada Badeer, Georgia Swee, Gil Halpern, Govind Thattai, Grant Herman, Grigory Sizov, Guangyi, Zhang, Guna Lakshminarayanan, Hamid Shojanazeri, Han Zou, Hannah Wang, Hanwen Zha, Haroun Habeeb, Harrison Rudolph, Helen Suk, Henry Aspegren, Hunter Goldman, Ibrahim Damlaj, Igor Molybog, Igor Tufanov, Irina-Elena Veliche, Itai Gat, Jake Weissman, James
  Geboski, James Kohli, Japhet Asher, Jean-Baptiste Gaya, Jeff Marcus, Jeff Tang, Jennifer Chan, Jenny Zhen, Jeremy Reizenstein, Jeremy Teboul, Jessica Zhong, Jian Jin, Jingyi Yang, Joe Cummings, Jon Carvill, Jon Shepard, Jonathan McPhie, Jonathan Torres, Josh Ginsburg, Junjie Wang, Kai Wu, Kam~Hou U, Karan Saxena, Karthik Prasad, Kartikay Khandelwal, Katayoun Zand, Kathy Matosich, Kaushik Veeraraghavan, Kelly Michelena, Keqian Li, Kun Huang, Kunal Chawla, Kushal Lakhotia, Kyle Huang, Lailin Chen, Lakshya Garg, Lavender A, Leandro Silva, Lee Bell, Lei Zhang, Liangpeng Guo, Licheng Yu, Liron Moshkovich, Luca Wehrstedt, Madian Khabsa, Manav Avalani, Manish Bhatt, Maria Tsimpoukelli, Martynas Mankus, Matan Hasson, Matthew Lennie, Matthias Reso, Maxim Groshev, Maxim Naumov, Maya Lathi, Meghan Keneally, Michael~L. Seltzer, Michal Valko, Michelle Restrepo, Mihir Patel, Mik Vyatskov, Mikayel Samvelyan, Mike Clark, Mike Macey, Mike Wang, Miquel~Jubert Hermoso, Mo~Metanat, Mohammad Rastegari, Munish Bansal, Nandhini
  Santhanam, Natascha Parks, Natasha White, Navyata Bawa, Nayan Singhal, Nick Egebo, Nicolas Usunier, Nikolay~Pavlovich Laptev, Ning Dong, Ning Zhang, Norman Cheng, Oleg Chernoguz, Olivia Hart, Omkar Salpekar, Ozlem Kalinli, Parkin Kent, Parth Parekh, Paul Saab, Pavan Balaji, Pedro Rittner, Philip Bontrager, Pierre Roux, Piotr Dollar, Polina Zvyagina, Prashant Ratanchandani, Pritish Yuvraj, Qian Liang, Rachad Alao, Rachel Rodriguez, Rafi Ayub, Raghotham Murthy, Raghu Nayani, Rahul Mitra, Raymond Li, Rebekkah Hogan, Robin Battey, Rocky Wang, Rohan Maheswari, Russ Howes, Ruty Rinott, Sai~Jayesh Bondu, Samyak Datta, Sara Chugh, Sara Hunt, Sargun Dhillon, Sasha Sidorov, Satadru Pan, Saurabh Verma, Seiji Yamamoto, Sharadh Ramaswamy, Shaun Lindsay, Shaun Lindsay, Sheng Feng, Shenghao Lin, Shengxin~Cindy Zha, Shiva Shankar, Shuqiang Zhang, Shuqiang Zhang, Sinong Wang, Sneha Agarwal, Soji Sajuyigbe, Soumith Chintala, Stephanie Max, Stephen Chen, Steve Kehoe, Steve Satterfield, Sudarshan Govindaprasad, Sumit Gupta,
  Sungmin Cho, Sunny Virk, Suraj Subramanian, Sy~Choudhury, Sydney Goldman, Tal Remez, Tamar Glaser, Tamara Best, Thilo Kohler, Thomas Robinson, Tianhe Li, Tianjun Zhang, Tim Matthews, Timothy Chou, Tzook Shaked, Varun Vontimitta, Victoria Ajayi, Victoria Montanez, Vijai Mohan, Vinay~Satish Kumar, Vishal Mangla, Vítor Albiero, Vlad Ionescu, Vlad Poenaru, Vlad~Tiberiu Mihailescu, Vladimir Ivanov, Wei Li, Wenchen Wang, Wenwen Jiang, Wes Bouaziz, Will Constable, Xiaocheng Tang, Xiaofang Wang, Xiaojian Wu, Xiaolan Wang, Xide Xia, Xilun Wu, Xinbo Gao, Yanjun Chen, Ye~Hu, Ye~Jia, Ye~Qi, Yenda Li, Yilin Zhang, Ying Zhang, Yossi Adi, Youngjin Nam, Yu, Wang, Yuchen Hao, Yundi Qian, Yuzi He, Zach Rait, Zachary DeVito, Zef Rosnbrick, Zhaoduo Wen, Zhenyu Yang, and Zhiwei Zhao.
\newblock The {L}lama 3 herd of models.
\newblock \emph{arXiv}, abs/2407.21783, 2024.
\newblock URL \url{https://arxiv.org/abs/2407.21783}.

\bibitem[Dubois et~al.(2024)Dubois, Liang, and Hashimoto]{dubois2024length}
Yann Dubois, Percy Liang, and Tatsunori Hashimoto.
\newblock Length-controlled {AlpacaEval}: A simple debiasing of automatic evaluators.
\newblock In \emph{Proceedings of the First Conference on Language Modeling}, 2024.
\newblock URL \url{https://openreview.net/forum?id=CybBmzWBX0}.

\bibitem[Ebrahimi et~al.(2023)Ebrahimi, Mager, Rijhwani, Rice, Oncevay, Baltazar, Cort{\'e}s, Monta{\~n}o, Ortega, Coto-solano, Cruz, Palmer, and Kann]{ebrahimi-etal-2023-findings}
Abteen Ebrahimi, Manuel Mager, Shruti Rijhwani, Enora Rice, Arturo Oncevay, Claudia Baltazar, Mar{\'i}a Cort{\'e}s, Cynthia Monta{\~n}o, John~E. Ortega, Rolando Coto-solano, Hilaria Cruz, Alexis Palmer, and Katharina Kann.
\newblock Findings of the {A}mericas{NLP} 2023 shared task on machine translation into indigenous languages.
\newblock In Manuel Mager, Abteen Ebrahimi, Arturo Oncevay, Enora Rice, Shruti Rijhwani, Alexis Palmer, and Katharina Kann (eds.), \emph{Proceedings of the Workshop on Natural Language Processing for Indigenous Languages of the Americas (AmericasNLP)}, pp.\  206--219, Toronto, Canada, July 2023. Association for Computational Linguistics.
\newblock \doi{10.18653/v1/2023.americasnlp-1.23}.
\newblock URL \url{https://aclanthology.org/2023.americasnlp-1.23/}.

\bibitem[ElSherief et~al.(2021)ElSherief, Ziems, Muchlinski, Anupindi, Seybolt, De~Choudhury, and Yang]{elsherief-etal-2021-latent}
Mai ElSherief, Caleb Ziems, David Muchlinski, Vaishnavi Anupindi, Jordyn Seybolt, Munmun De~Choudhury, and Diyi Yang.
\newblock Latent hatred: A benchmark for understanding implicit hate speech.
\newblock In Marie-Francine Moens, Xuanjing Huang, Lucia Specia, and Scott Wen-tau Yih (eds.), \emph{Proceedings of the 2021 Conference on Empirical Methods in Natural Language Processing}, pp.\  345--363, Online and Punta Cana, Dominican Republic, November 2021. Association for Computational Linguistics.
\newblock \doi{10.18653/v1/2021.emnlp-main.29}.
\newblock URL \url{https://aclanthology.org/2021.emnlp-main.29/}.

\bibitem[Fourrier et~al.(2023)Fourrier, Habib, Wolf, and Tunstall]{lighteval}
Clémentine Fourrier, Nathan Habib, Thomas Wolf, and Lewis Tunstall.
\newblock {LightEval}: A lightweight framework for {LLM} evaluation.
\newblock \url{https://github.com/huggingface/lighteval}, 2023.

\bibitem[Fujii et~al.(2024)Fujii, Nakamura, Loem, Iida, Ohi, Hattori, Shota, Mizuki, Yokota, and Okazaki]{fujii2024continual}
Kazuki Fujii, Taishi Nakamura, Mengsay Loem, Hiroki Iida, Masanari Ohi, Kakeru Hattori, Hirai Shota, Sakae Mizuki, Rio Yokota, and Naoaki Okazaki.
\newblock Continual pre-training for cross-lingual {LLM} adaptation: Enhancing {J}apanese language capabilities.
\newblock In \emph{Proceedings of the First Conference on Language Modeling}, 2024.
\newblock URL \url{https://openreview.net/forum?id=TQdd1VhWbe}.

\bibitem[Gao et~al.(2023)Gao, Tow, Abbasi, Biderman, Black, DiPofi, Foster, Golding, Hsu, Le~Noac'h, Li, McDonell, Muennighoff, Ociepa, Phang, Reynolds, Schoelkopf, Skowron, Sutawika, Tang, Thite, Wang, Wang, and Zou]{eval-harness}
Leo Gao, Jonathan Tow, Baber Abbasi, Stella Biderman, Sid Black, Anthony DiPofi, Charles Foster, Laurence Golding, Jeffrey Hsu, Alain Le~Noac'h, Haonan Li, Kyle McDonell, Niklas Muennighoff, Chris Ociepa, Jason Phang, Laria Reynolds, Hailey Schoelkopf, Aviya Skowron, Lintang Sutawika, Eric Tang, Anish Thite, Ben Wang, Kevin Wang, and Andy Zou.
\newblock A framework for few-shot language model evaluation.
\newblock \url{https://zenodo.org/records/10256836}, 2023.

\bibitem[{Gemma Team} et~al.(2024){Gemma Team}, Riviere, Pathak, Sessa, Hardin, Bhupatiraju, Hussenot, Mesnard, Shahriari, Ramé, Ferret, Liu, Tafti, Friesen, Casbon, Ramos, Kumar, Lan, Jerome, Tsitsulin, Vieillard, Stanczyk, Girgin, Momchev, Hoffman, Thakoor, Grill, Neyshabur, Bachem, Walton, Severyn, Parrish, Ahmad, Hutchison, Abdagic, Carl, Shen, Brock, Coenen, Laforge, Paterson, Bastian, Piot, Wu, Royal, Chen, Kumar, Perry, Welty, Choquette-Choo, Sinopalnikov, Weinberger, Vijaykumar, Rogozińska, Herbison, Bandy, Wang, Noland, Moreira, Senter, Eltyshev, Visin, Rasskin, Wei, Cameron, Martins, Hashemi, Klimczak-Plucińska, Batra, Dhand, Nardini, Mein, Zhou, Svensson, Stanway, Chan, Zhou, Carrasqueira, Iljazi, Becker, Fernandez, van Amersfoort, Gordon, Lipschultz, Newlan, yeong Ji, Mohamed, Badola, Black, Millican, McDonell, Nguyen, Sodhia, Greene, Sjoesund, Usui, Sifre, Heuermann, Lago, McNealus, Soares, Kilpatrick, Dixon, Martins, Reid, Singh, Iverson, Görner, Velloso, Wirth, Davidow, Miller, Rahtz,
  Watson, Risdal, Kazemi, Moynihan, Zhang, Kahng, Park, Rahman, Khatwani, Dao, Bardoliwalla, Devanathan, Dumai, Chauhan, Wahltinez, Botarda, Barnes, Barham, Michel, Jin, Georgiev, Culliton, Kuppala, Comanescu, Merhej, Jana, Rokni, Agarwal, Mullins, Saadat, Carthy, Cogan, Perrin, Arnold, Krause, Dai, Garg, Sheth, Ronstrom, Chan, Jordan, Yu, Eccles, Hennigan, Kocisky, Doshi, Jain, Yadav, Meshram, Dharmadhikari, Barkley, Wei, Ye, Han, Kwon, Xu, Shen, Gong, Wei, Cotruta, Kirk, Rao, Giang, Peran, Warkentin, Collins, Barral, Ghahramani, Hadsell, Sculley, Banks, Dragan, Petrov, Vinyals, Dean, Hassabis, Kavukcuoglu, Farabet, Buchatskaya, Borgeaud, Fiedel, Joulin, Kenealy, Dadashi, and Andreev]{gemmateam2024gemma2improvingopen}
{Gemma Team}, Morgane Riviere, Shreya Pathak, Pier~Giuseppe Sessa, Cassidy Hardin, Surya Bhupatiraju, Léonard Hussenot, Thomas Mesnard, Bobak Shahriari, Alexandre Ramé, Johan Ferret, Peter Liu, Pouya Tafti, Abe Friesen, Michelle Casbon, Sabela Ramos, Ravin Kumar, Charline~Le Lan, Sammy Jerome, Anton Tsitsulin, Nino Vieillard, Piotr Stanczyk, Sertan Girgin, Nikola Momchev, Matt Hoffman, Shantanu Thakoor, Jean-Bastien Grill, Behnam Neyshabur, Olivier Bachem, Alanna Walton, Aliaksei Severyn, Alicia Parrish, Aliya Ahmad, Allen Hutchison, Alvin Abdagic, Amanda Carl, Amy Shen, Andy Brock, Andy Coenen, Anthony Laforge, Antonia Paterson, Ben Bastian, Bilal Piot, Bo~Wu, Brandon Royal, Charlie Chen, Chintu Kumar, Chris Perry, Chris Welty, Christopher~A. Choquette-Choo, Danila Sinopalnikov, David Weinberger, Dimple Vijaykumar, Dominika Rogozińska, Dustin Herbison, Elisa Bandy, Emma Wang, Eric Noland, Erica Moreira, Evan Senter, Evgenii Eltyshev, Francesco Visin, Gabriel Rasskin, Gary Wei, Glenn Cameron, Gus Martins,
  Hadi Hashemi, Hanna Klimczak-Plucińska, Harleen Batra, Harsh Dhand, Ivan Nardini, Jacinda Mein, Jack Zhou, James Svensson, Jeff Stanway, Jetha Chan, Jin~Peng Zhou, Joana Carrasqueira, Joana Iljazi, Jocelyn Becker, Joe Fernandez, Joost van Amersfoort, Josh Gordon, Josh Lipschultz, Josh Newlan, Ju~yeong Ji, Kareem Mohamed, Kartikeya Badola, Kat Black, Katie Millican, Keelin McDonell, Kelvin Nguyen, Kiranbir Sodhia, Kish Greene, Lars~Lowe Sjoesund, Lauren Usui, Laurent Sifre, Lena Heuermann, Leticia Lago, Lilly McNealus, Livio~Baldini Soares, Logan Kilpatrick, Lucas Dixon, Luciano Martins, Machel Reid, Manvinder Singh, Mark Iverson, Martin Görner, Mat Velloso, Mateo Wirth, Matt Davidow, Matt Miller, Matthew Rahtz, Matthew Watson, Meg Risdal, Mehran Kazemi, Michael Moynihan, Ming Zhang, Minsuk Kahng, Minwoo Park, Mofi Rahman, Mohit Khatwani, Natalie Dao, Nenshad Bardoliwalla, Nesh Devanathan, Neta Dumai, Nilay Chauhan, Oscar Wahltinez, Pankil Botarda, Parker Barnes, Paul Barham, Paul Michel, Pengchong Jin,
  Petko Georgiev, Phil Culliton, Pradeep Kuppala, Ramona Comanescu, Ramona Merhej, Reena Jana, Reza~Ardeshir Rokni, Rishabh Agarwal, Ryan Mullins, Samaneh Saadat, Sara~Mc Carthy, Sarah Cogan, Sarah Perrin, Sébastien M.~R. Arnold, Sebastian Krause, Shengyang Dai, Shruti Garg, Shruti Sheth, Sue Ronstrom, Susan Chan, Timothy Jordan, Ting Yu, Tom Eccles, Tom Hennigan, Tomas Kocisky, Tulsee Doshi, Vihan Jain, Vikas Yadav, Vilobh Meshram, Vishal Dharmadhikari, Warren Barkley, Wei Wei, Wenming Ye, Woohyun Han, Woosuk Kwon, Xiang Xu, Zhe Shen, Zhitao Gong, Zichuan Wei, Victor Cotruta, Phoebe Kirk, Anand Rao, Minh Giang, Ludovic Peran, Tris Warkentin, Eli Collins, Joelle Barral, Zoubin Ghahramani, Raia Hadsell, D.~Sculley, Jeanine Banks, Anca Dragan, Slav Petrov, Oriol Vinyals, Jeff Dean, Demis Hassabis, Koray Kavukcuoglu, Clement Farabet, Elena Buchatskaya, Sebastian Borgeaud, Noah Fiedel, Armand Joulin, Kathleen Kenealy, Robert Dadashi, and Alek Andreev.
\newblock Gemma 2: Improving open language models at a practical size.
\newblock \emph{arXiv}, abs/2408.00118, 2024.
\newblock URL \url{https://arxiv.org/abs/2408.00118}.

\bibitem[Geng et~al.(2025)Geng, Zhu, Li, Lai, Zou, She, GUO, Zhao, Li, Li, Su, Zhao, Lyu, Zhang, Chen, Yang, and Huang]{geng2025why}
Xiang Geng, Ming Zhu, Jiahuan Li, Zhejian Lai, Wei Zou, Shuaijie She, Jiaxin GUO, Xiaofeng Zhao, Yinglu Li, Yuang Li, Chang Su, Yanqing Zhao, Xinglin Lyu, Min Zhang, Jiajun Chen, Hao Yang, and Shujian Huang.
\newblock Why not transform chat large language models to non-{E}nglish?
\newblock \emph{Frontiers of Computer Science}, 2025.
\newblock ISSN 2095-2228.
\newblock \doi{https://doi.org/10.1007/s11704-025-50646-z}.
\newblock URL \url{https://journal.hep.com.cn/fcs/EN/10.1007/s11704-025-50646-z}.

\bibitem[Goddard et~al.(2024)Goddard, Siriwardhana, Ehghaghi, Meyers, Karpukhin, Benedict, McQuade, and Solawetz]{goddard-etal-2024-arcees}
Charles Goddard, Shamane Siriwardhana, Malikeh Ehghaghi, Luke Meyers, Vladimir Karpukhin, Brian Benedict, Mark McQuade, and Jacob Solawetz.
\newblock Arcee`s {M}erge{K}it: A toolkit for merging large language models.
\newblock In Franck Dernoncourt, Daniel Preo{\c{t}}iuc-Pietro, and Anastasia Shimorina (eds.), \emph{Proceedings of the 2024 Conference on Empirical Methods in Natural Language Processing: Industry Track}, pp.\  477--485, Miami, Florida, US, November 2024. Association for Computational Linguistics.
\newblock \doi{10.18653/v1/2024.emnlp-industry.36}.
\newblock URL \url{https://aclanthology.org/2024.emnlp-industry.36/}.

\bibitem[Han et~al.(2025{\natexlab{a}})Han, Eriguchi, Xu, Hoang, Carpuat, and Khayrallah]{han2024adaptersalteringllmvocabularies}
HyoJung Han, Akiko Eriguchi, Haoran Xu, Hieu Hoang, Marine Carpuat, and Huda Khayrallah.
\newblock Adapters for altering {LLM} vocabularies: What languages benefit the most?
\newblock In \emph{Proceedings of the Thirteenth International Conference on Learning Representations}, 2025{\natexlab{a}}.
\newblock URL \url{https://openreview.net/forum?id=KxQRHOre9D}.

\bibitem[Han et~al.(2025{\natexlab{b}})Han, Suk, An, Kim, Kim, Yang, Choi, and Shin]{han2025trillion7btechnicalreport}
Sungjun Han, Juyoung Suk, Suyeong An, Hyungguk Kim, Kyuseok Kim, Wonsuk Yang, Seungtaek Choi, and Jamin Shin.
\newblock Trillion 7b technical report.
\newblock \emph{arXiv}, abs/2504.15431, 2025{\natexlab{b}}.
\newblock URL \url{https://arxiv.org/abs/2504.15431}.

\bibitem[Hartvigsen et~al.(2022)Hartvigsen, Gabriel, Palangi, Sap, Ray, and Kamar]{hartvigsen-etal-2022-toxigen}
Thomas Hartvigsen, Saadia Gabriel, Hamid Palangi, Maarten Sap, Dipankar Ray, and Ece Kamar.
\newblock {T}oxi{G}en: A large-scale machine-generated dataset for adversarial and implicit hate speech detection.
\newblock In Smaranda Muresan, Preslav Nakov, and Aline Villavicencio (eds.), \emph{Proceedings of the 60th Annual Meeting of the Association for Computational Linguistics (Volume 1: Long Papers)}, pp.\  3309--3326, Dublin, Ireland, May 2022. Association for Computational Linguistics.
\newblock \doi{10.18653/v1/2022.acl-long.234}.
\newblock URL \url{https://aclanthology.org/2022.acl-long.234/}.

\bibitem[Hasan et~al.(2021)Hasan, Bhattacharjee, Islam, Mubasshir, Li, Kang, Rahman, and Shahriyar]{hasan-etal-2021-xl}
Tahmid Hasan, Abhik Bhattacharjee, Md.~Saiful Islam, Kazi Mubasshir, Yuan-Fang Li, Yong-Bin Kang, M.~Sohel Rahman, and Rifat Shahriyar.
\newblock {XL}-sum: Large-scale multilingual abstractive summarization for 44 languages.
\newblock In Chengqing Zong, Fei Xia, Wenjie Li, and Roberto Navigli (eds.), \emph{Findings of the Association for Computational Linguistics: ACL-IJCNLP 2021}, pp.\  4693--4703, Online, August 2021. Association for Computational Linguistics.
\newblock \doi{10.18653/v1/2021.findings-acl.413}.
\newblock URL \url{https://aclanthology.org/2021.findings-acl.413/}.

\bibitem[Hendrycks et~al.(2021)Hendrycks, Burns, Basart, Zou, Mazeika, Song, and Steinhardt]{hendrycks2021measuring}
Dan Hendrycks, Collin Burns, Steven Basart, Andy Zou, Mantas Mazeika, Dawn Song, and Jacob Steinhardt.
\newblock Measuring massive multitask language understanding.
\newblock In \emph{Proceedings of the Nineth International Conference on Learning Representations}, 2021.
\newblock URL \url{https://openreview.net/forum?id=d7KBjmI3GmQ}.

\bibitem[Hong et~al.(2024)Hong, Lee, and Cho]{hong-etal-2024-accelerating}
Jimin Hong, Gibbeum Lee, and Jaewoong Cho.
\newblock Accelerating multilingual language model for excessively tokenized languages.
\newblock In Lun-Wei Ku, Andre Martins, and Vivek Srikumar (eds.), \emph{Findings of the Association for Computational Linguistics: ACL 2024}, pp.\  11095--11111, Bangkok, Thailand, August 2024. Association for Computational Linguistics.
\newblock \doi{10.18653/v1/2024.findings-acl.660}.
\newblock URL \url{https://aclanthology.org/2024.findings-acl.660/}.

\bibitem[Hu et~al.(2022)Hu, yelong shen, Wallis, Allen-Zhu, Li, Wang, Wang, and Chen]{hu2022lora}
Edward~J Hu, yelong shen, Phillip Wallis, Zeyuan Allen-Zhu, Yuanzhi Li, Shean Wang, Lu~Wang, and Weizhu Chen.
\newblock Lo{RA}: Low-rank adaptation of large language models.
\newblock In \emph{Proceedings of the Tenth International Conference on Learning Representations}, 2022.
\newblock URL \url{https://openreview.net/forum?id=nZeVKeeFYf9}.

\bibitem[Huang et~al.(2024)Huang, Li, Hsu, Chen, Lin, Hsiao, Tsai, and Lee]{huang-etal-2024-chat}
Shih-Cheng Huang, Pin-Zu Li, Yu-chi Hsu, Kuang-Ming Chen, Yu~Tung Lin, Shih-Kai Hsiao, Richard Tsai, and Hung-yi Lee.
\newblock Chat vector: A simple approach to equip {LLM}s with instruction following and model alignment in new languages.
\newblock In Lun-Wei Ku, Andre Martins, and Vivek Srikumar (eds.), \emph{Proceedings of the 62nd Annual Meeting of the Association for Computational Linguistics (Volume 1: Long Papers)}, pp.\  10943--10959, Bangkok, Thailand, August 2024. Association for Computational Linguistics.
\newblock \doi{10.18653/v1/2024.acl-long.590}.
\newblock URL \url{https://aclanthology.org/2024.acl-long.590/}.

\bibitem[Kim et~al.(2024)Kim, Choi, and Jeong]{kim2024efficienteffectivevocabularyexpansion}
Seungduk Kim, Seungtaek Choi, and Myeongho Jeong.
\newblock Efficient and effective vocabulary expansion towards multilingual large language models.
\newblock \emph{arXiv}, abs/2402.14714, 2024.
\newblock URL \url{https://arxiv.org/abs/2402.14714}.

\bibitem[Kudugunta et~al.(2023)Kudugunta, Caswell, Zhang, Garcia, Xin, Kusupati, Stella, Bapna, and Firat]{kudugunta2023madlad}
Sneha Kudugunta, Isaac~Rayburn Caswell, Biao Zhang, Xavier Garcia, Derrick Xin, Aditya Kusupati, Romi Stella, Ankur Bapna, and Orhan Firat.
\newblock {MADLAD}-400: A multilingual and document-level large audited dataset.
\newblock In \emph{Proceedings of the Thirty-seventh Conference on Neural Information Processing Systems Datasets and Benchmarks Track}, 2023.
\newblock URL \url{https://openreview.net/forum?id=Y45ZCxslFx}.

\bibitem[Larcher et~al.(2023)Larcher, Piau, Finardi, Gengo, Esposito, and Caridá]{larcher2023cabritaclosinggapforeign}
Celio Larcher, Marcos Piau, Paulo Finardi, Pedro Gengo, Piero Esposito, and Vinicius Caridá.
\newblock Cabrita: closing the gap for foreign languages.
\newblock \emph{arXiv}, abs/2308.11878, 2023.
\newblock URL \url{https://arxiv.org/abs/2308.11878}.

\bibitem[Lhoest et~al.(2021)Lhoest, Villanova~del Moral, Jernite, Thakur, von Platen, Patil, Chaumond, Drame, Plu, Tunstall, Davison, {\v{S}}a{\v{s}}ko, Chhablani, Malik, Brandeis, Le~Scao, Sanh, Xu, Patry, McMillan-Major, Schmid, Gugger, Delangue, Matussi{\`e}re, Debut, Bekman, Cistac, Goehringer, Mustar, Lagunas, Rush, and Wolf]{lhoest-etal-2021-datasets}
Quentin Lhoest, Albert Villanova~del Moral, Yacine Jernite, Abhishek Thakur, Patrick von Platen, Suraj Patil, Julien Chaumond, Mariama Drame, Julien Plu, Lewis Tunstall, Joe Davison, Mario {\v{S}}a{\v{s}}ko, Gunjan Chhablani, Bhavitvya Malik, Simon Brandeis, Teven Le~Scao, Victor Sanh, Canwen Xu, Nicolas Patry, Angelina McMillan-Major, Philipp Schmid, Sylvain Gugger, Cl{\'e}ment Delangue, Th{\'e}o Matussi{\`e}re, Lysandre Debut, Stas Bekman, Pierric Cistac, Thibault Goehringer, Victor Mustar, Fran{\c{c}}ois Lagunas, Alexander Rush, and Thomas Wolf.
\newblock Datasets: A community library for natural language processing.
\newblock In Heike Adel and Shuming Shi (eds.), \emph{Proceedings of the 2021 Conference on Empirical Methods in Natural Language Processing: System Demonstrations}, pp.\  175--184, Online and Punta Cana, Dominican Republic, November 2021. Association for Computational Linguistics.
\newblock \doi{10.18653/v1/2021.emnlp-demo.21}.
\newblock URL \url{https://aclanthology.org/2021.emnlp-demo.21/}.

\bibitem[Li et~al.(2023)Li, Zhang, Dubois, Taori, Gulrajani, Guestrin, Liang, and Hashimoto]{alpaca_eval}
Xuechen Li, Tianyi Zhang, Yann Dubois, Rohan Taori, Ishaan Gulrajani, Carlos Guestrin, Percy Liang, and Tatsunori~B. Hashimoto.
\newblock {AlpacaEval}: An automatic evaluator of instruction-following models.
\newblock \url{https://github.com/tatsu-lab/alpaca_eval}, 2023.

\bibitem[Lin(2004)]{lin-2004-rouge}
Chin-Yew Lin.
\newblock {ROUGE}: A package for automatic evaluation of summaries.
\newblock In \emph{Text Summarization Branches Out}, pp.\  74--81, Barcelona, Spain, July 2004. Association for Computational Linguistics.
\newblock URL \url{https://aclanthology.org/W04-1013/}.

\bibitem[Lin et~al.(2024)Lin, Ji, Tiedemann, Martins, and Schütze]{lin2024mala500massivelanguageadaptation}
Peiqin Lin, Shaoxiong Ji, Jörg Tiedemann, André F.~T. Martins, and Hinrich Schütze.
\newblock {MaLA-500}: Massive language adaptation of large language models.
\newblock \emph{arXiv}, abs/2401.13303, 2024.
\newblock URL \url{https://arxiv.org/abs/2401.13303}.

\bibitem[Lin et~al.(2022)Lin, Hilton, and Evans]{lin-etal-2022-truthfulqa}
Stephanie Lin, Jacob Hilton, and Owain Evans.
\newblock {T}ruthful{QA}: Measuring how models mimic human falsehoods.
\newblock In Smaranda Muresan, Preslav Nakov, and Aline Villavicencio (eds.), \emph{Proceedings of the 60th Annual Meeting of the Association for Computational Linguistics (Volume 1: Long Papers)}, pp.\  3214--3252, Dublin, Ireland, May 2022. Association for Computational Linguistics.
\newblock \doi{10.18653/v1/2022.acl-long.229}.
\newblock URL \url{https://aclanthology.org/2022.acl-long.229/}.

\bibitem[Mahdizadeh~Sani et~al.(2025)Mahdizadeh~Sani, Sadeghi, Vu, Yaghoobzadeh, and Haffari]{mahdizadeh-sani-etal-2025-extending}
Samin Mahdizadeh~Sani, Pouya Sadeghi, Thuy-Trang Vu, Yadollah Yaghoobzadeh, and Gholamreza Haffari.
\newblock Extending {LLM}s to new languages: A case study of llama and {P}ersian adaptation.
\newblock In Owen Rambow, Leo Wanner, Marianna Apidianaki, Hend Al-Khalifa, Barbara~Di Eugenio, and Steven Schockaert (eds.), \emph{Proceedings of the 31st International Conference on Computational Linguistics}, pp.\  8868--8884, Abu Dhabi, UAE, January 2025. Association for Computational Linguistics.
\newblock URL \url{https://aclanthology.org/2025.coling-main.594/}.

\bibitem[Maynez et~al.(2023)Maynez, Agrawal, and Gehrmann]{maynez-etal-2023-benchmarking}
Joshua Maynez, Priyanka Agrawal, and Sebastian Gehrmann.
\newblock Benchmarking large language model capabilities for conditional generation.
\newblock In Anna Rogers, Jordan Boyd-Graber, and Naoaki Okazaki (eds.), \emph{Proceedings of the 61st Annual Meeting of the Association for Computational Linguistics (Volume 1: Long Papers)}, pp.\  9194--9213, Toronto, Canada, July 2023. Association for Computational Linguistics.
\newblock \doi{10.18653/v1/2023.acl-long.511}.
\newblock URL \url{https://aclanthology.org/2023.acl-long.511/}.

\bibitem[Minixhofer et~al.(2022)Minixhofer, Paischer, and Rekabsaz]{minixhofer-etal-2022-wechsel}
Benjamin Minixhofer, Fabian Paischer, and Navid Rekabsaz.
\newblock {WECHSEL}: Effective initialization of subword embeddings for cross-lingual transfer of monolingual language models.
\newblock In Marine Carpuat, Marie-Catherine de~Marneffe, and Ivan~Vladimir Meza~Ruiz (eds.), \emph{Proceedings of the 2022 Conference of the North American Chapter of the Association for Computational Linguistics: Human Language Technologies}, pp.\  3992--4006, Seattle, United States, July 2022. Association for Computational Linguistics.
\newblock \doi{10.18653/v1/2022.naacl-main.293}.
\newblock URL \url{https://aclanthology.org/2022.naacl-main.293/}.

\bibitem[Minixhofer et~al.(2024)Minixhofer, Ponti, and Vuli{\'c}]{minixhofer2024zeroshot}
Benjamin Minixhofer, Edoardo Ponti, and Ivan Vuli{\'c}.
\newblock Zero-shot tokenizer transfer.
\newblock In \emph{Proceedings of the Thirty-eighth Annual Conference on Neural Information Processing Systems}, 2024.
\newblock URL \url{https://openreview.net/forum?id=RwBObRsIzC}.

\bibitem[Mundra et~al.(2024)Mundra, Khandavally, Dabre, Puduppully, Kunchukuttan, and Khapra]{mundra-etal-2024-empirical}
Nandini Mundra, Aditya Nanda~Kishore Khandavally, Raj Dabre, Ratish Puduppully, Anoop Kunchukuttan, and Mitesh~M Khapra.
\newblock An empirical comparison of vocabulary expansion and initialization approaches for language models.
\newblock In Libby Barak and Malihe Alikhani (eds.), \emph{Proceedings of the 28th Conference on Computational Natural Language Learning}, pp.\  84--104, Miami, FL, USA, November 2024. Association for Computational Linguistics.
\newblock \doi{10.18653/v1/2024.conll-1.8}.
\newblock URL \url{https://aclanthology.org/2024.conll-1.8/}.

\bibitem[Nguyen et~al.(2024)Nguyen, Zhang, Li, Aljunied, Hu, Shen, Chia, Li, Wang, Tan, Cheng, Chen, Deng, Yang, Liu, Zhang, and Bing]{nguyen-etal-2024-seallms}
Xuan-Phi Nguyen, Wenxuan Zhang, Xin Li, Mahani Aljunied, Zhiqiang Hu, Chenhui Shen, Yew~Ken Chia, Xingxuan Li, Jianyu Wang, Qingyu Tan, Liying Cheng, Guanzheng Chen, Yue Deng, Sen Yang, Chaoqun Liu, Hang Zhang, and Lidong Bing.
\newblock {S}ea{LLM}s - large language models for {S}outheast {A}sia.
\newblock In Yixin Cao, Yang Feng, and Deyi Xiong (eds.), \emph{Proceedings of the 62nd Annual Meeting of the Association for Computational Linguistics (Volume 3: System Demonstrations)}, pp.\  294--304, Bangkok, Thailand, August 2024. Association for Computational Linguistics.
\newblock \doi{10.18653/v1/2024.acl-demos.28}.
\newblock URL \url{https://aclanthology.org/2024.acl-demos.28/}.

\bibitem[{NLLB Team} et~al.(2022){NLLB Team}, Costa-jussà, Cross, Çelebi, Elbayad, Heafield, Heffernan, Kalbassi, Lam, Licht, Maillard, Sun, Wang, Wenzek, Youngblood, Akula, Barrault, Mejia-Gonzalez, Hansanti, Hoffman, Jarrett, Sadagopan, Rowe, Spruit, Tran, Andrews, Ayan, Bhosale, Edunov, Fan, Gao, Goswami, Guzmán, Koehn, Mourachko, Ropers, Saleem, Schwenk, and Wang]{nllb-22}
{NLLB Team}, Marta~R. Costa-jussà, James Cross, Onur Çelebi, Maha Elbayad, Kenneth Heafield, Kevin Heffernan, Elahe Kalbassi, Janice Lam, Daniel Licht, Jean Maillard, Anna Sun, Skyler Wang, Guillaume Wenzek, Al~Youngblood, Bapi Akula, Loic Barrault, Gabriel Mejia-Gonzalez, Prangthip Hansanti, John Hoffman, Semarley Jarrett, Kaushik~Ram Sadagopan, Dirk Rowe, Shannon Spruit, Chau Tran, Pierre Andrews, Necip~Fazil Ayan, Shruti Bhosale, Sergey Edunov, Angela Fan, Cynthia Gao, Vedanuj Goswami, Francisco Guzmán, Philipp Koehn, Alexandre Mourachko, Christophe Ropers, Safiyyah Saleem, Holger Schwenk, and Jeff Wang.
\newblock No language left behind: Scaling human-centered machine translation.
\newblock \emph{arXiv}, abs/2207.04672, 2022.
\newblock URL \url{https://arxiv.org/abs/2207.04672}.

\bibitem[Ostendorff \& Rehm(2023)Ostendorff and Rehm]{ostendorff2023efficientlanguagemodeltraining}
Malte Ostendorff and Georg Rehm.
\newblock Efficient language model training through cross-lingual and progressive transfer learning.
\newblock \emph{arXiv}, abs/2301.09626, 2023.
\newblock URL \url{https://arxiv.org/abs/2301.09626}.

\bibitem[Petrov et~al.(2023)Petrov, Malfa, Torr, and Bibi]{petrov2023language}
Aleksandar Petrov, Emanuele~La Malfa, Philip Torr, and Adel Bibi.
\newblock Language model tokenizers introduce unfairness between languages.
\newblock In \emph{Proceedings of the Thirty-seventh Conference on Neural Information Processing Systems}, 2023.
\newblock URL \url{https://openreview.net/forum?id=78yDLKi95p}.

\bibitem[Pipatanakul et~al.(2023)Pipatanakul, Jirabovonvisut, Manakul, Sripaisarnmongkol, Patomwong, Chokchainant, and Tharnpipitchai]{pipatanakul2023typhoonthailargelanguage}
Kunat Pipatanakul, Phatrasek Jirabovonvisut, Potsawee Manakul, Sittipong Sripaisarnmongkol, Ruangsak Patomwong, Pathomporn Chokchainant, and Kasima Tharnpipitchai.
\newblock Typhoon: {Thai} large language models.
\newblock \emph{arXiv}, abs/2312.13951, 2023.
\newblock URL \url{https://arxiv.org/abs/2312.13951}.

\bibitem[Popovi{\'c}(2015)]{popovic-2015-chrf}
Maja Popovi{\'c}.
\newblock chr{F}: character n-gram {F}-score for automatic {MT} evaluation.
\newblock In Ond{\v{r}}ej Bojar, Rajan Chatterjee, Christian Federmann, Barry Haddow, Chris Hokamp, Matthias Huck, Varvara Logacheva, and Pavel Pecina (eds.), \emph{Proceedings of the Tenth Workshop on Statistical Machine Translation}, pp.\  392--395, Lisbon, Portugal, September 2015. Association for Computational Linguistics.
\newblock \doi{10.18653/v1/W15-3049}.
\newblock URL \url{https://aclanthology.org/W15-3049/}.

\bibitem[Popovi{\'c}(2017)]{popovic-2017-chrf}
Maja Popovi{\'c}.
\newblock chr{F}++: words helping character n-grams.
\newblock In Ond{\v{r}}ej Bojar, Christian Buck, Rajen Chatterjee, Christian Federmann, Yvette Graham, Barry Haddow, Matthias Huck, Antonio~Jimeno Yepes, Philipp Koehn, and Julia Kreutzer (eds.), \emph{Proceedings of the Second Conference on Machine Translation}, pp.\  612--618, Copenhagen, Denmark, September 2017. Association for Computational Linguistics.
\newblock \doi{10.18653/v1/W17-4770}.
\newblock URL \url{https://aclanthology.org/W17-4770/}.

\bibitem[Post(2018)]{post-2018-call}
Matt Post.
\newblock A call for clarity in reporting {BLEU} scores.
\newblock In Ond{\v{r}}ej Bojar, Rajen Chatterjee, Christian Federmann, Mark Fishel, Yvette Graham, Barry Haddow, Matthias Huck, Antonio~Jimeno Yepes, Philipp Koehn, Christof Monz, Matteo Negri, Aur{\'e}lie N{\'e}v{\'e}ol, Mariana Neves, Matt Post, Lucia Specia, Marco Turchi, and Karin Verspoor (eds.), \emph{Proceedings of the Third Conference on Machine Translation: Research Papers}, pp.\  186--191, Brussels, Belgium, October 2018. Association for Computational Linguistics.
\newblock \doi{10.18653/v1/W18-6319}.
\newblock URL \url{https://aclanthology.org/W18-6319/}.

\bibitem[Qwen et~al.(2025)Qwen, :, Yang, Yang, Zhang, Hui, Zheng, Yu, Li, Liu, Huang, Wei, Lin, Yang, Tu, Zhang, Yang, Yang, Zhou, Lin, Dang, Lu, Bao, Yang, Yu, Li, Xue, Zhang, Zhu, Men, Lin, Li, Tang, Xia, Ren, Ren, Fan, Su, Zhang, Wan, Liu, Cui, Zhang, and Qiu]{qwen2.5}
Qwen, :, An~Yang, Baosong Yang, Beichen Zhang, Binyuan Hui, Bo~Zheng, Bowen Yu, Chengyuan Li, Dayiheng Liu, Fei Huang, Haoran Wei, Huan Lin, Jian Yang, Jianhong Tu, Jianwei Zhang, Jianxin Yang, Jiaxi Yang, Jingren Zhou, Junyang Lin, Kai Dang, Keming Lu, Keqin Bao, Kexin Yang, Le~Yu, Mei Li, Mingfeng Xue, Pei Zhang, Qin Zhu, Rui Men, Runji Lin, Tianhao Li, Tianyi Tang, Tingyu Xia, Xingzhang Ren, Xuancheng Ren, Yang Fan, Yang Su, Yichang Zhang, Yu~Wan, Yuqiong Liu, Zeyu Cui, Zhenru Zhang, and Zihan Qiu.
\newblock Qwen2.5 technical report.
\newblock \emph{arXiv}, abs/2412.15115, 2025.
\newblock URL \url{https://arxiv.org/abs/2412.15115}.

\bibitem[Remy et~al.(2024)Remy, Delobelle, Avetisyan, Khabibullina, de~Lhoneux, and Demeester]{remy2024transtokenization}
Fran{\c{c}}ois Remy, Pieter Delobelle, Hayastan Avetisyan, Alfiya Khabibullina, Miryam de~Lhoneux, and Thomas Demeester.
\newblock Trans-tokenization and cross-lingual vocabulary transfers: Language adaptation of {LLM}s for low-resource {NLP}.
\newblock In \emph{Proceedings of the First Conference on Language Modeling}, 2024.
\newblock URL \url{https://openreview.net/forum?id=sBxvoDhvao}.

\bibitem[Shi et~al.(2023)Shi, Suzgun, Freitag, Wang, Srivats, Vosoughi, Chung, Tay, Ruder, Zhou, Das, and Wei]{shi2023language}
Freda Shi, Mirac Suzgun, Markus Freitag, Xuezhi Wang, Suraj Srivats, Soroush Vosoughi, Hyung~Won Chung, Yi~Tay, Sebastian Ruder, Denny Zhou, Dipanjan Das, and Jason Wei.
\newblock Language models are multilingual chain-of-thought reasoners.
\newblock In \emph{Proceedings of the Eleventh International Conference on Learning Representations}, 2023.
\newblock URL \url{https://openreview.net/forum?id=fR3wGCk-IXp}.

\bibitem[Singh et~al.(2024)Singh, Vargus, D{'}souza, Karlsson, Mahendiran, Ko, Shandilya, Patel, Mataciunas, O{'}Mahony, Zhang, Hettiarachchi, Wilson, Machado, Moura, Krzemi{\'n}ski, Fadaei, Ergun, Okoh, Alaagib, Mudannayake, Alyafeai, Chien, Ruder, Guthikonda, Alghamdi, Gehrmann, Muennighoff, Bartolo, Kreutzer, {\"U}st{\"u}n, Fadaee, and Hooker]{singh-etal-2024-aya}
Shivalika Singh, Freddie Vargus, Daniel D{'}souza, B{\"o}rje Karlsson, Abinaya Mahendiran, Wei-Yin Ko, Herumb Shandilya, Jay Patel, Deividas Mataciunas, Laura O{'}Mahony, Mike Zhang, Ramith Hettiarachchi, Joseph Wilson, Marina Machado, Luisa Moura, Dominik Krzemi{\'n}ski, Hakimeh Fadaei, Irem Ergun, Ifeoma Okoh, Aisha Alaagib, Oshan Mudannayake, Zaid Alyafeai, Vu~Chien, Sebastian Ruder, Surya Guthikonda, Emad Alghamdi, Sebastian Gehrmann, Niklas Muennighoff, Max Bartolo, Julia Kreutzer, Ahmet {\"U}st{\"u}n, Marzieh Fadaee, and Sara Hooker.
\newblock Aya dataset: An open-access collection for multilingual instruction tuning.
\newblock In Lun-Wei Ku, Andre Martins, and Vivek Srikumar (eds.), \emph{Proceedings of the 62nd Annual Meeting of the Association for Computational Linguistics (Volume 1: Long Papers)}, pp.\  11521--11567, Bangkok, Thailand, August 2024. Association for Computational Linguistics.
\newblock \doi{10.18653/v1/2024.acl-long.620}.
\newblock URL \url{https://aclanthology.org/2024.acl-long.620/}.

\bibitem[Singh et~al.(2025)Singh, Romanou, Fourrier, Adelani, Ngui, Vila-Suero, Limkonchotiwat, Marchisio, Leong, Susanto, Ng, Longpre, Ruder, Ko, Bosselut, Oh, Martins, Choshen, Ippolito, Ferrante, Fadaee, Ermis, and Hooker]{singh2024globalmmluunderstandingaddressing}
Shivalika Singh, Angelika Romanou, Cl{\'e}mentine Fourrier, David~Ifeoluwa Adelani, Jian~Gang Ngui, Daniel Vila-Suero, Peerat Limkonchotiwat, Kelly Marchisio, Wei~Qi Leong, Yosephine Susanto, Raymond Ng, Shayne Longpre, Sebastian Ruder, Wei-Yin Ko, Antoine Bosselut, Alice Oh, Andre Martins, Leshem Choshen, Daphne Ippolito, Enzo Ferrante, Marzieh Fadaee, Beyza Ermis, and Sara Hooker.
\newblock Global {MMLU}: Understanding and addressing cultural and linguistic biases in multilingual evaluation.
\newblock In Wanxiang Che, Joyce Nabende, Ekaterina Shutova, and Mohammad~Taher Pilehvar (eds.), \emph{Proceedings of the 63rd Annual Meeting of the Association for Computational Linguistics (Volume 1: Long Papers)}, pp.\  18761--18799, Vienna, Austria, July 2025. Association for Computational Linguistics.
\newblock ISBN 979-8-89176-251-0.
\newblock \doi{10.18653/v1/2025.acl-long.919}.
\newblock URL \url{https://aclanthology.org/2025.acl-long.919/}.

\bibitem[Srivastava et~al.(2023)Srivastava, Rastogi, Rao, Shoeb, Abid, Fisch, Brown, Santoro, Gupta, Garriga-Alonso, Kluska, Lewkowycz, Agarwal, Power, Ray, Warstadt, Kocurek, Safaya, Tazarv, Xiang, Parrish, Nie, Hussain, Askell, Dsouza, Slone, Rahane, Iyer, Andreassen, Madotto, Santilli, Stuhlm{\"u}ller, Dai, La, Lampinen, Zou, Jiang, Chen, Vuong, Gupta, Gottardi, Norelli, Venkatesh, Gholamidavoodi, Tabassum, Menezes, Kirubarajan, Mullokandov, Sabharwal, Herrick, Efrat, Erdem, Karaka{\c{s}}, Roberts, Loe, Zoph, Bojanowski, {\"O}zyurt, Hedayatnia, Neyshabur, Inden, Stein, Ekmekci, Lin, Howald, Orinion, Diao, Dour, Stinson, Argueta, Ferri, Singh, Rathkopf, Meng, Baral, Wu, Callison-Burch, Waites, Voigt, Manning, Potts, Ramirez, Rivera, Siro, Raffel, Ashcraft, Garbacea, Sileo, Garrette, Hendrycks, Kilman, Roth, Freeman, Khashabi, Levy, Gonz{\'a}lez, Perszyk, Hernandez, Chen, Ippolito, Gilboa, Dohan, Drakard, Jurgens, Datta, Ganguli, Emelin, Kleyko, Yuret, Chen, Tam, Hupkes, Misra, Buzan, Mollo, Yang, Lee,
  Schrader, Shutova, Cubuk, Segal, Hagerman, Barnes, Donoway, Pavlick, Rodol{\`a}, Lam, Chu, Tang, Erdem, Chang, Chi, Dyer, Jerzak, Kim, Manyasi, Zheltonozhskii, Xia, Siar, Mart{\'\i}nez-Plumed, Happ{\'e}, Chollet, Rong, Mishra, Winata, de~Melo, Kruszewski, Parascandolo, Mariani, Wang, Jaimovitch-Lopez, Betz, Gur-Ari, Galijasevic, Kim, Rashkin, Hajishirzi, Mehta, Bogar, Shevlin, Schuetze, Yakura, Zhang, Wong, Ng, Noble, Jumelet, Geissinger, Kernion, Hilton, Lee, Fisac, Simon, Koppel, Zheng, Zou, Kocon, Thompson, Wingfield, Kaplan, Radom, Sohl-Dickstein, Phang, Wei, Yosinski, Novikova, Bosscher, Marsh, Kim, Taal, Engel, Alabi, Xu, Song, Tang, Waweru, Burden, Miller, Balis, Batchelder, Berant, Frohberg, Rozen, Hernandez-Orallo, Boudeman, Guerr, Jones, Tenenbaum, Rule, Chua, Kanclerz, Livescu, Krauth, Gopalakrishnan, Ignatyeva, Markert, Dhole, Gimpel, Omondi, Mathewson, Chiafullo, Shkaruta, Shridhar, McDonell, Richardson, Reynolds, Gao, Zhang, Dugan, Qin, Contreras-Ochando, Morency, Moschella, Lam, Noble,
  Schmidt, He, Oliveros-Col{\'o}n, Metz, Senel, Bosma, Sap, Hoeve, Farooqi, Faruqui, Mazeika, Baturan, Marelli, Maru, Ramirez-Quintana, Tolkiehn, Giulianelli, Lewis, Potthast, Leavitt, Hagen, Schubert, Baitemirova, Arnaud, McElrath, Yee, Cohen, Gu, Ivanitskiy, Starritt, Strube, Sw{\k{e}}drowski, Bevilacqua, Yasunaga, Kale, Cain, Xu, Suzgun, Walker, Tiwari, Bansal, Aminnaseri, Geva, Gheini, T, Peng, Chi, Lee, Krakover, Cameron, Roberts, Doiron, Martinez, Nangia, Deckers, Muennighoff, Keskar, Iyer, Constant, Fiedel, Wen, Zhang, Agha, Elbaghdadi, Levy, Evans, Casares, Doshi, Fung, Liang, Vicol, Alipoormolabashi, Liao, Liang, Chang, Eckersley, Htut, Hwang, Mi{\l}kowski, Patil, Pezeshkpour, Oli, Mei, Lyu, Chen, Banjade, Rudolph, Gabriel, Habacker, Risco, Milli{\`e}re, Garg, Barnes, Saurous, Arakawa, Raymaekers, Frank, Sikand, Novak, Sitelew, Bras, Liu, Jacobs, Zhang, Salakhutdinov, Chi, Lee, Stovall, Teehan, Yang, Singh, Mohammad, Anand, Dillavou, Shleifer, Wiseman, Gruetter, Bowman, Schoenholz, Han, Kwatra, Rous,
  Ghazarian, Ghosh, Casey, Bischoff, Gehrmann, Schuster, Sadeghi, Hamdan, Zhou, Srivastava, Shi, Singh, Asaadi, Gu, Pachchigar, Toshniwal, Upadhyay, Debnath, Shakeri, Thormeyer, Melzi, Reddy, Makini, Lee, Torene, Hatwar, Dehaene, Divic, Ermon, Biderman, Lin, Prasad, Piantadosi, Shieber, Misherghi, Kiritchenko, Mishra, Linzen, Schuster, Li, Yu, Ali, Hashimoto, Wu, Desbordes, Rothschild, Phan, Wang, Nkinyili, Schick, Kornev, Tunduny, Gerstenberg, Chang, Neeraj, Khot, Shultz, Shaham, Misra, Demberg, Nyamai, Raunak, Ramasesh, vinay~uday prabhu, Padmakumar, Srikumar, Fedus, Saunders, Zhang, Vossen, Ren, Tong, Zhao, Wu, Shen, Yaghoobzadeh, Lakretz, Song, Bahri, Choi, Yang, Hao, Chen, Belinkov, Hou, Hou, Bai, Seid, Zhao, Wang, Wang, Wang, and Wu]{srivastava2023beyond}
Aarohi Srivastava, Abhinav Rastogi, Abhishek Rao, Abu Awal~Md Shoeb, Abubakar Abid, Adam Fisch, Adam~R. Brown, Adam Santoro, Aditya Gupta, Adri{\`a} Garriga-Alonso, Agnieszka Kluska, Aitor Lewkowycz, Akshat Agarwal, Alethea Power, Alex Ray, Alex Warstadt, Alexander~W. Kocurek, Ali Safaya, Ali Tazarv, Alice Xiang, Alicia Parrish, Allen Nie, Aman Hussain, Amanda Askell, Amanda Dsouza, Ambrose Slone, Ameet Rahane, Anantharaman~S. Iyer, Anders~Johan Andreassen, Andrea Madotto, Andrea Santilli, Andreas Stuhlm{\"u}ller, Andrew~M. Dai, Andrew La, Andrew~Kyle Lampinen, Andy Zou, Angela Jiang, Angelica Chen, Anh Vuong, Animesh Gupta, Anna Gottardi, Antonio Norelli, Anu Venkatesh, Arash Gholamidavoodi, Arfa Tabassum, Arul Menezes, Arun Kirubarajan, Asher Mullokandov, Ashish Sabharwal, Austin Herrick, Avia Efrat, Aykut Erdem, Ayla Karaka{\c{s}}, B.~Ryan Roberts, Bao~Sheng Loe, Barret Zoph, Bart{\l}omiej Bojanowski, Batuhan {\"O}zyurt, Behnam Hedayatnia, Behnam Neyshabur, Benjamin Inden, Benno Stein, Berk Ekmekci,
  Bill~Yuchen Lin, Blake Howald, Bryan Orinion, Cameron Diao, Cameron Dour, Catherine Stinson, Cedrick Argueta, Cesar Ferri, Chandan Singh, Charles Rathkopf, Chenlin Meng, Chitta Baral, Chiyu Wu, Chris Callison-Burch, Christopher Waites, Christian Voigt, Christopher~D Manning, Christopher Potts, Cindy Ramirez, Clara~E. Rivera, Clemencia Siro, Colin Raffel, Courtney Ashcraft, Cristina Garbacea, Damien Sileo, Dan Garrette, Dan Hendrycks, Dan Kilman, Dan Roth, C.~Daniel Freeman, Daniel Khashabi, Daniel Levy, Daniel~Mosegu{\'\i} Gonz{\'a}lez, Danielle Perszyk, Danny Hernandez, Danqi Chen, Daphne Ippolito, Dar Gilboa, David Dohan, David Drakard, David Jurgens, Debajyoti Datta, Deep Ganguli, Denis Emelin, Denis Kleyko, Deniz Yuret, Derek Chen, Derek Tam, Dieuwke Hupkes, Diganta Misra, Dilyar Buzan, Dimitri~Coelho Mollo, Diyi Yang, Dong-Ho Lee, Dylan Schrader, Ekaterina Shutova, Ekin~Dogus Cubuk, Elad Segal, Eleanor Hagerman, Elizabeth Barnes, Elizabeth Donoway, Ellie Pavlick, Emanuele Rodol{\`a}, Emma Lam, Eric
  Chu, Eric Tang, Erkut Erdem, Ernie Chang, Ethan~A Chi, Ethan Dyer, Ethan Jerzak, Ethan Kim, Eunice~Engefu Manyasi, Evgenii Zheltonozhskii, Fanyue Xia, Fatemeh Siar, Fernando Mart{\'\i}nez-Plumed, Francesca Happ{\'e}, Francois Chollet, Frieda Rong, Gaurav Mishra, Genta~Indra Winata, Gerard de~Melo, Germ{\`a}n Kruszewski, Giambattista Parascandolo, Giorgio Mariani, Gloria~Xinyue Wang, Gonzalo Jaimovitch-Lopez, Gregor Betz, Guy Gur-Ari, Hana Galijasevic, Hannah Kim, Hannah Rashkin, Hannaneh Hajishirzi, Harsh Mehta, Hayden Bogar, Henry Francis~Anthony Shevlin, Hinrich Schuetze, Hiromu Yakura, Hongming Zhang, Hugh~Mee Wong, Ian Ng, Isaac Noble, Jaap Jumelet, Jack Geissinger, Jackson Kernion, Jacob Hilton, Jaehoon Lee, Jaime~Fern{\'a}ndez Fisac, James~B Simon, James Koppel, James Zheng, James Zou, Jan Kocon, Jana Thompson, Janelle Wingfield, Jared Kaplan, Jarema Radom, Jascha Sohl-Dickstein, Jason Phang, Jason Wei, Jason Yosinski, Jekaterina Novikova, Jelle Bosscher, Jennifer Marsh, Jeremy Kim, Jeroen Taal, Jesse
  Engel, Jesujoba Alabi, Jiacheng Xu, Jiaming Song, Jillian Tang, Joan Waweru, John Burden, John Miller, John~U. Balis, Jonathan Batchelder, Jonathan Berant, J{\"o}rg Frohberg, Jos Rozen, Jose Hernandez-Orallo, Joseph Boudeman, Joseph Guerr, Joseph Jones, Joshua~B. Tenenbaum, Joshua~S. Rule, Joyce Chua, Kamil Kanclerz, Karen Livescu, Karl Krauth, Karthik Gopalakrishnan, Katerina Ignatyeva, Katja Markert, Kaustubh Dhole, Kevin Gimpel, Kevin Omondi, Kory~Wallace Mathewson, Kristen Chiafullo, Ksenia Shkaruta, Kumar Shridhar, Kyle McDonell, Kyle Richardson, Laria Reynolds, Leo Gao, Li~Zhang, Liam Dugan, Lianhui Qin, Lidia Contreras-Ochando, Louis-Philippe Morency, Luca Moschella, Lucas Lam, Lucy Noble, Ludwig Schmidt, Luheng He, Luis Oliveros-Col{\'o}n, Luke Metz, L{\"u}tfi~Kerem Senel, Maarten Bosma, Maarten Sap, Maartje~Ter Hoeve, Maheen Farooqi, Manaal Faruqui, Mantas Mazeika, Marco Baturan, Marco Marelli, Marco Maru, Maria~Jose Ramirez-Quintana, Marie Tolkiehn, Mario Giulianelli, Martha Lewis, Martin
  Potthast, Matthew~L Leavitt, Matthias Hagen, M{\'a}ty{\'a}s Schubert, Medina~Orduna Baitemirova, Melody Arnaud, Melvin McElrath, Michael~Andrew Yee, Michael Cohen, Michael Gu, Michael Ivanitskiy, Michael Starritt, Michael Strube, Micha{\l} Sw{\k{e}}drowski, Michele Bevilacqua, Michihiro Yasunaga, Mihir Kale, Mike Cain, Mimee Xu, Mirac Suzgun, Mitch Walker, Mo~Tiwari, Mohit Bansal, Moin Aminnaseri, Mor Geva, Mozhdeh Gheini, Mukund~Varma T, Nanyun Peng, Nathan~Andrew Chi, Nayeon Lee, Neta Gur-Ari Krakover, Nicholas Cameron, Nicholas Roberts, Nick Doiron, Nicole Martinez, Nikita Nangia, Niklas Deckers, Niklas Muennighoff, Nitish~Shirish Keskar, Niveditha~S. Iyer, Noah Constant, Noah Fiedel, Nuan Wen, Oliver Zhang, Omar Agha, Omar Elbaghdadi, Omer Levy, Owain Evans, Pablo Antonio~Moreno Casares, Parth Doshi, Pascale Fung, Paul~Pu Liang, Paul Vicol, Pegah Alipoormolabashi, Peiyuan Liao, Percy Liang, Peter~W Chang, Peter Eckersley, Phu~Mon Htut, Pinyu Hwang, Piotr Mi{\l}kowski, Piyush Patil, Pouya Pezeshkpour,
  Priti Oli, Qiaozhu Mei, Qing Lyu, Qinlang Chen, Rabin Banjade, Rachel~Etta Rudolph, Raefer Gabriel, Rahel Habacker, Ramon Risco, Rapha{\"e}l Milli{\`e}re, Rhythm Garg, Richard Barnes, Rif~A. Saurous, Riku Arakawa, Robbe Raymaekers, Robert Frank, Rohan Sikand, Roman Novak, Roman Sitelew, Ronan~Le Bras, Rosanne Liu, Rowan Jacobs, Rui Zhang, Russ Salakhutdinov, Ryan~Andrew Chi, Seungjae~Ryan Lee, Ryan Stovall, Ryan Teehan, Rylan Yang, Sahib Singh, Saif~M. Mohammad, Sajant Anand, Sam Dillavou, Sam Shleifer, Sam Wiseman, Samuel Gruetter, Samuel~R. Bowman, Samuel~Stern Schoenholz, Sanghyun Han, Sanjeev Kwatra, Sarah~A. Rous, Sarik Ghazarian, Sayan Ghosh, Sean Casey, Sebastian Bischoff, Sebastian Gehrmann, Sebastian Schuster, Sepideh Sadeghi, Shadi Hamdan, Sharon Zhou, Shashank Srivastava, Sherry Shi, Shikhar Singh, Shima Asaadi, Shixiang~Shane Gu, Shubh Pachchigar, Shubham Toshniwal, Shyam Upadhyay, Shyamolima~Shammie Debnath, Siamak Shakeri, Simon Thormeyer, Simone Melzi, Siva Reddy, Sneha~Priscilla Makini,
  Soo-Hwan Lee, Spencer Torene, Sriharsha Hatwar, Stanislas Dehaene, Stefan Divic, Stefano Ermon, Stella Biderman, Stephanie Lin, Stephen Prasad, Steven Piantadosi, Stuart Shieber, Summer Misherghi, Svetlana Kiritchenko, Swaroop Mishra, Tal Linzen, Tal Schuster, Tao Li, Tao Yu, Tariq Ali, Tatsunori Hashimoto, Te-Lin Wu, Th{\'e}o Desbordes, Theodore Rothschild, Thomas Phan, Tianle Wang, Tiberius Nkinyili, Timo Schick, Timofei Kornev, Titus Tunduny, Tobias Gerstenberg, Trenton Chang, Trishala Neeraj, Tushar Khot, Tyler Shultz, Uri Shaham, Vedant Misra, Vera Demberg, Victoria Nyamai, Vikas Raunak, Vinay~Venkatesh Ramasesh, vinay~uday prabhu, Vishakh Padmakumar, Vivek Srikumar, William Fedus, William Saunders, William Zhang, Wout Vossen, Xiang Ren, Xiaoyu Tong, Xinran Zhao, Xinyi Wu, Xudong Shen, Yadollah Yaghoobzadeh, Yair Lakretz, Yangqiu Song, Yasaman Bahri, Yejin Choi, Yichi Yang, Sophie Hao, Yifu Chen, Yonatan Belinkov, Yu~Hou, Yufang Hou, Yuntao Bai, Zachary Seid, Zhuoye Zhao, Zijian Wang, Zijie~J. Wang,
  Zirui Wang, and Ziyi Wu.
\newblock Beyond the imitation game: Quantifying and extrapolating the capabilities of language models.
\newblock \emph{Transactions on Machine Learning Research}, 2023.
\newblock ISSN 2835-8856.
\newblock URL \url{https://openreview.net/forum?id=uyTL5Bvosj}.
\newblock Featured Certification.

\bibitem[Suzgun et~al.(2023)Suzgun, Scales, Sch{\"a}rli, Gehrmann, Tay, Chung, Chowdhery, Le, Chi, Zhou, and Wei]{suzgun-etal-2023-challenging}
Mirac Suzgun, Nathan Scales, Nathanael Sch{\"a}rli, Sebastian Gehrmann, Yi~Tay, Hyung~Won Chung, Aakanksha Chowdhery, Quoc Le, Ed~Chi, Denny Zhou, and Jason Wei.
\newblock Challenging {BIG}-bench tasks and whether chain-of-thought can solve them.
\newblock In Anna Rogers, Jordan Boyd-Graber, and Naoaki Okazaki (eds.), \emph{Findings of the Association for Computational Linguistics: ACL 2023}, pp.\  13003--13051, Toronto, Canada, July 2023. Association for Computational Linguistics.
\newblock \doi{10.18653/v1/2023.findings-acl.824}.
\newblock URL \url{https://aclanthology.org/2023.findings-acl.824/}.

\bibitem[Tang et~al.(2024)Tang, Luo, Huang, Zhang, Wang, Zhao, Wei, and Wen]{tang-etal-2024-language}
Tianyi Tang, Wenyang Luo, Haoyang Huang, Dongdong Zhang, Xiaolei Wang, Xin Zhao, Furu Wei, and Ji-Rong Wen.
\newblock Language-specific neurons: The key to multilingual capabilities in large language models.
\newblock In Lun-Wei Ku, Andre Martins, and Vivek Srikumar (eds.), \emph{Proceedings of the 62nd Annual Meeting of the Association for Computational Linguistics (Volume 1: Long Papers)}, pp.\  5701--5715, Bangkok, Thailand, August 2024. Association for Computational Linguistics.
\newblock \doi{10.18653/v1/2024.acl-long.309}.
\newblock URL \url{https://aclanthology.org/2024.acl-long.309/}.

\bibitem[Tao et~al.(2024)Tao, Zhang, Huang, Ma, Huang, Zhao, and Feng]{tao-etal-2024-unlocking}
Mingxu Tao, Chen Zhang, Quzhe Huang, Tianyao Ma, Songfang Huang, Dongyan Zhao, and Yansong Feng.
\newblock Unlocking the potential of model merging for low-resource languages.
\newblock In Yaser Al-Onaizan, Mohit Bansal, and Yun-Nung Chen (eds.), \emph{Findings of the Association for Computational Linguistics: EMNLP 2024}, pp.\  8705--8720, Miami, Florida, USA, November 2024. Association for Computational Linguistics.
\newblock \doi{10.18653/v1/2024.findings-emnlp.508}.
\newblock URL \url{https://aclanthology.org/2024.findings-emnlp.508/}.

\bibitem[Tejaswi et~al.(2024)Tejaswi, Gupta, and Choi]{tejaswi-etal-2024-exploring}
Atula Tejaswi, Nilesh Gupta, and Eunsol Choi.
\newblock Exploring design choices for building language-specific {LLM}s.
\newblock In Yaser Al-Onaizan, Mohit Bansal, and Yun-Nung Chen (eds.), \emph{Findings of the Association for Computational Linguistics: EMNLP 2024}, pp.\  10485--10500, Miami, Florida, USA, November 2024. Association for Computational Linguistics.
\newblock \doi{10.18653/v1/2024.findings-emnlp.614}.
\newblock URL \url{https://aclanthology.org/2024.findings-emnlp.614/}.

\bibitem[Toraman(2024)]{toraman2024llamaturkadaptingopensourcegenerative}
Cagri Toraman.
\newblock {LlamaTurk}: Adapting open-source generative large language models for low-resource language.
\newblock \emph{arXiv}, abs/2405.07745, 2024.
\newblock URL \url{https://arxiv.org/abs/2405.07745}.

\bibitem[Wendler et~al.(2024)Wendler, Veselovsky, Monea, and West]{wendler-etal-2024-llamas}
Chris Wendler, Veniamin Veselovsky, Giovanni Monea, and Robert West.
\newblock Do llamas work in {E}nglish? on the latent language of multilingual transformers.
\newblock In Lun-Wei Ku, Andre Martins, and Vivek Srikumar (eds.), \emph{Proceedings of the 62nd Annual Meeting of the Association for Computational Linguistics (Volume 1: Long Papers)}, pp.\  15366--15394, Bangkok, Thailand, August 2024. Association for Computational Linguistics.
\newblock \doi{10.18653/v1/2024.acl-long.820}.
\newblock URL \url{https://aclanthology.org/2024.acl-long.820/}.

\bibitem[Wolf et~al.(2020)Wolf, Debut, Sanh, Chaumond, Delangue, Moi, Cistac, Rault, Louf, Funtowicz, Davison, Shleifer, von Platen, Ma, Jernite, Plu, Xu, Le~Scao, Gugger, Drame, Lhoest, and Rush]{wolf-etal-2020-transformers}
Thomas Wolf, Lysandre Debut, Victor Sanh, Julien Chaumond, Clement Delangue, Anthony Moi, Pierric Cistac, Tim Rault, Remi Louf, Morgan Funtowicz, Joe Davison, Sam Shleifer, Patrick von Platen, Clara Ma, Yacine Jernite, Julien Plu, Canwen Xu, Teven Le~Scao, Sylvain Gugger, Mariama Drame, Quentin Lhoest, and Alexander Rush.
\newblock Transformers: State-of-the-art natural language processing.
\newblock In Qun Liu and David Schlangen (eds.), \emph{Proceedings of the 2020 Conference on Empirical Methods in Natural Language Processing: System Demonstrations}, pp.\  38--45, Online, October 2020. Association for Computational Linguistics.
\newblock \doi{10.18653/v1/2020.emnlp-demos.6}.
\newblock URL \url{https://aclanthology.org/2020.emnlp-demos.6/}.

\bibitem[Wortsman et~al.(2022)Wortsman, Ilharco, Gadre, Roelofs, Gontijo-Lopes, Morcos, Namkoong, Farhadi, Carmon, Kornblith, and Schmidt]{pmlr-v162-wortsman22a}
Mitchell Wortsman, Gabriel Ilharco, Samir~Ya Gadre, Rebecca Roelofs, Raphael Gontijo-Lopes, Ari~S Morcos, Hongseok Namkoong, Ali Farhadi, Yair Carmon, Simon Kornblith, and Ludwig Schmidt.
\newblock Model soups: averaging weights of multiple fine-tuned models improves accuracy without increasing inference time.
\newblock In Kamalika Chaudhuri, Stefanie Jegelka, Le~Song, Csaba Szepesvari, Gang Niu, and Sivan Sabato (eds.), \emph{Proceedings of the 39th International Conference on Machine Learning}, volume 162 of \emph{Proceedings of Machine Learning Research}, pp.\  23965--23998. PMLR, 17--23 Jul 2022.
\newblock URL \url{https://proceedings.mlr.press/v162/wortsman22a.html}.

\bibitem[Xue et~al.(2021)Xue, Constant, Roberts, Kale, Al-Rfou, Siddhant, Barua, and Raffel]{xue-etal-2021-mt5}
Linting Xue, Noah Constant, Adam Roberts, Mihir Kale, Rami Al-Rfou, Aditya Siddhant, Aditya Barua, and Colin Raffel.
\newblock m{T}5: A massively multilingual pre-trained text-to-text transformer.
\newblock In Kristina Toutanova, Anna Rumshisky, Luke Zettlemoyer, Dilek Hakkani-Tur, Iz~Beltagy, Steven Bethard, Ryan Cotterell, Tanmoy Chakraborty, and Yichao Zhou (eds.), \emph{Proceedings of the 2021 Conference of the North American Chapter of the Association for Computational Linguistics: Human Language Technologies}, pp.\  483--498, Online, June 2021. Association for Computational Linguistics.
\newblock \doi{10.18653/v1/2021.naacl-main.41}.
\newblock URL \url{https://aclanthology.org/2021.naacl-main.41/}.

\bibitem[Yadav et~al.(2023)Yadav, Tam, Choshen, Raffel, and Bansal]{yadav2023tiesmerging}
Prateek Yadav, Derek Tam, Leshem Choshen, Colin Raffel, and Mohit Bansal.
\newblock {TIES}-merging: Resolving interference when merging models.
\newblock In \emph{Proceedings of the Thirty-seventh Conference on Neural Information Processing Systems}, 2023.
\newblock URL \url{https://openreview.net/forum?id=xtaX3WyCj1}.

\bibitem[Yamaguchi et~al.(2024{\natexlab{a}})Yamaguchi, Villavicencio, and Aletras]{yamaguchi-etal-2024-empirical}
Atsuki Yamaguchi, Aline Villavicencio, and Nikolaos Aletras.
\newblock An empirical study on cross-lingual vocabulary adaptation for efficient language model inference.
\newblock In Yaser Al-Onaizan, Mohit Bansal, and Yun-Nung Chen (eds.), \emph{Findings of the Association for Computational Linguistics: EMNLP 2024}, pp.\  6760--6785, Miami, Florida, USA, November 2024{\natexlab{a}}. Association for Computational Linguistics.
\newblock \doi{10.18653/v1/2024.findings-emnlp.396}.
\newblock URL \url{https://aclanthology.org/2024.findings-emnlp.396/}.

\bibitem[Yamaguchi et~al.(2024{\natexlab{b}})Yamaguchi, Villavicencio, and Aletras]{yamaguchi2024effectivelyexpandvocabularyllms}
Atsuki Yamaguchi, Aline Villavicencio, and Nikolaos Aletras.
\newblock How can we effectively expand the vocabulary of {LLM}s with 0.01{GB} of target language text?
\newblock \emph{arXiv}, abs/2406.11477, 2024{\natexlab{b}}.
\newblock URL \url{https://arxiv.org/abs/2406.11477}.

\bibitem[Yang et~al.(2024)Yang, Yang, Hui, Zheng, Yu, Zhou, Li, Li, Liu, Huang, Dong, Wei, Lin, Tang, Wang, Yang, Tu, Zhang, Ma, Yang, Xu, Zhou, Bai, He, Lin, Dang, Lu, Chen, Yang, Li, Xue, Ni, Zhang, Wang, Peng, Men, Gao, Lin, Wang, Bai, Tan, Zhu, Li, Liu, Ge, Deng, Zhou, Ren, Zhang, Wei, Ren, Liu, Fan, Yao, Zhang, Wan, Chu, Liu, Cui, Zhang, Guo, and Fan]{yang2024qwen2technicalreport}
An~Yang, Baosong Yang, Binyuan Hui, Bo~Zheng, Bowen Yu, Chang Zhou, Chengpeng Li, Chengyuan Li, Dayiheng Liu, Fei Huang, Guanting Dong, Haoran Wei, Huan Lin, Jialong Tang, Jialin Wang, Jian Yang, Jianhong Tu, Jianwei Zhang, Jianxin Ma, Jianxin Yang, Jin Xu, Jingren Zhou, Jinze Bai, Jinzheng He, Junyang Lin, Kai Dang, Keming Lu, Keqin Chen, Kexin Yang, Mei Li, Mingfeng Xue, Na~Ni, Pei Zhang, Peng Wang, Ru~Peng, Rui Men, Ruize Gao, Runji Lin, Shijie Wang, Shuai Bai, Sinan Tan, Tianhang Zhu, Tianhao Li, Tianyu Liu, Wenbin Ge, Xiaodong Deng, Xiaohuan Zhou, Xingzhang Ren, Xinyu Zhang, Xipin Wei, Xuancheng Ren, Xuejing Liu, Yang Fan, Yang Yao, Yichang Zhang, Yu~Wan, Yunfei Chu, Yuqiong Liu, Zeyu Cui, Zhenru Zhang, Zhifang Guo, and Zhihao Fan.
\newblock Qwen2 technical report.
\newblock \emph{arXiv}, abs/2407.10671, 2024.
\newblock URL \url{https://arxiv.org/abs/2407.10671}.

\bibitem[Yang et~al.(2025)Yang, Li, Yang, Zhang, Hui, Zheng, Yu, Gao, Huang, Lv, Zheng, Liu, Zhou, Huang, Hu, Ge, Wei, Lin, Tang, Yang, Tu, Zhang, Yang, Yang, Zhou, Zhou, Lin, Dang, Bao, Yang, Yu, Deng, Li, Xue, Li, Zhang, Wang, Zhu, Men, Gao, Liu, Luo, Li, Tang, Yin, Ren, Wang, Zhang, Ren, Fan, Su, Zhang, Zhang, Wan, Liu, Wang, Cui, Zhang, Zhou, and Qiu]{yang2025qwen3technicalreport}
An~Yang, Anfeng Li, Baosong Yang, Beichen Zhang, Binyuan Hui, Bo~Zheng, Bowen Yu, Chang Gao, Chengen Huang, Chenxu Lv, Chujie Zheng, Dayiheng Liu, Fan Zhou, Fei Huang, Feng Hu, Hao Ge, Haoran Wei, Huan Lin, Jialong Tang, Jian Yang, Jianhong Tu, Jianwei Zhang, Jianxin Yang, Jiaxi Yang, Jing Zhou, Jingren Zhou, Junyang Lin, Kai Dang, Keqin Bao, Kexin Yang, Le~Yu, Lianghao Deng, Mei Li, Mingfeng Xue, Mingze Li, Pei Zhang, Peng Wang, Qin Zhu, Rui Men, Ruize Gao, Shixuan Liu, Shuang Luo, Tianhao Li, Tianyi Tang, Wenbiao Yin, Xingzhang Ren, Xinyu Wang, Xinyu Zhang, Xuancheng Ren, Yang Fan, Yang Su, Yichang Zhang, Yinger Zhang, Yu~Wan, Yuqiong Liu, Zekun Wang, Zeyu Cui, Zhenru Zhang, Zhipeng Zhou, and Zihan Qiu.
\newblock {Qwen3} technical report.
\newblock \emph{arXiv}, abs/2505.09388, 2025.
\newblock URL \url{https://arxiv.org/abs/2505.09388}.

\bibitem[Yao et~al.(2021)Yao, Huang, Wang, Dong, and Wei]{yao-etal-2021-adapt}
Yunzhi Yao, Shaohan Huang, Wenhui Wang, Li~Dong, and Furu Wei.
\newblock Adapt-and-distill: Developing small, fast and effective pretrained language models for domains.
\newblock In Chengqing Zong, Fei Xia, Wenjie Li, and Roberto Navigli (eds.), \emph{Findings of the Association for Computational Linguistics: ACL-IJCNLP 2021}, pp.\  460--470, Online, August 2021. Association for Computational Linguistics.
\newblock \doi{10.18653/v1/2021.findings-acl.40}.
\newblock URL \url{https://aclanthology.org/2021.findings-acl.40/}.

\bibitem[Yong et~al.(2023)Yong, Schoelkopf, Muennighoff, Aji, Adelani, Almubarak, Bari, Sutawika, Kasai, Baruwa, Winata, Biderman, Raff, Radev, and Nikoulina]{yong-etal-2023-bloom}
Zheng~Xin Yong, Hailey Schoelkopf, Niklas Muennighoff, Alham~Fikri Aji, David~Ifeoluwa Adelani, Khalid Almubarak, M~Saiful Bari, Lintang Sutawika, Jungo Kasai, Ahmed Baruwa, Genta Winata, Stella Biderman, Edward Raff, Dragomir Radev, and Vassilina Nikoulina.
\newblock {BLOOM}+1: Adding language support to {BLOOM} for zero-shot prompting.
\newblock In Anna Rogers, Jordan Boyd-Graber, and Naoaki Okazaki (eds.), \emph{Proceedings of the 61st Annual Meeting of the Association for Computational Linguistics (Volume 1: Long Papers)}, pp.\  11682--11703, Toronto, Canada, July 2023. Association for Computational Linguistics.
\newblock \doi{10.18653/v1/2023.acl-long.653}.
\newblock URL \url{https://aclanthology.org/2023.acl-long.653/}.

\bibitem[Yu et~al.(2024)Yu, Yu, Yu, Huang, and Li]{yu2024language}
Le~Yu, Bowen Yu, Haiyang Yu, Fei Huang, and Yongbin Li.
\newblock Language models are super mario: Absorbing abilities from homologous models as a free lunch.
\newblock In \emph{Proceedings of the Forty-first International Conference on Machine Learning}, 2024.
\newblock URL \url{https://openreview.net/forum?id=fq0NaiU8Ex}.

\bibitem[Zhao et~al.(2024{\natexlab{a}})Zhao, Zhang, Gao, Zhang, Gui, and Huang]{zhao2024llamaenglishempiricalstudy}
Jun Zhao, Zhihao Zhang, Luhui Gao, Qi~Zhang, Tao Gui, and Xuanjing Huang.
\newblock {LLaMA} beyond {English}: An empirical study on language capability transfer.
\newblock \emph{arXiv}, abs/2401.01055, 2024{\natexlab{a}}.
\newblock URL \url{https://arxiv.org/abs/2401.01055}.

\bibitem[Zhao et~al.(2024{\natexlab{b}})Zhao, Zhang, Chen, Kawaguchi, and Bing]{zhao2024how}
Yiran Zhao, Wenxuan Zhang, Guizhen Chen, Kenji Kawaguchi, and Lidong Bing.
\newblock How do large language models handle multilingualism?
\newblock In \emph{Proceedings of the Thirty-eighth Annual Conference on Neural Information Processing Systems}, 2024{\natexlab{b}}.
\newblock URL \url{https://openreview.net/forum?id=ctXYOoAgRy}.

\bibitem[Zheng et~al.(2023)Zheng, Chiang, Sheng, Zhuang, Wu, Zhuang, Lin, Li, Li, Xing, Zhang, Gonzalez, and Stoica]{zheng2023judging}
Lianmin Zheng, Wei-Lin Chiang, Ying Sheng, Siyuan Zhuang, Zhanghao Wu, Yonghao Zhuang, Zi~Lin, Zhuohan Li, Dacheng Li, Eric Xing, Hao Zhang, Joseph~E. Gonzalez, and Ion Stoica.
\newblock Judging {LLM}-as-a-judge with {MT}-bench and chatbot arena.
\newblock In \emph{Proceedings of the Thirty-seventh Conference on Neural Information Processing Systems Datasets and Benchmarks Track}, 2023.
\newblock URL \url{https://openreview.net/forum?id=uccHPGDlao}.

\bibitem[Zhou et~al.(2023)Zhou, Lu, Mishra, Brahma, Basu, Luan, Zhou, and Hou]{zhou2023instructionfollowingevaluationlargelanguage}
Jeffrey Zhou, Tianjian Lu, Swaroop Mishra, Siddhartha Brahma, Sujoy Basu, Yi~Luan, Denny Zhou, and Le~Hou.
\newblock Instruction-following evaluation for large language models.
\newblock \emph{arXiv}, abs/2311.07911, 2023.
\newblock URL \url{https://arxiv.org/abs/2311.07911}.

\end{thebibliography}
\bibliographystyle{tmlr}

\appendix

\section{Experimental Setup} \label{appendix:setup}
\subsection{Chat Template and Special Tokens}
Model-specific chat templates and special tokens are accessible via the following links:
\begin{itemize}
    \item \textbf{Qwen2.5}: \url{https://huggingface.co/Qwen/Qwen2.5-7B-Instruct/blob/main/tokenizer_config.json}

     \item \textbf{Llama 3.1}: \url{https://huggingface.co/meta-llama/Llama-3.1-8B-Instruct/blob/main/tokenizer_config.json}

     \item \textbf{Qwen3}: \url{https://huggingface.co/Qwen/Qwen3-14B/blob/main/tokenizer_config.json}
\end{itemize}

Below are excerpts from the chat templates of each model with placeholders for a prompt and output:

\paragraph{Qwen2.5.}

\begin{verbatim}
<|im_start|>system
You are Qwen, created by Alibaba Cloud. 
You are a helpful assistant.
<|im_end|>
<|im_start|>user
{prompt}
<|im_end|>
<|im_start|>assistant
{output}
\end{verbatim}

\paragraph{Llama 3.1.}

\begin{verbatim}
<|begin_of_text|>
<|start_header_id|>system
<|end_header_id|>
Cutting Knowledge Date: December 2023
Today Date: 26 Jul 2024
<|eot_id|>
<|start_header_id|>user
<|end_header_id|>
{prompt}
<|eot_id|>
<|start_header_id|>assistant
<|end_header_id|>
{output}
\end{verbatim}

\paragraph{Qwen3.}
\begin{verbatim}
<|im_start|>user
{prompt}
<|im_end|>
<|im_start|>assistant
<think>
</think>
{output}
\end{verbatim}

\subsection{Prompt Template}
We translate the English prompt templates provided by \citet{ahia-etal-2023-languages} for \textsc{sum} with a machine translation API, following \citet{yong-etal-2023-bloom}.
For \textsc{mt} and \textsc{mc}, we formulate a task-specific English prompt, followed by machine translation for each language. 
For the remaining tasks, except for \textsc{MT-Bench}, we use the default templates provided in \texttt{lm-evaluation-harness}~\citep{eval-harness}.
For \textsc{MT-Bench}, we use the default template provided in LightEval~\citep{lighteval}.
Table \ref{tab:prompt} shows the prompt templates used in our evaluation.
Note that we do not make any changes to the task-specific prompt to allow for a fair comparison between models with and without a chat template (i.e., Base and Chat, respectively).

\renewcommand*{\arraystretch}{1.0}
\begin{table*}[t]
\caption{Prompt template for each task and language.}
\begin{center}
\resizebox{0.8\linewidth}{!}{%
    \includegraphics[]{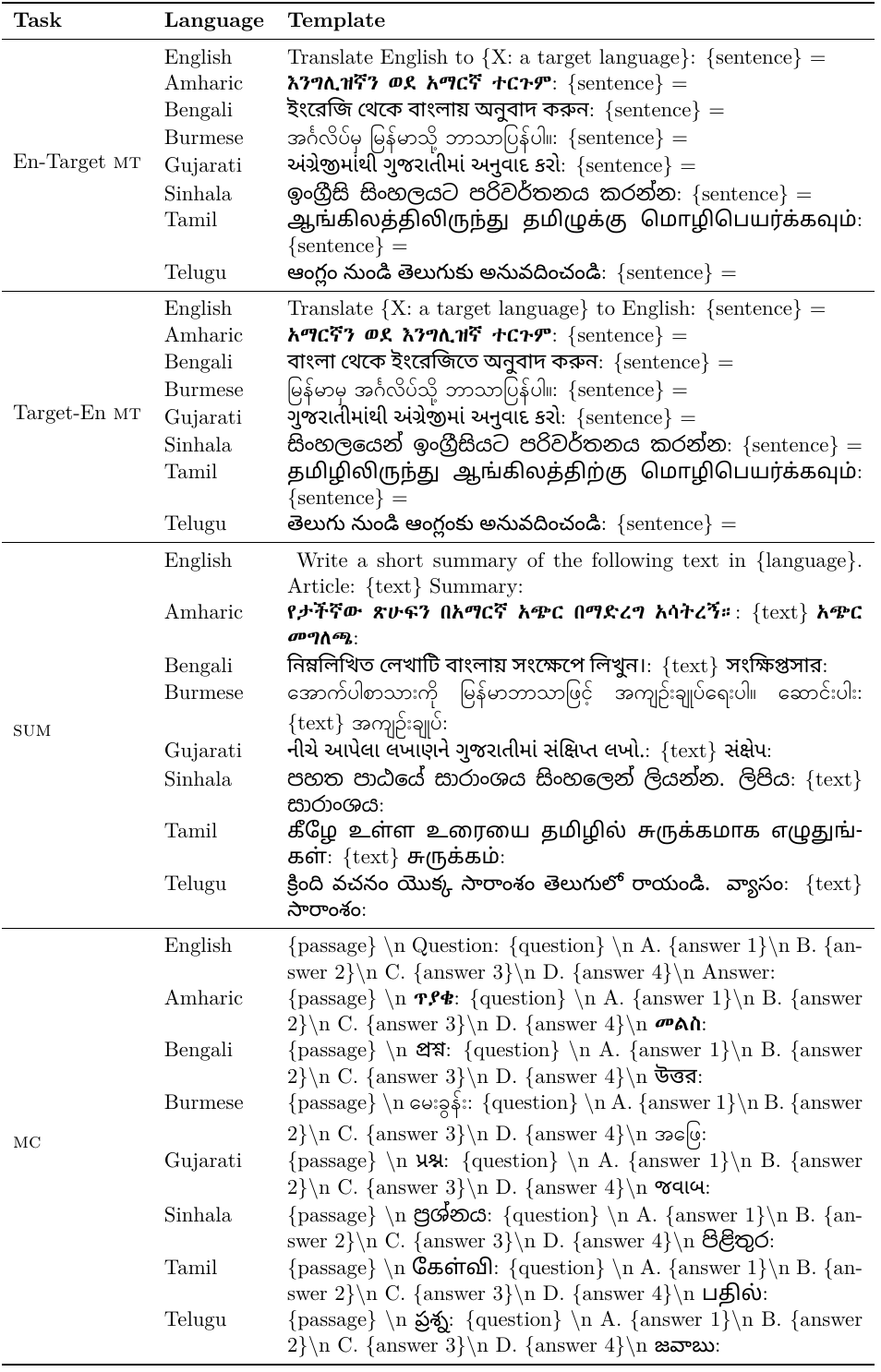}
}
\end{center}
\label{tab:prompt}
\end{table*}

\begin{table}[H]
\begin{center}
\small
\caption{Hyperparameters for continual pre-training.}
\begin{tabular}{lc}
\toprule
\textbf{Hyperparameters} & \textbf{Values} \\
\midrule
Batch size & 32\\
Number of training steps & 30,517\\
Adam $\epsilon$ & 1e-8\\
Adam $\beta_1$ & 0.9\\
Adam $\beta_2$ & 0.999\\
Sequence length & 512\\
Learning rate & 5e-5\\
Learning rate scheduler & cosine\\
Warmup steps & First 5\% of steps \\
Weight decay & 0.01\\
Attention dropout & 0.0 \\
Training precision & BF16\\
\bottomrule
\end{tabular}%
\label{tab:hyperparams_cpt}
\end{center}
\end{table}

\begin{table}[H]
\begin{center}
\small
\caption{Parameters for non-greedy generative tasks: \textsc{mt} and \textsc{sum}.}
\begin{tabular}{lc}
\toprule
\textbf{Parameters} & \textbf{Values} \\
\midrule
Temperature & 0.8\\
Repetition penalty & 1.1\\
Top $k$ & 40\\
Top $p$ & 0.9\\
Beam width & 5\\
Sampling & True\\
Early stopping & True\\
Maximum number of generated tokens & 128\\
\bottomrule
\end{tabular}%
\label{tab:params_generation}
\end{center}
\end{table}

\subsection{Implementation Details}

Our experimental design is based on the findings from \citet{tejaswi-etal-2024-exploring}. They report that (1) \textit{there are no significant gains when adding more than 10K tokens to the source vocabulary}, and (2) \textit{additional CPT in the order of millions of tokens is sufficient for model adaptation}.
Given this, we set the vocabulary size of the auxiliary target language tokenizer $|\mathcal{V}_\text{aux}|$ to 50K across languages and the number of new target tokens $k$ to 10K.
We train each model for 500M tokens with a batch size of 32, a maximum learning rate of 5e-5, and a sequence length of 512.

\paragraph{Tokenizer Training.}
We train tokenizers using Hugging Face Tokenizers.

\paragraph{Preprocessing.}
We preprocess datasets with Hugging Face Datasets~\citep{lhoest-etal-2021-datasets}.

\paragraph{Continual Pre-training.}

We implement our models using PyTorch~\citep{10.1145/3620665.3640366} and Hugging Face Transformers~\citep{wolf-etal-2020-transformers}.
Table \ref{tab:hyperparams_cpt} lists the hyperparameters in CPT.

\paragraph{Model Merging.}
To ensure a smooth transition between layers, we use a $0.3:0.7=\text{Chat}:\text{Chat+VE}$ mixing ratio for the top and bottom layers of all merged models, favoring Chat+VE as these layers are adjacent to the embeddings and language modeling head of Chat+VE.
For the second top and bottom layers, we use $0.5:0.5=\text{Chat}:\text{Chat+VE}$, balancing the contributions of Chat and Chat+VE.
For Qwen3, we use a $0.1:0.9=\text{Chat}:\text{Chat+VE}$ mixing ratio for all layers.

\paragraph{Evaluation.}
We use Hugging Face LightEval\footnote{\url{https://github.com/huggingface/lighteval}} for evaluation on all tasks except for \textsc{AlpacaEval}, \textsc{IFEval}, \textsc{GSM8K} and \textsc{MGSM}.
For \textsc{AlpacaEval}, we use the official implementation available on GitHub\footnote{\url{https://github.com/tatsu-lab/alpaca_eval}} (v0.6.6).
For \textsc{IFEval}, \textsc{GSM8K}, and \textsc{MGSM}, we use \texttt{lm-evaluation-harness}~\citep{eval-harness}.
To compute ROUGE-L, we split sentences with an mT5~\citep{xue-etal-2021-mt5} tokenizer as preprocessing following \citet{maynez-etal-2023-benchmarking} and subsequently call \texttt{rouge\_scorer}\footnote{\url{https://github.com/google-research/google-research/tree/master/rouge}} to compute the metric.
To compute chrF and chrF++, we use SacreBLEU~\citep{post-2018-call}.
For safety evaluation, we follow \citet{cahyawijaya-etal-2024-cendol} and use their implementation available on GitHub: \url{https://github.com/IndoNLP/cendol}.

Table \ref{tab:params_generation} lists the parameters used during evaluation for generative tasks: \textsc{mt} and \textsc{sum}.
To make a fair comparison, we do not conduct any generation parameter tuning and use the same ones across all approaches.

\paragraph{Hardware.}
We use either a single NVIDIA A100 (80GB), NVIDIA H100 (80GB), or NVIDIA GH200 (96GB) for CPT.
For CPT with Qwen3 14B, we use a single AMD MI300X GPU.
For evaluation, we use a single NVIDIA A100 (80GB) for all Llama 3.1 models, a single NVIDIA H100 (80GB) for all Qwen2.5 models, and a single AMD MI300X GPU for all Qwen3 models to ensure accurate measurement of inference efficiency.

\paragraph{Model Links.}
We list all the source model URLs in the following:
\begin{itemize}
    \item \textbf{Qwen2.5} (Chat): \url{https://huggingface.co/Qwen/Qwen2.5-7B-Instruct}

    \item \textbf{Qwen2.5} (Base): \url{https://huggingface.co/Qwen/Qwen2.5-7B}
    
    \item \textbf{Llama 3.1} (Chat): \url{https://huggingface.co/meta-llama/Llama-3.1-8B-Instruct}

    \item \textbf{Llama 3.1} (Base): \url{https://huggingface.co/meta-llama/Llama-3.1-8B}

    \item \textbf{Qwen3} (Base): \url{https://huggingface.co/Qwen/Qwen3-14B-Base}

    \item \textbf{Qwen3} (Chat): \url{https://huggingface.co/Qwen/Qwen3-14B}
\end{itemize}

\section{Results} \label{appendix:results}
\subsection{Task Performance}

\renewcommand*{\arraystretch}{1.0}
\begin{table*}[t]
\caption{Qwen2.5 2 7B chat, instruction-following, and safety performance.
Darker \textcolor{blue}{blue} and  \textcolor{red}{red} shades indicate higher positive and negative relative performance change over \colorbox{gray!25}{Chat} per language and task, respectively. 
Experiments are limited to models that use the chat template, except for \textsc{IFEval} to verify the performance gain of CV over its adapted base model, Base+VE.
}
\begin{center}
\resizebox{\linewidth}{!}{%
\begin{tabular}{llllllll|lllllll|lllllll}
\toprule
& \multicolumn{7}{c|}{\textbf{\textsc{IFEval}}} & \multicolumn{7}{c|}{\textbf{\textsc{GSM8K}}} & \multicolumn{7}{c}{\textbf{\textsc{MT-Bench}}}\\  
 & \multicolumn{1}{c}{\footnotesize am} & \multicolumn{1}{c}{\footnotesize bn} & \multicolumn{1}{c}{\footnotesize my} & \multicolumn{1}{c}{\footnotesize gu} & \multicolumn{1}{c}{\footnotesize si} & \multicolumn{1}{c}{\footnotesize ta} & \multicolumn{1}{c|}{\footnotesize te} 
 & \multicolumn{1}{c}{\footnotesize am} & \multicolumn{1}{c}{\footnotesize bn} & \multicolumn{1}{c}{\footnotesize my} & \multicolumn{1}{c}{\footnotesize gu} & \multicolumn{1}{c}{\footnotesize si} & \multicolumn{1}{c}{\footnotesize ta} & \multicolumn{1}{c|}{\footnotesize te} 
 & \multicolumn{1}{c}{\footnotesize am} & \multicolumn{1}{c}{\footnotesize bn} & \multicolumn{1}{c}{\footnotesize my} & \multicolumn{1}{c}{\footnotesize gu} & \multicolumn{1}{c}{\footnotesize si} & \multicolumn{1}{c}{\footnotesize ta} & \multicolumn{1}{c}{\footnotesize te}\\
 \midrule

Base+VE & \cellcolor[rgb]{1.00,0.63,0.63}.19 & \cellcolor[rgb]{1.00,0.61,0.61}.16 & \cellcolor[rgb]{1.00,0.60,0.60}.14 & \cellcolor[rgb]{1.00,0.61,0.61}.15 & \cellcolor[rgb]{1.00,0.62,0.62}.16 & \cellcolor[rgb]{1.00,0.62,0.62}.17 & \cellcolor[rgb]{1.00,0.60,0.60}.14 & - & - & - & - & - & - & - & - & - & - & - & - & - & -\\
CV & \cellcolor[rgb]{1.00,0.79,0.79}.41 & \cellcolor[rgb]{1.00,0.76,0.76}.36 & \cellcolor[rgb]{1.00,0.82,0.82}.45 & \cellcolor[rgb]{1.00,0.76,0.76}.36 & \cellcolor[rgb]{1.00,0.77,0.77}.38 & \cellcolor[rgb]{1.00,0.77,0.77}.38 & \cellcolor[rgb]{1.00,0.76,0.76}.37 & \cellcolor[rgb]{1.00,0.97,0.97}.65 & \cellcolor[rgb]{1.00,0.97,0.97}.66 & \cellcolor[rgb]{0.98,0.98,1.00}.73 & \cellcolor[rgb]{1.00,0.91,0.91}.57 & \cellcolor[rgb]{1.00,0.90,0.90}.56 & \cellcolor[rgb]{1.00,0.58,0.58}.11 & \cellcolor[rgb]{1.00,0.68,0.68}.25 & \cellcolor[rgb]{1.00,0.85,0.85}2.92 & \cellcolor[rgb]{1.00,0.90,0.90}3.35 & \cellcolor[rgb]{1.00,0.92,0.92}3.54 & \cellcolor[rgb]{1.00,0.91,0.91}3.41 & \cellcolor[rgb]{1.00,0.90,0.90}3.39 & \cellcolor[rgb]{1.00,0.88,0.88}3.22 & \cellcolor[rgb]{1.00,0.90,0.90}3.33\\
\rowcolor{gray!25}
Chat & .70 & .70 & .70 & .70 & .70 & .70 & .70 & .70 & .70 & .70 & .70 & .70 & .70 & .70 & 4.19 & 4.19 & 4.19 & 4.19 & 4.19 & 4.19 & 4.19\\
Chat+CPT & \cellcolor[rgb]{1.00,0.78,0.78}.40 & \cellcolor[rgb]{1.00,0.81,0.81}.43 & \cellcolor[rgb]{1.00,0.79,0.79}.41 & \cellcolor[rgb]{1.00,0.72,0.72}.31 & \cellcolor[rgb]{1.00,0.74,0.74}.34 & \cellcolor[rgb]{1.00,0.75,0.75}.35 & \cellcolor[rgb]{1.00,0.75,0.75}.36 & \cellcolor[rgb]{1.00,0.95,0.95}.63 & \cellcolor[rgb]{1.00,1.00,1.00}.70 & \cellcolor[rgb]{1.00,0.97,0.97}.66 & \cellcolor[rgb]{1.00,0.99,0.99}.68 & \cellcolor[rgb]{1.00,1.00,1.00}.70 & \cellcolor[rgb]{1.00,0.93,0.93}.60 & \cellcolor[rgb]{0.99,0.99,1.00}.71 & \cellcolor[rgb]{1.00,0.92,0.92}3.55 & \cellcolor[rgb]{1.00,0.92,0.92}3.51 & \cellcolor[rgb]{1.00,0.92,0.92}3.52 & \cellcolor[rgb]{1.00,0.85,0.85}2.93 & \cellcolor[rgb]{1.00,0.88,0.88}3.19 & \cellcolor[rgb]{1.00,0.90,0.90}3.38 & \cellcolor[rgb]{1.00,0.89,0.89}3.29\\
Chat+VE & \cellcolor[rgb]{1.00,0.78,0.78}.39 & \cellcolor[rgb]{1.00,0.77,0.77}.38 & \cellcolor[rgb]{1.00,0.82,0.82}.45 & \cellcolor[rgb]{1.00,0.77,0.77}.39 & \cellcolor[rgb]{1.00,0.78,0.78}.40 & \cellcolor[rgb]{1.00,0.78,0.78}.39 & \cellcolor[rgb]{1.00,0.79,0.79}.40 & \cellcolor[rgb]{1.00,0.81,0.81}.43 & \cellcolor[rgb]{1.00,0.95,0.95}.63 & \cellcolor[rgb]{0.97,0.97,1.00}.74 & \cellcolor[rgb]{1.00,0.83,0.83}.47 & \cellcolor[rgb]{1.00,0.90,0.90}.57 & \cellcolor[rgb]{1.00,0.70,0.70}.29 & \cellcolor[rgb]{1.00,0.81,0.81}.44 & \cellcolor[rgb]{1.00,0.87,0.87}3.12 & \cellcolor[rgb]{1.00,0.91,0.91}3.43 & \cellcolor[rgb]{1.00,0.94,0.94}3.67 & \cellcolor[rgb]{1.00,0.90,0.90}3.31 & \cellcolor[rgb]{1.00,0.92,0.92}3.50 & \cellcolor[rgb]{1.00,0.90,0.90}3.34 & \cellcolor[rgb]{1.00,0.88,0.88}3.21\\
ElChat & \cellcolor[rgb]{1.00,0.87,0.87}.52 & \cellcolor[rgb]{1.00,0.88,0.88}.54 & \cellcolor[rgb]{1.00,0.92,0.92}.59 & \cellcolor[rgb]{1.00,0.87,0.87}.52 & \cellcolor[rgb]{1.00,0.88,0.88}.53 & \cellcolor[rgb]{1.00,0.87,0.87}.52 & \cellcolor[rgb]{1.00,0.89,0.89}.55 & \cellcolor[rgb]{1.00,0.97,0.97}.66 & \cellcolor[rgb]{0.98,0.98,1.00}.72 & \cellcolor[rgb]{0.97,0.97,1.00}.74 & \cellcolor[rgb]{1.00,0.99,0.99}.68 & \cellcolor[rgb]{0.98,0.98,1.00}.72 & \cellcolor[rgb]{0.99,0.99,1.00}.71 & \cellcolor[rgb]{1.00,1.00,1.00}.70 & \cellcolor[rgb]{1.00,0.89,0.89}3.30 & \cellcolor[rgb]{1.00,0.93,0.93}3.59 & \cellcolor[rgb]{1.00,0.95,0.95}3.81 & \cellcolor[rgb]{1.00,0.93,0.93}3.58 & \cellcolor[rgb]{1.00,0.95,0.95}3.75 & \cellcolor[rgb]{1.00,0.94,0.94}3.69 & \cellcolor[rgb]{1.00,0.93,0.93}3.58\\

\midrule
& \multicolumn{7}{c|}{\textbf{\textsc{TruthfulQA}}} & \multicolumn{7}{c|}{\textbf{\textsc{ToxiGen}}} & \multicolumn{7}{c}{\textbf{\textsc{ImplicitHate}}}\\  
 & \multicolumn{1}{c}{\footnotesize am} & \multicolumn{1}{c}{\footnotesize bn} & \multicolumn{1}{c}{\footnotesize my} & \multicolumn{1}{c}{\footnotesize gu} & \multicolumn{1}{c}{\footnotesize si} & \multicolumn{1}{c}{\footnotesize ta} & \multicolumn{1}{c|}{\footnotesize te} 
 & \multicolumn{1}{c}{\footnotesize am} & \multicolumn{1}{c}{\footnotesize bn} & \multicolumn{1}{c}{\footnotesize my} & \multicolumn{1}{c}{\footnotesize gu} & \multicolumn{1}{c}{\footnotesize si} & \multicolumn{1}{c}{\footnotesize ta} & \multicolumn{1}{c|}{\footnotesize te} 
 & \multicolumn{1}{c}{\footnotesize am} & \multicolumn{1}{c}{\footnotesize bn} & \multicolumn{1}{c}{\footnotesize my} & \multicolumn{1}{c}{\footnotesize gu} & \multicolumn{1}{c}{\footnotesize si} & \multicolumn{1}{c}{\footnotesize ta} & \multicolumn{1}{c}{\footnotesize te}\\
 \midrule

CV & \cellcolor[rgb]{0.62,0.62,1.00}.55 & \cellcolor[rgb]{0.94,0.94,1.00}.57 & \cellcolor[rgb]{0.71,0.71,1.00}.52 & \cellcolor[rgb]{0.80,0.80,1.00}.58 & \cellcolor[rgb]{0.71,0.71,1.00}.50 & \cellcolor[rgb]{0.55,0.55,1.00}.62 & \cellcolor[rgb]{0.56,0.56,1.00}.60 & \cellcolor[rgb]{0.50,0.50,1.00}.21 & \cellcolor[rgb]{0.55,0.55,1.00}.23 & \cellcolor[rgb]{0.50,0.50,1.00}.22 & \cellcolor[rgb]{0.50,0.50,1.00}.24 & \cellcolor[rgb]{0.50,0.50,1.00}.23 & \cellcolor[rgb]{0.50,0.50,1.00}.22 & \cellcolor[rgb]{0.50,0.50,1.00}.22 & \cellcolor[rgb]{0.63,0.63,1.00}.16 & \cellcolor[rgb]{0.63,0.63,1.00}.20 & \cellcolor[rgb]{0.50,0.50,1.00}.19 & \cellcolor[rgb]{0.50,0.50,1.00}.22 & \cellcolor[rgb]{0.50,0.50,1.00}.20 & \cellcolor[rgb]{0.50,0.50,1.00}.21 & \cellcolor[rgb]{0.50,0.50,1.00}.19\\
\rowcolor{gray!25}
Chat & .31 & .50 & .33 & .41 & .32 & .33 & .32 & .10 & .12 & .07 & .10 & .09 & .09 & .09 & .09 & .12 & .09 & .10 & .07 & .09 & .09\\
Chat+CPT & \cellcolor[rgb]{0.62,0.62,1.00}.55 & \cellcolor[rgb]{0.91,0.91,1.00}.59 & \cellcolor[rgb]{0.93,0.93,1.00}.38 & \cellcolor[rgb]{0.82,0.82,1.00}.56 & \cellcolor[rgb]{0.76,0.76,1.00}.47 & \cellcolor[rgb]{0.59,0.59,1.00}.59 & \cellcolor[rgb]{0.64,0.64,1.00}.55 & \cellcolor[rgb]{0.70,0.70,1.00}.17 & \cellcolor[rgb]{0.79,0.79,1.00}.17 & \cellcolor[rgb]{0.79,0.79,1.00}.10 & \cellcolor[rgb]{0.84,0.84,1.00}.13 & \cellcolor[rgb]{0.77,0.77,1.00}.13 & \cellcolor[rgb]{0.75,0.75,1.00}.14 & \cellcolor[rgb]{0.80,0.80,1.00}.12 & \cellcolor[rgb]{0.61,0.61,1.00}.16 & \cellcolor[rgb]{0.81,0.81,1.00}.16 & \cellcolor[rgb]{0.85,0.85,1.00}.11 & \cellcolor[rgb]{0.81,0.81,1.00}.13 & \cellcolor[rgb]{0.59,0.59,1.00}.13 & \cellcolor[rgb]{0.75,0.75,1.00}.14 & \cellcolor[rgb]{0.79,0.79,1.00}.12\\
Chat+VE & \cellcolor[rgb]{0.56,0.56,1.00}.59 & \cellcolor[rgb]{0.95,0.95,1.00}.56 & \cellcolor[rgb]{0.73,0.73,1.00}.51 & \cellcolor[rgb]{0.78,0.78,1.00}.60 & \cellcolor[rgb]{0.73,0.73,1.00}.49 & \cellcolor[rgb]{0.60,0.60,1.00}.59 & \cellcolor[rgb]{0.62,0.62,1.00}.56 & \cellcolor[rgb]{0.63,0.63,1.00}.18 & \cellcolor[rgb]{0.64,0.64,1.00}.21 & \cellcolor[rgb]{0.50,0.50,1.00}.19 & \cellcolor[rgb]{0.50,0.50,1.00}.21 & \cellcolor[rgb]{0.50,0.50,1.00}.21 & \cellcolor[rgb]{0.50,0.50,1.00}.19 & \cellcolor[rgb]{0.50,0.50,1.00}.20 & \cellcolor[rgb]{0.75,0.75,1.00}.14 & \cellcolor[rgb]{0.69,0.69,1.00}.19 & \cellcolor[rgb]{0.50,0.50,1.00}.18 & \cellcolor[rgb]{0.50,0.50,1.00}.20 & \cellcolor[rgb]{0.50,0.50,1.00}.19 & \cellcolor[rgb]{0.54,0.54,1.00}.18 & \cellcolor[rgb]{0.50,0.50,1.00}.18\\
ElChat & \cellcolor[rgb]{0.63,0.63,1.00}.54 & \cellcolor[rgb]{0.90,0.90,1.00}.60 & \cellcolor[rgb]{0.76,0.76,1.00}.49 & \cellcolor[rgb]{0.80,0.80,1.00}.58 & \cellcolor[rgb]{0.67,0.67,1.00}.53 & \cellcolor[rgb]{0.58,0.58,1.00}.60 & \cellcolor[rgb]{0.54,0.54,1.00}.61 & \cellcolor[rgb]{0.93,0.93,1.00}.12 & \cellcolor[rgb]{0.76,0.76,1.00}.18 & \cellcolor[rgb]{0.50,0.50,1.00}.17 & \cellcolor[rgb]{0.64,0.64,1.00}.17 & \cellcolor[rgb]{0.63,0.63,1.00}.16 & \cellcolor[rgb]{0.63,0.63,1.00}.16 & \cellcolor[rgb]{0.57,0.57,1.00}.16 & \cellcolor[rgb]{0.98,0.98,1.00}.09 & \cellcolor[rgb]{0.79,0.79,1.00}.17 & \cellcolor[rgb]{0.55,0.55,1.00}.17 & \cellcolor[rgb]{0.64,0.64,1.00}.17 & \cellcolor[rgb]{0.51,0.51,1.00}.14 & \cellcolor[rgb]{0.72,0.72,1.00}.15 & \cellcolor[rgb]{0.66,0.66,1.00}.15\\

 \bottomrule
\end{tabular}
}
\end{center}
\label{tab:chat_performance_qwen25}
\end{table*}

\renewcommand*{\arraystretch}{1.0}
\begin{table*}[t]
\caption{Llama 3.1 8B chat, instruction-following, and safety performance.
Darker \textcolor{blue}{blue} and  \textcolor{red}{red} shades indicate higher positive and negative relative performance change over \colorbox{gray!25}{Chat} per language and task, respectively.
(L) stands for linear merging.
Experiments are limited to models that use the chat template, except for \textsc{IFEval} to verify the performance gain of CV over its adapted base model, Base+VE.
}
\begin{center}
\resizebox{\linewidth}{!}{%
\begin{tabular}{llllllll|lllllll|lllllll}
\toprule
& \multicolumn{7}{c|}{\textbf{\textsc{IFEval}}} & \multicolumn{7}{c|}{\textbf{\textsc{GSM8K}}} & \multicolumn{7}{c}{\textbf{\textsc{MT-Bench}}}\\  
 & \multicolumn{1}{c}{\footnotesize am} & \multicolumn{1}{c}{\footnotesize bn} & \multicolumn{1}{c}{\footnotesize my} & \multicolumn{1}{c}{\footnotesize gu} & \multicolumn{1}{c}{\footnotesize si} & \multicolumn{1}{c}{\footnotesize ta} & \multicolumn{1}{c|}{\footnotesize te} 
 & \multicolumn{1}{c}{\footnotesize am} & \multicolumn{1}{c}{\footnotesize bn} & \multicolumn{1}{c}{\footnotesize my} & \multicolumn{1}{c}{\footnotesize gu} & \multicolumn{1}{c}{\footnotesize si} & \multicolumn{1}{c}{\footnotesize ta} & \multicolumn{1}{c|}{\footnotesize te} 
 & \multicolumn{1}{c}{\footnotesize am} & \multicolumn{1}{c}{\footnotesize bn} & \multicolumn{1}{c}{\footnotesize my} & \multicolumn{1}{c}{\footnotesize gu} & \multicolumn{1}{c}{\footnotesize si} & \multicolumn{1}{c}{\footnotesize ta} & \multicolumn{1}{c}{\footnotesize te}\\
 \midrule
Base+VE & \cellcolor[rgb]{1.00,0.61,0.61}.17 & \cellcolor[rgb]{1.00,0.59,0.59}.14 & \cellcolor[rgb]{1.00,0.59,0.59}.13 & \cellcolor[rgb]{1.00,0.57,0.57}.11 & \cellcolor[rgb]{1.00,0.60,0.60}.15 & \cellcolor[rgb]{1.00,0.58,0.58}.12 & \cellcolor[rgb]{1.00,0.58,0.58}.12 & - & - & - & - & - & - & - & - & - & - & - & - & - & -\\
CV & \cellcolor[rgb]{1.00,0.75,0.75}.38 & \cellcolor[rgb]{1.00,0.77,0.77}.39 & \cellcolor[rgb]{1.00,0.79,0.79}.42 & \cellcolor[rgb]{1.00,0.74,0.74}.35 & \cellcolor[rgb]{1.00,0.78,0.78}.41 & \cellcolor[rgb]{1.00,0.74,0.74}.35 & \cellcolor[rgb]{1.00,0.74,0.74}.35 & \cellcolor[rgb]{1.00,0.56,0.56}.10 & \cellcolor[rgb]{1.00,0.66,0.66}.27 & \cellcolor[rgb]{1.00,0.70,0.70}.34 & \cellcolor[rgb]{1.00,0.68,0.68}.30 & \cellcolor[rgb]{1.00,0.75,0.75}.43 & \cellcolor[rgb]{1.00,0.74,0.74}.40 & \cellcolor[rgb]{1.00,0.78,0.78}.48 & \cellcolor[rgb]{1.00,0.87,0.87}2.93 & \cellcolor[rgb]{1.00,0.89,0.89}3.10 & \cellcolor[rgb]{1.00,0.94,0.94}3.49 & \cellcolor[rgb]{1.00,0.86,0.86}2.84 & \cellcolor[rgb]{1.00,0.88,0.88}3.03 & \cellcolor[rgb]{1.00,0.88,0.88}3.04 & \cellcolor[rgb]{1.00,0.86,0.86}2.88\\
\rowcolor{gray!25}
Chat & .73 & .73 & .73 & .73 & .73 & .73 & .73 & .84 & .84 & .84 & .84 & .84 & .84 & .84 & 3.93 & 3.93 & 3.93 & 3.93 & 3.93 & 3.93 & 3.93\\
Chat+CPT & \cellcolor[rgb]{1.00,0.62,0.62}.18 & \cellcolor[rgb]{1.00,0.71,0.71}.31 & \cellcolor[rgb]{1.00,0.75,0.75}.37 & \cellcolor[rgb]{1.00,0.66,0.66}.23 & \cellcolor[rgb]{1.00,0.72,0.72}.32 & \cellcolor[rgb]{1.00,0.74,0.74}.36 & \cellcolor[rgb]{1.00,0.74,0.74}.36 & \cellcolor[rgb]{1.00,0.77,0.77}.46 & \cellcolor[rgb]{1.00,0.80,0.80}.51 & \cellcolor[rgb]{1.00,0.84,0.84}.58 & \cellcolor[rgb]{1.00,0.75,0.75}.43 & \cellcolor[rgb]{1.00,0.80,0.80}.51 & \cellcolor[rgb]{1.00,0.80,0.80}.51 & \cellcolor[rgb]{1.00,0.76,0.76}.45 & \cellcolor[rgb]{1.00,0.81,0.81}2.44 & \cellcolor[rgb]{1.00,0.78,0.78}2.23 & \cellcolor[rgb]{1.00,0.83,0.83}2.61 & \cellcolor[rgb]{1.00,0.81,0.81}2.48 & \cellcolor[rgb]{1.00,0.85,0.85}2.76 & \cellcolor[rgb]{1.00,0.77,0.77}2.15 & \cellcolor[rgb]{1.00,0.85,0.85}2.72\\
Chat+VE & \cellcolor[rgb]{1.00,0.73,0.73}.33 & \cellcolor[rgb]{1.00,0.75,0.75}.37 & \cellcolor[rgb]{1.00,0.75,0.75}.37 & \cellcolor[rgb]{1.00,0.72,0.72}.32 & \cellcolor[rgb]{1.00,0.75,0.75}.37 & \cellcolor[rgb]{1.00,0.75,0.75}.36 & \cellcolor[rgb]{1.00,0.69,0.69}.28 & \cellcolor[rgb]{1.00,0.65,0.65}.26 & \cellcolor[rgb]{1.00,0.82,0.82}.55 & \cellcolor[rgb]{1.00,0.72,0.72}.38 & \cellcolor[rgb]{1.00,0.67,0.67}.29 & \cellcolor[rgb]{1.00,0.76,0.76}.44 & \cellcolor[rgb]{1.00,0.70,0.70}.35 & \cellcolor[rgb]{1.00,0.54,0.54}.07 & \cellcolor[rgb]{1.00,0.79,0.79}2.28 & \cellcolor[rgb]{1.00,0.86,0.86}2.85 & \cellcolor[rgb]{1.00,0.84,0.84}2.64 & \cellcolor[rgb]{1.00,0.79,0.79}2.29 & \cellcolor[rgb]{1.00,0.82,0.82}2.54 & \cellcolor[rgb]{1.00,0.85,0.85}2.71 & \cellcolor[rgb]{1.00,0.80,0.80}2.34\\
ElChat & \cellcolor[rgb]{1.00,0.81,0.81}.45 & \cellcolor[rgb]{1.00,0.80,0.80}.44 & \cellcolor[rgb]{1.00,0.86,0.86}.54 & \cellcolor[rgb]{1.00,0.80,0.80}.44 & \cellcolor[rgb]{1.00,0.82,0.82}.47 & \cellcolor[rgb]{1.00,0.85,0.85}.51 & \cellcolor[rgb]{1.00,0.82,0.82}.47 & \cellcolor[rgb]{1.00,0.83,0.83}.57 & \cellcolor[rgb]{1.00,0.84,0.84}.57 & \cellcolor[rgb]{1.00,0.93,0.93}.72 & \cellcolor[rgb]{1.00,0.83,0.83}.56 & \cellcolor[rgb]{1.00,0.85,0.85}.60 & \cellcolor[rgb]{1.00,0.83,0.83}.56 & \cellcolor[rgb]{1.00,0.81,0.81}.52 & \cellcolor[rgb]{1.00,0.86,0.86}2.86 & \cellcolor[rgb]{1.00,0.85,0.85}2.72 & \cellcolor[rgb]{1.00,0.95,0.95}3.56 & \cellcolor[rgb]{1.00,0.86,0.86}2.81 & \cellcolor[rgb]{1.00,0.85,0.85}2.79 & \cellcolor[rgb]{1.00,0.88,0.88}2.96 & \cellcolor[rgb]{1.00,0.85,0.85}2.76\\
\midrule
ElChat \textbackslash Merge & \cellcolor[rgb]{1.00,0.68,0.68}.27 & \cellcolor[rgb]{1.00,0.75,0.75}.38 & \cellcolor[rgb]{1.00,0.77,0.77}.41 & \cellcolor[rgb]{1.00,0.70,0.70}.30 & \cellcolor[rgb]{1.00,0.75,0.75}.38 & \cellcolor[rgb]{1.00,0.73,0.73}.33 & \cellcolor[rgb]{1.00,0.71,0.71}.31 & \cellcolor[rgb]{1.00,0.60,0.60}.17 & \cellcolor[rgb]{1.00,0.82,0.82}.54 & \cellcolor[rgb]{1.00,0.84,0.84}.58 & \cellcolor[rgb]{1.00,0.75,0.75}.42 & \cellcolor[rgb]{1.00,0.83,0.83}.55 & \cellcolor[rgb]{1.00,0.78,0.78}.47 & \cellcolor[rgb]{1.00,0.55,0.55}.09 & \cellcolor[rgb]{1.00,0.79,0.79}2.29 & \cellcolor[rgb]{1.00,0.85,0.85}2.79 & \cellcolor[rgb]{1.00,0.86,0.86}2.86 & \cellcolor[rgb]{1.00,0.81,0.81}2.42 & \cellcolor[rgb]{1.00,0.84,0.84}2.70 & \cellcolor[rgb]{1.00,0.79,0.79}2.28 & \cellcolor[rgb]{1.00,0.81,0.81}2.40\\
ElChat \textbackslash Copy & \cellcolor[rgb]{1.00,0.77,0.77}.40 & \cellcolor[rgb]{1.00,0.75,0.75}.37 & \cellcolor[rgb]{1.00,0.79,0.79}.43 & \cellcolor[rgb]{1.00,0.74,0.74}.35 & \cellcolor[rgb]{1.00,0.77,0.77}.40 & \cellcolor[rgb]{1.00,0.78,0.78}.41 & \cellcolor[rgb]{1.00,0.77,0.77}.39 & \cellcolor[rgb]{1.00,0.59,0.59}.15 & \cellcolor[rgb]{1.00,0.76,0.76}.44 & \cellcolor[rgb]{1.00,0.89,0.89}.66 & \cellcolor[rgb]{1.00,0.58,0.58}.14 & \cellcolor[rgb]{1.00,0.75,0.75}.42 & \cellcolor[rgb]{1.00,0.74,0.74}.41 & \cellcolor[rgb]{1.00,0.71,0.71}.36 & \cellcolor[rgb]{1.00,0.83,0.83}2.61 & \cellcolor[rgb]{1.00,0.86,0.86}2.82 & \cellcolor[rgb]{1.00,0.89,0.89}3.09 & \cellcolor[rgb]{1.00,0.81,0.81}2.45 & \cellcolor[rgb]{1.00,0.81,0.81}2.41 & \cellcolor[rgb]{1.00,0.84,0.84}2.70 & \cellcolor[rgb]{1.00,0.81,0.81}2.48\\
ElChat (L) & \cellcolor[rgb]{1.00,0.83,0.83}.48 & \cellcolor[rgb]{1.00,0.80,0.80}.44 & \cellcolor[rgb]{1.00,0.88,0.88}.55 & \cellcolor[rgb]{1.00,0.79,0.79}.42 & \cellcolor[rgb]{1.00,0.85,0.85}.50 & \cellcolor[rgb]{1.00,0.85,0.85}.50 & \cellcolor[rgb]{1.00,0.83,0.83}.48 & \cellcolor[rgb]{1.00,0.83,0.83}.56 & \cellcolor[rgb]{1.00,0.83,0.83}.55 & \cellcolor[rgb]{1.00,0.94,0.94}.74 & \cellcolor[rgb]{1.00,0.82,0.82}.55 & \cellcolor[rgb]{1.00,0.86,0.86}.60 & \cellcolor[rgb]{1.00,0.83,0.83}.55 & \cellcolor[rgb]{1.00,0.80,0.80}.51 & \cellcolor[rgb]{1.00,0.87,0.87}2.91 & \cellcolor[rgb]{1.00,0.88,0.88}2.97 & \cellcolor[rgb]{1.00,0.95,0.95}3.58 & \cellcolor[rgb]{1.00,0.84,0.84}2.69 & \cellcolor[rgb]{1.00,0.85,0.85}2.76 & \cellcolor[rgb]{1.00,0.91,0.91}3.23 & \cellcolor[rgb]{1.00,0.85,0.85}2.75\\

\midrule
& \multicolumn{7}{c|}{\textbf{\textsc{TruthfulQA}}} & \multicolumn{7}{c|}{\textbf{\textsc{ToxiGen}}} & \multicolumn{7}{c}{\textbf{\textsc{ImplicitHate}}}\\  
 & \multicolumn{1}{c}{\footnotesize am} & \multicolumn{1}{c}{\footnotesize bn} & \multicolumn{1}{c}{\footnotesize my} & \multicolumn{1}{c}{\footnotesize gu} & \multicolumn{1}{c}{\footnotesize si} & \multicolumn{1}{c}{\footnotesize ta} & \multicolumn{1}{c|}{\footnotesize te} 
 & \multicolumn{1}{c}{\footnotesize am} & \multicolumn{1}{c}{\footnotesize bn} & \multicolumn{1}{c}{\footnotesize my} & \multicolumn{1}{c}{\footnotesize gu} & \multicolumn{1}{c}{\footnotesize si} & \multicolumn{1}{c}{\footnotesize ta} & \multicolumn{1}{c|}{\footnotesize te} 
 & \multicolumn{1}{c}{\footnotesize am} & \multicolumn{1}{c}{\footnotesize bn} & \multicolumn{1}{c}{\footnotesize my} & \multicolumn{1}{c}{\footnotesize gu} & \multicolumn{1}{c}{\footnotesize si} & \multicolumn{1}{c}{\footnotesize ta} & \multicolumn{1}{c}{\footnotesize te}\\
 \midrule

CV & \cellcolor[rgb]{1.00,0.66,0.66}.11 & \cellcolor[rgb]{1.00,0.85,0.85}.30 & \cellcolor[rgb]{0.65,0.65,1.00}.25 & \cellcolor[rgb]{1.00,0.61,0.61}.10 & \cellcolor[rgb]{1.00,0.75,0.75}.15 & \cellcolor[rgb]{1.00,0.97,0.97}.42 & \cellcolor[rgb]{1.00,0.99,0.99}.39 & \cellcolor[rgb]{0.50,0.50,1.00}.17 & \cellcolor[rgb]{0.69,0.69,1.00}.19 & \cellcolor[rgb]{0.50,0.50,1.00}.19 & \cellcolor[rgb]{0.50,0.50,1.00}.21 & \cellcolor[rgb]{0.50,0.50,1.00}.17 & \cellcolor[rgb]{0.50,0.50,1.00}.18 & \cellcolor[rgb]{0.50,0.50,1.00}.20 & \cellcolor[rgb]{0.88,0.88,1.00}.11 & \cellcolor[rgb]{0.73,0.73,1.00}.17 & \cellcolor[rgb]{0.50,0.50,1.00}.17 & \cellcolor[rgb]{0.50,0.50,1.00}.20 & \cellcolor[rgb]{0.54,0.54,1.00}.15 & \cellcolor[rgb]{0.59,0.59,1.00}.16 & \cellcolor[rgb]{0.50,0.50,1.00}.17\\
\rowcolor{gray!25}
Chat & .34 & .43 & .15 & .46 & .30 & .45 & .40 & .08 & .12 & .07 & .09 & .08 & .09 & .08 & .09 & .11 & .07 & .10 & .08 & .09 & .08\\
Chat+CPT & \cellcolor[rgb]{1.00,0.50,0.50}.00 & \cellcolor[rgb]{1.00,0.57,0.57}.06 & \cellcolor[rgb]{0.50,0.50,1.00}.80 & \cellcolor[rgb]{0.70,0.70,1.00}.74 & \cellcolor[rgb]{0.50,0.50,1.00}.94 & \cellcolor[rgb]{1.00,0.50,0.50}.00 & \cellcolor[rgb]{0.61,0.61,1.00}.71 & \cellcolor[rgb]{0.50,0.50,1.00}.16 & \cellcolor[rgb]{0.77,0.77,1.00}.17 & \cellcolor[rgb]{0.64,0.64,1.00}.11 & \cellcolor[rgb]{0.74,0.74,1.00}.14 & \cellcolor[rgb]{0.67,0.67,1.00}.13 & \cellcolor[rgb]{0.81,0.81,1.00}.12 & \cellcolor[rgb]{0.71,0.71,1.00}.13 & \cellcolor[rgb]{0.54,0.54,1.00}.18 & \cellcolor[rgb]{0.76,0.76,1.00}.17 & \cellcolor[rgb]{0.65,0.65,1.00}.12 & \cellcolor[rgb]{0.74,0.74,1.00}.15 & \cellcolor[rgb]{0.59,0.59,1.00}.14 & \cellcolor[rgb]{0.80,0.80,1.00}.13 & \cellcolor[rgb]{0.69,0.69,1.00}.13\\
Chat+VE & \cellcolor[rgb]{1.00,0.95,0.95}.30 & \cellcolor[rgb]{1.00,0.84,0.84}.29 & \cellcolor[rgb]{1.00,0.82,0.82}.09 & \cellcolor[rgb]{1.00,0.56,0.56}.06 & \cellcolor[rgb]{0.99,0.99,1.00}.30 & \cellcolor[rgb]{1.00,0.80,0.80}.27 & \cellcolor[rgb]{1.00,0.76,0.76}.21 & \cellcolor[rgb]{0.50,0.50,1.00}.18 & \cellcolor[rgb]{0.58,0.58,1.00}.22 & \cellcolor[rgb]{0.50,0.50,1.00}.17 & \cellcolor[rgb]{0.50,0.50,1.00}.22 & \cellcolor[rgb]{0.50,0.50,1.00}.19 & \cellcolor[rgb]{0.50,0.50,1.00}.21 & \cellcolor[rgb]{0.50,0.50,1.00}.23 & \cellcolor[rgb]{0.76,0.76,1.00}.14 & \cellcolor[rgb]{0.62,0.62,1.00}.20 & \cellcolor[rgb]{0.50,0.50,1.00}.17 & \cellcolor[rgb]{0.50,0.50,1.00}.21 & \cellcolor[rgb]{0.50,0.50,1.00}.17 & \cellcolor[rgb]{0.50,0.50,1.00}.20 & \cellcolor[rgb]{0.50,0.50,1.00}.21\\
ElChat & \cellcolor[rgb]{0.87,0.87,1.00}.43 & \cellcolor[rgb]{0.99,0.99,1.00}.44 & \cellcolor[rgb]{0.65,0.65,1.00}.25 & \cellcolor[rgb]{1.00,0.66,0.66}.15 & \cellcolor[rgb]{0.70,0.70,1.00}.48 & \cellcolor[rgb]{1.00,0.75,0.75}.22 & \cellcolor[rgb]{0.68,0.68,1.00}.65 & \cellcolor[rgb]{0.88,0.88,1.00}.10 & \cellcolor[rgb]{0.74,0.74,1.00}.18 & \cellcolor[rgb]{0.50,0.50,1.00}.14 & \cellcolor[rgb]{0.57,0.57,1.00}.18 & \cellcolor[rgb]{0.63,0.63,1.00}.14 & \cellcolor[rgb]{0.52,0.52,1.00}.17 & \cellcolor[rgb]{0.59,0.59,1.00}.15 & \cellcolor[rgb]{1.00,0.83,0.83}.06 & \cellcolor[rgb]{0.80,0.80,1.00}.16 & \cellcolor[rgb]{0.54,0.54,1.00}.14 & \cellcolor[rgb]{0.62,0.62,1.00}.17 & \cellcolor[rgb]{0.74,0.74,1.00}.12 & \cellcolor[rgb]{0.64,0.64,1.00}.16 & \cellcolor[rgb]{0.66,0.66,1.00}.13\\
\midrule
ElChat \textbackslash Merge & \cellcolor[rgb]{0.50,0.50,1.00}.89 & \cellcolor[rgb]{1.00,0.89,0.89}.33 & \cellcolor[rgb]{1.00,0.81,0.81}.09 & \cellcolor[rgb]{1.00,0.86,0.86}.33 & \cellcolor[rgb]{1.00,0.97,0.97}.28 & \cellcolor[rgb]{0.64,0.64,1.00}.77 & \cellcolor[rgb]{0.83,0.83,1.00}.53 & \cellcolor[rgb]{0.57,0.57,1.00}.15 & \cellcolor[rgb]{0.61,0.61,1.00}.21 & \cellcolor[rgb]{0.50,0.50,1.00}.18 & \cellcolor[rgb]{0.50,0.50,1.00}.21 & \cellcolor[rgb]{0.50,0.50,1.00}.19 & \cellcolor[rgb]{0.50,0.50,1.00}.21 & \cellcolor[rgb]{0.50,0.50,1.00}.22 & \cellcolor[rgb]{1.00,0.81,0.81}.06 & \cellcolor[rgb]{0.74,0.74,1.00}.17 & \cellcolor[rgb]{0.50,0.50,1.00}.15 & \cellcolor[rgb]{0.56,0.56,1.00}.18 & \cellcolor[rgb]{0.58,0.58,1.00}.14 & \cellcolor[rgb]{0.60,0.60,1.00}.16 & \cellcolor[rgb]{0.62,0.62,1.00}.14\\
ElChat \textbackslash Copy & \cellcolor[rgb]{0.88,0.88,1.00}.42 & \cellcolor[rgb]{0.97,0.97,1.00}.45 & \cellcolor[rgb]{0.57,0.57,1.00}.27 & \cellcolor[rgb]{1.00,0.67,0.67}.16 & \cellcolor[rgb]{0.64,0.64,1.00}.52 & \cellcolor[rgb]{1.00,0.79,0.79}.26 & \cellcolor[rgb]{0.74,0.74,1.00}.60 & \cellcolor[rgb]{0.89,0.89,1.00}.10 & \cellcolor[rgb]{0.69,0.69,1.00}.19 & \cellcolor[rgb]{0.50,0.50,1.00}.15 & \cellcolor[rgb]{0.50,0.50,1.00}.19 & \cellcolor[rgb]{0.50,0.50,1.00}.16 & \cellcolor[rgb]{0.50,0.50,1.00}.18 & \cellcolor[rgb]{0.54,0.54,1.00}.16 & \cellcolor[rgb]{1.00,0.82,0.82}.06 & \cellcolor[rgb]{0.74,0.74,1.00}.17 & \cellcolor[rgb]{0.50,0.50,1.00}.15 & \cellcolor[rgb]{0.55,0.55,1.00}.18 & \cellcolor[rgb]{0.59,0.59,1.00}.14 & \cellcolor[rgb]{0.60,0.60,1.00}.16 & \cellcolor[rgb]{0.62,0.62,1.00}.14\\
ElChat (L) & \cellcolor[rgb]{0.86,0.86,1.00}.43 & \cellcolor[rgb]{0.98,0.98,1.00}.44 & \cellcolor[rgb]{0.70,0.70,1.00}.23 & \cellcolor[rgb]{1.00,0.65,0.65}.14 & \cellcolor[rgb]{0.72,0.72,1.00}.46 & \cellcolor[rgb]{1.00,0.76,0.76}.24 & \cellcolor[rgb]{0.70,0.70,1.00}.63 & \cellcolor[rgb]{0.88,0.88,1.00}.10 & \cellcolor[rgb]{0.74,0.74,1.00}.18 & \cellcolor[rgb]{0.50,0.50,1.00}.14 & \cellcolor[rgb]{0.57,0.57,1.00}.18 & \cellcolor[rgb]{0.63,0.63,1.00}.14 & \cellcolor[rgb]{0.54,0.54,1.00}.17 & \cellcolor[rgb]{0.58,0.58,1.00}.15 & \cellcolor[rgb]{1.00,0.83,0.83}.06 & \cellcolor[rgb]{0.80,0.80,1.00}.16 & \cellcolor[rgb]{0.56,0.56,1.00}.14 & \cellcolor[rgb]{0.63,0.63,1.00}.17 & \cellcolor[rgb]{0.74,0.74,1.00}.12 & \cellcolor[rgb]{0.65,0.65,1.00}.16 & \cellcolor[rgb]{0.66,0.66,1.00}.13\\

 \bottomrule
\end{tabular}
}
\end{center}

\label{tab:chat_performance_llama31}
\end{table*}

\renewcommand*{\arraystretch}{1.0}
\begin{table}[t]
\caption{Qwen3 14B chat, instruction-following, and safety performance.
Darker \textcolor{blue}{blue} and  \textcolor{red}{red} shades indicate higher positive and negative relative performance change over \colorbox{gray!25}{Chat} per language and task, respectively. 
Experiments are limited to models that use the chat template, except for \textsc{IFEval} to verify the performance gain of CV over its adapted base model, Base+VE.
}
\begin{center}
\begin{tabular}{llll|lll|lll}
\toprule
& \multicolumn{3}{c|}{\textbf{\textsc{IFEval}}} & \multicolumn{3}{c|}{\textbf{\textsc{GSM8K}}} & \multicolumn{3}{c}{\textbf{\textsc{MT-Bench}}}\\  
 & \multicolumn{1}{c}{\footnotesize am} & \multicolumn{1}{c}{\footnotesize bn} & \multicolumn{1}{c|}{\footnotesize te} 
 & \multicolumn{1}{c}{\footnotesize am} & \multicolumn{1}{c}{\footnotesize bn} & \multicolumn{1}{c|}{\footnotesize te} 
 & \multicolumn{1}{c}{\footnotesize am} & \multicolumn{1}{c}{\footnotesize bn} & \multicolumn{1}{c}{\footnotesize te}\\
 \midrule

Base+VE & \cellcolor[rgb]{1.00,0.71,0.71}.27 & \cellcolor[rgb]{1.00,0.71,0.71}.27 & \cellcolor[rgb]{1.00,0.70,0.70}.26 & - & - & - & - & - & -\\
CV & \cellcolor[rgb]{1.00,0.68,0.68}.24 & \cellcolor[rgb]{1.00,0.75,0.75}.33 & \cellcolor[rgb]{1.00,0.75,0.75}.33 & \cellcolor[rgb]{1.00,0.66,0.66}.27 & \cellcolor[rgb]{1.00,0.92,0.92}.69 & \cellcolor[rgb]{1.00,0.74,0.74}.40 & \cellcolor[rgb]{1.00,0.92,0.92}3.58 & \cellcolor[rgb]{1.00,0.95,0.95}3.78 & \cellcolor[rgb]{1.00,0.94,0.94}3.77\\
\rowcolor{gray!25}
Chat & .65 & .65 & .65 & .83 & .83 & .83 & 4.24 & 4.24 & 4.24\\
Chat+CPT & \cellcolor[rgb]{1.00,0.93,0.93}.57 & \cellcolor[rgb]{1.00,0.98,0.98}.62 & \cellcolor[rgb]{1.00,0.99,0.99}.63 & \cellcolor[rgb]{1.00,0.98,0.98}.80 & \cellcolor[rgb]{0.99,0.99,1.00}.84 & \cellcolor[rgb]{0.98,0.98,1.00}.86 & \cellcolor[rgb]{1.00,0.98,0.98}4.07 & \cellcolor[rgb]{1.00,0.99,0.99}4.18 & \cellcolor[rgb]{0.99,0.99,1.00}4.29\\
Chat+VE & \cellcolor[rgb]{1.00,0.93,0.93}.56 & \cellcolor[rgb]{1.00,0.95,0.95}.58 & \cellcolor[rgb]{1.00,0.90,0.90}.52 & \cellcolor[rgb]{1.00,0.95,0.95}.74 & \cellcolor[rgb]{1.00,0.98,0.98}.79 & \cellcolor[rgb]{1.00,0.81,0.81}.51 & \cellcolor[rgb]{1.00,0.92,0.92}3.51 & \cellcolor[rgb]{1.00,0.99,0.99}4.18 & \cellcolor[rgb]{1.00,0.97,0.97}4.00\\
\bottomrule
\end{tabular}
\end{center}
\label{tab:chat_performance_qwen3}
\end{table}

\begin{table}[t]
\centering
\small
\caption{\textsc{MGSM} performance in Bengali (bn) and Telugu (te) by model. The experimental settings are the same as \textsc{GSM8K}. (L) stands for linear merging.}
\begin{subtable}{0.3\linewidth}
    \centering
    \caption{Qwen2.5 7B}
    \begin{tabular}{lcc}
        \toprule
        \textbf{Model} & \multicolumn{2}{c}{\textbf{\textsc{EM}}} \\
        & bn & te\\
        \midrule
        \rowcolor{gray!25}
        Chat & .23 & .06\\
        CV & .60 & .27\\
        Chat+CPT & .24 & .14\\
        Chat+VE & .39 & .27\\ \midrule
        ElChat (Ours) & .46 & .35\\
        \bottomrule
    \end{tabular}    
\end{subtable}%
\hspace{1em}
\begin{subtable}{0.3\linewidth}
    \centering
    \caption{Llama 3.1 8B}
    \begin{tabular}{lcc}
        \toprule
        \textbf{Model} & \multicolumn{2}{c}{\textbf{\textsc{EM}}} \\
        & bn & te\\
        \midrule
        \rowcolor{gray!25}
        Chat & .30 & .12\\
        CV & .31 & .24\\
        Chat+CPT & .12 & .02\\
        Chat+VE & .26 & .28\\
        \midrule
        ElChat (Ours) & .51 & .41\\
        \midrule
        ElChat \textbackslash Copy & .28 & .36\\
        ElChat \textbackslash Merge & .43 & .30\\
        ElChat (L) & .53 & .42\\
        \bottomrule
    \end{tabular}
\end{subtable}
\hspace{1em}
\begin{subtable}{0.3\linewidth}
    \centering
    \caption{Qwen3 14B}
    \begin{tabular}{lcc}
        \toprule
        \textbf{Model} & \multicolumn{2}{c}{\textbf{\textsc{EM}}} \\
        & bn & te\\
        \midrule
        \rowcolor{gray!25}
        Chat & .34 & .17\\
        CV & .47 & .28\\
        Chat+CPT & .30 & .16\\
        Chat+VE & .37 & .20\\
        \midrule
        ElChat (Ours) & .48 & .22\\
        \bottomrule
    \end{tabular}
\end{subtable}
\label{tab:mgsm_aux}
\end{table}

\renewcommand*{\arraystretch}{1.0}
\begin{table}[t]
\caption{Win rates over GPT-4 (1106 Preview) on \textsc{AlpacaEval} v2.0.
Darker \textcolor{red}{red} shades indicate larger negative relative performance change over \colorbox{gray!25}{Chat} per language and base model, respectively.
Experiments are limited to Amharic (am), Bengali (bn), and Telugu (te) due to computational constraints.
}
\begin{center}
\begin{tabular}{llll|lll|lll}
\toprule
& \multicolumn{3}{c|}{\textbf{Qwen2.5 7B}} & \multicolumn{3}{c|}{\textbf{Llama 3.1 8B}} & \multicolumn{3}{c}{\textbf{Qwen3 14B}}\\  
 & \multicolumn{1}{c}{\footnotesize am} & \multicolumn{1}{c}{\footnotesize bn} & \multicolumn{1}{c|}{\footnotesize te} 
 & \multicolumn{1}{c}{\footnotesize am} & \multicolumn{1}{c}{\footnotesize bn} & \multicolumn{1}{c|}{\footnotesize te} 
 & \multicolumn{1}{c}{\footnotesize am} & \multicolumn{1}{c}{\footnotesize bn} & \multicolumn{1}{c}{\footnotesize te}\\
 \midrule

CV & \cellcolor[rgb]{1.00,0.54,0.54}2.60 & \cellcolor[rgb]{1.00,0.52,0.52}1.27 & \cellcolor[rgb]{1.00,0.54,0.54}2.50 & \cellcolor[rgb]{1.00,0.53,0.53}2.38 & \cellcolor[rgb]{1.00,0.54,0.54}2.91 & \cellcolor[rgb]{1.00,0.54,0.54}2.43 & \cellcolor[rgb]{1.00,0.60,0.60}6.44 & \cellcolor[rgb]{1.00,0.58,0.58}4.90 & \cellcolor[rgb]{1.00,0.61,0.61}6.95\\
\rowcolor{gray!25}
Chat & 30.80 & 30.80 & 30.80 & 34.43 & 34.43 & 34.43 & 30.66 & 30.66 & 30.66\\
Chat+CPT & \cellcolor[rgb]{1.00,0.57,0.57}4.67 & \cellcolor[rgb]{1.00,0.55,0.55}3.07 & \cellcolor[rgb]{1.00,0.56,0.56}3.75 & \cellcolor[rgb]{1.00,0.53,0.53}2.16 & \cellcolor[rgb]{1.00,0.52,0.52}1.68 & \cellcolor[rgb]{1.00,0.54,0.54}2.53 & \cellcolor[rgb]{1.00,0.66,0.66}10.08 & \cellcolor[rgb]{1.00,0.90,0.90}24.22 & \cellcolor[rgb]{1.00,0.94,0.94}27.09\\
Chat+VE & \cellcolor[rgb]{1.00,0.53,0.53}2.09 & \cellcolor[rgb]{1.00,0.56,0.56}3.65 & \cellcolor[rgb]{1.00,0.53,0.53}2.08 & \cellcolor[rgb]{1.00,0.53,0.53}2.33 & \cellcolor[rgb]{1.00,0.53,0.53}2.10 & \cellcolor[rgb]{1.00,0.52,0.52}1.62 & \cellcolor[rgb]{1.00,0.64,0.64}8.42 & \cellcolor[rgb]{1.00,0.65,0.65}9.12 & \cellcolor[rgb]{1.00,0.61,0.61}6.74\\
\bottomrule
\end{tabular}
\end{center}
\label{tab:alpacaeval}
\end{table}

\renewcommand*{\arraystretch}{1.0}
\begin{table*}[t]
\caption{Qwen2.5 7B task performance.
Darker \textcolor{blue}{blue} and  \textcolor{red}{red} shades indicate higher positive and negative relative performance change over \colorbox{gray!25}{Chat} per language and task, respectively. 
Note that \textsc{gmmlu} does not cover Burmese (my), Gujarati (gu), and Tamil (ta).
}
\begin{center}
\resizebox{\linewidth}{!}{%
\begin{tabular}{llllllll|lllllll|lllllll}
\toprule
 & \multicolumn{7}{c|}{\textbf{Target \textsc{sum}}} & \multicolumn{7}{c|}{\textbf{English $\rightarrow$ Target \textsc{mt}}} & \multicolumn{7}{c}{\textbf{Target \textsc{mc}}}\\  
 & \multicolumn{1}{c}{\footnotesize am} & \multicolumn{1}{c}{\footnotesize bn} & \multicolumn{1}{c}{\footnotesize my} & \multicolumn{1}{c}{\footnotesize gu} & \multicolumn{1}{c}{\footnotesize si} & \multicolumn{1}{c}{\footnotesize ta} & \multicolumn{1}{c|}{\footnotesize te} 
 & \multicolumn{1}{c}{\footnotesize am} & \multicolumn{1}{c}{\footnotesize bn} & \multicolumn{1}{c}{\footnotesize my} & \multicolumn{1}{c}{\footnotesize gu} & \multicolumn{1}{c}{\footnotesize si} & \multicolumn{1}{c}{\footnotesize ta} & \multicolumn{1}{c|}{\footnotesize te} 
 & \multicolumn{1}{c}{\footnotesize am} & \multicolumn{1}{c}{\footnotesize bn} & \multicolumn{1}{c}{\footnotesize my} & \multicolumn{1}{c}{\footnotesize gu} & \multicolumn{1}{c}{\footnotesize si} & \multicolumn{1}{c}{\footnotesize ta} & \multicolumn{1}{c}{\footnotesize te}\\
\bottomrule
\end{tabular}
}
\end{center}
\label{tab:task_performance_qwen25}
\end{table*}

\renewcommand*{\arraystretch}{1.0}
\begin{table*}[t]
\caption{Llama 3.1 8B task performance.
Darker \textcolor{blue}{blue} and  \textcolor{red}{red} shades indicate higher positive and negative relative performance change over \colorbox{gray!25}{Chat} per language and task, respectively.
(L) stands for linear merging.
Note that \textsc{gmmlu} does not cover Burmese (my), Gujarati (gu), and Tamil (ta).
}
\begin{center}
\resizebox{\linewidth}{!}{%
\begin{tabular}{llllllll|lllllll|lllllll}
\toprule
 & \multicolumn{7}{c|}{\textbf{Target \textsc{sum}}} & \multicolumn{7}{c|}{\textbf{English $\rightarrow$ Target \textsc{mt}}} & \multicolumn{7}{c}{\textbf{Target \textsc{mc}}}\\  
 & \multicolumn{1}{c}{\footnotesize am} & \multicolumn{1}{c}{\footnotesize bn} & \multicolumn{1}{c}{\footnotesize my} & \multicolumn{1}{c}{\footnotesize gu} & \multicolumn{1}{c}{\footnotesize si} & \multicolumn{1}{c}{\footnotesize ta} & \multicolumn{1}{c|}{\footnotesize te} 
 & \multicolumn{1}{c}{\footnotesize am} & \multicolumn{1}{c}{\footnotesize bn} & \multicolumn{1}{c}{\footnotesize my} & \multicolumn{1}{c}{\footnotesize gu} & \multicolumn{1}{c}{\footnotesize si} & \multicolumn{1}{c}{\footnotesize ta} & \multicolumn{1}{c|}{\footnotesize te} 
 & \multicolumn{1}{c}{\footnotesize am} & \multicolumn{1}{c}{\footnotesize bn} & \multicolumn{1}{c}{\footnotesize my} & \multicolumn{1}{c}{\footnotesize gu} & \multicolumn{1}{c}{\footnotesize si} & \multicolumn{1}{c}{\footnotesize ta} & \multicolumn{1}{c}{\footnotesize te}\\
\bottomrule
\end{tabular}
}
\end{center}

\label{tab:task_performance_llama31}
\end{table*}

\renewcommand*{\arraystretch}{1.0}
\begin{table*}[t]
\caption{Qwen3 14B task performance.
Darker \textcolor{blue}{blue} and  \textcolor{red}{red} shades indicate higher positive and negative relative performance change over \colorbox{gray!25}{Chat} per language and task, respectively.
}
\begin{center}
\begin{tabular}{llll|lll|lll}
\toprule
 & \multicolumn{3}{c|}{\textbf{Target \textsc{sum}}} & \multicolumn{3}{c|}{\textbf{English $\rightarrow$ Target \textsc{mt}}} & \multicolumn{3}{c}{\textbf{Target \textsc{mc}}}\\  
 & \multicolumn{1}{c}{\footnotesize am} & \multicolumn{1}{c}{\footnotesize bn} & \multicolumn{1}{c|}{\footnotesize te} 
 & \multicolumn{1}{c}{\footnotesize am} & \multicolumn{1}{c}{\footnotesize bn} & \multicolumn{1}{c|}{\footnotesize te} 
 & \multicolumn{1}{c}{\footnotesize am} & \multicolumn{1}{c}{\footnotesize bn} & \multicolumn{1}{c}{\footnotesize te}\\

\midrule
Base & \cellcolor[rgb]{1.00,1.00,1.00}17.0\textsubscript{{\tiny0.0}} & \cellcolor[rgb]{1.00,0.95,0.95}28.7\textsubscript{{\tiny0.2}} & \cellcolor[rgb]{1.00,0.95,0.95}20.7\textsubscript{{\tiny0.1}} & \cellcolor[rgb]{1.00,0.99,0.99}13.5\textsubscript{{\tiny0.3}} & \cellcolor[rgb]{0.99,0.99,1.00}42.2\textsubscript{{\tiny0.1}} & \cellcolor[rgb]{0.96,0.96,1.00}36.4\textsubscript{{\tiny0.1}} & \cellcolor[rgb]{0.97,0.97,1.00}.56 & \cellcolor[rgb]{1.00,1.00,1.00}.84 & \cellcolor[rgb]{0.97,0.97,1.00}.81\\
Base+CPT & \cellcolor[rgb]{1.00,1.00,1.00}17.0\textsubscript{{\tiny0.0}} & \cellcolor[rgb]{1.00,0.99,0.99}31.1\textsubscript{{\tiny0.0}} & \cellcolor[rgb]{1.00,0.96,0.96}20.9\textsubscript{{\tiny0.0}} & \cellcolor[rgb]{0.50,0.50,1.00}28.4\textsubscript{{\tiny0.1}} & \cellcolor[rgb]{0.97,0.97,1.00}43.5\textsubscript{{\tiny0.0}} & \cellcolor[rgb]{0.93,0.93,1.00}38.7\textsubscript{{\tiny0.1}} & \cellcolor[rgb]{0.77,0.77,1.00}.77 & \cellcolor[rgb]{0.99,0.99,1.00}.85 & \cellcolor[rgb]{0.98,0.98,1.00}.79\\
\bottomrule
\end{tabular}
\end{center}
\label{tab:task_performance_qwen3}
\end{table*}

\renewcommand*{\arraystretch}{1.0}
\begin{table*}[t]
\caption{Qwen2.5 7B task performance.
Darker \textcolor{blue}{blue} and  \textcolor{red}{red} shades indicate higher positive and negative relative performance change over \colorbox{gray!25}{Chat} per language and task, respectively.
}
\begin{center}
\resizebox{\linewidth}{!}{%
\begin{tabular}{llllllll|lllllll}
\toprule
 & \multicolumn{7}{c|}{\textbf{Target \textsc{sum}}} & \multicolumn{7}{c}{\textbf{English $\rightarrow$ Target \textsc{mt}}}\\  
 & \multicolumn{7}{c|}{[\textbf{ROUGE-L}]} & \multicolumn{7}{c}{[\textbf{chrF++}]}\\
 & \multicolumn{1}{c}{\footnotesize am} & \multicolumn{1}{c}{\footnotesize bn} & \multicolumn{1}{c}{\footnotesize my} & \multicolumn{1}{c}{\footnotesize gu} & \multicolumn{1}{c}{\footnotesize si} & \multicolumn{1}{c}{\footnotesize ta} & \multicolumn{1}{c|}{\footnotesize te} 
 & \multicolumn{1}{c}{\footnotesize am} & \multicolumn{1}{c}{\footnotesize bn} & \multicolumn{1}{c}{\footnotesize my} & \multicolumn{1}{c}{\footnotesize gu} & \multicolumn{1}{c}{\footnotesize si} & \multicolumn{1}{c}{\footnotesize ta} & \multicolumn{1}{c}{\footnotesize te}\\

\midrule
Base & \cellcolor[rgb]{0.88,0.88,1.00}1.8\textsubscript{{\tiny0.1}} & \cellcolor[rgb]{0.85,0.85,1.00}1.7\textsubscript{{\tiny0.0}} & \cellcolor[rgb]{0.59,0.59,1.00}7.2\textsubscript{{\tiny0.0}} & \cellcolor[rgb]{1.00,0.96,0.96}2.8\textsubscript{{\tiny0.1}} & \cellcolor[rgb]{0.91,0.91,1.00}42.4\textsubscript{{\tiny0.5}} & \cellcolor[rgb]{0.94,0.94,1.00}2.4\textsubscript{{\tiny0.0}} & \cellcolor[rgb]{0.87,0.87,1.00}18.7\textsubscript{{\tiny0.1}} & \cellcolor[rgb]{0.86,0.86,1.00}3.8\textsubscript{{\tiny0.1}} & \cellcolor[rgb]{0.97,0.97,1.00}28.3\textsubscript{{\tiny0.1}} & \cellcolor[rgb]{1.00,0.95,0.95}12.3\textsubscript{{\tiny0.2}} & \cellcolor[rgb]{0.98,0.98,1.00}19.6\textsubscript{{\tiny0.1}} & \cellcolor[rgb]{1.00,0.86,0.86}9.9\textsubscript{{\tiny0.0}} & \cellcolor[rgb]{1.00,0.98,0.98}20.7\textsubscript{{\tiny0.1}} & \cellcolor[rgb]{1.00,0.95,0.95}15.3\textsubscript{{\tiny0.1}}\\
\bottomrule
\end{tabular}
}
\end{center}

\label{tab:task_performance_qwen25_aux}
\end{table*}

\renewcommand*{\arraystretch}{1.0}
\begin{table*}[t]
\caption{Llama 3.1 8B task performance.
Darker \textcolor{blue}{blue} and  \textcolor{red}{red} shades indicate higher positive and negative relative performance change over \colorbox{gray!25}{Chat} per language and task, respectively.
(L) stands for linear merging.
}
\begin{center}
\resizebox{\linewidth}{!}{%
\begin{tabular}{llllllll|lllllll}
\toprule
 & \multicolumn{7}{c|}{\textbf{Target \textsc{sum}}} & \multicolumn{7}{c}{\textbf{English $\rightarrow$ Target \textsc{mt}}}\\  
 & \multicolumn{7}{c|}{[\textbf{ROUGE-L}]} & \multicolumn{7}{c}{[\textbf{chrF++}]}\\
 & \multicolumn{1}{c}{\footnotesize am} & \multicolumn{1}{c}{\footnotesize bn} & \multicolumn{1}{c}{\footnotesize my} & \multicolumn{1}{c}{\footnotesize gu} & \multicolumn{1}{c}{\footnotesize si} & \multicolumn{1}{c}{\footnotesize ta} & \multicolumn{1}{c|}{\footnotesize te} 
 & \multicolumn{1}{c}{\footnotesize am} & \multicolumn{1}{c}{\footnotesize bn} & \multicolumn{1}{c}{\footnotesize my} & \multicolumn{1}{c}{\footnotesize gu} & \multicolumn{1}{c}{\footnotesize si} & \multicolumn{1}{c}{\footnotesize ta} & \multicolumn{1}{c}{\footnotesize te}\\

\midrule
Base & \cellcolor[rgb]{1.00,0.98,0.98}1.6\textsubscript{{\tiny0.1}} & \cellcolor[rgb]{0.68,0.68,1.00}1.9\textsubscript{{\tiny0.0}} & \cellcolor[rgb]{0.75,0.75,1.00}6.2\textsubscript{{\tiny0.1}} & \cellcolor[rgb]{0.99,0.99,1.00}3.2\textsubscript{{\tiny0.0}} & \cellcolor[rgb]{1.00,0.89,0.89}36.3\textsubscript{{\tiny0.4}} & \cellcolor[rgb]{1.00,0.83,0.83}2.0\textsubscript{{\tiny0.1}} & \cellcolor[rgb]{1.00,0.93,0.93}18.3\textsubscript{{\tiny0.2}} & \cellcolor[rgb]{0.97,0.97,1.00}8.4\textsubscript{{\tiny0.1}} & \cellcolor[rgb]{1.00,1.00,1.00}35.6\textsubscript{{\tiny0.0}} & \cellcolor[rgb]{1.00,1.00,1.00}16.9\textsubscript{{\tiny0.1}} & \cellcolor[rgb]{1.00,0.98,0.98}27.0\textsubscript{{\tiny0.1}} & \cellcolor[rgb]{0.96,0.96,1.00}22.5\textsubscript{{\tiny0.1}} & \cellcolor[rgb]{1.00,0.99,0.99}31.5\textsubscript{{\tiny0.2}} & \cellcolor[rgb]{1.00,0.97,0.97}27.3\textsubscript{{\tiny0.1}}\\
\bottomrule
\end{tabular}
}
\end{center}

\label{tab:task_performance_llama31_aux}
\end{table*}

\renewcommand*{\arraystretch}{1.0}
\begin{table*}[t]
\caption{Qwen3 14B task performance.
Darker \textcolor{blue}{blue} and  \textcolor{red}{red} shades indicate higher positive and negative relative performance change over \colorbox{gray!25}{Chat} per language and task, respectively.
}
\begin{center}
\begin{tabular}{llll|lll}
\toprule
 & \multicolumn{3}{c|}{\textbf{Target \textsc{sum}}} & \multicolumn{3}{c}{\textbf{English $\rightarrow$ Target \textsc{mt}}}\\  
 & \multicolumn{3}{c|}{[\textbf{ROUGE-L}]} & \multicolumn{3}{c}{[\textbf{chrF++}]}\\
 & \multicolumn{1}{c}{\footnotesize am} & \multicolumn{1}{c}{\footnotesize bn} & \multicolumn{1}{c|}{\footnotesize te} 
 & \multicolumn{1}{c}{\footnotesize am} & \multicolumn{1}{c}{\footnotesize bn} & \multicolumn{1}{c}{\footnotesize te}\\
\midrule

Base & \cellcolor[rgb]{0.96,0.96,1.00}2.5\textsubscript{{\tiny0.0}} & \cellcolor[rgb]{1.00,1.00,1.00}0.0\textsubscript{{\tiny0.0}} & \cellcolor[rgb]{1.00,0.90,0.90}2.4\textsubscript{{\tiny0.0}} & \cellcolor[rgb]{1.00,1.00,1.00}11.7\textsubscript{{\tiny0.3}} & \cellcolor[rgb]{0.99,0.99,1.00}38.3\textsubscript{{\tiny0.1}} & \cellcolor[rgb]{0.96,0.96,1.00}32.6\textsubscript{{\tiny0.1}}\\
Base+CPT & \cellcolor[rgb]{1.00,0.97,0.97}2.2\textsubscript{{\tiny0.0}} & \cellcolor[rgb]{1.00,1.00,1.00}0.0\textsubscript{{\tiny0.0}} & \cellcolor[rgb]{1.00,0.90,0.90}2.4\textsubscript{{\tiny0.1}} & \cellcolor[rgb]{0.50,0.50,1.00}25.8\textsubscript{{\tiny0.1}} & \cellcolor[rgb]{0.97,0.97,1.00}39.3\textsubscript{{\tiny0.0}} & \cellcolor[rgb]{0.92,0.92,1.00}34.6\textsubscript{{\tiny0.2}}\\
Base+VE & \cellcolor[rgb]{0.94,0.94,1.00}2.6\textsubscript{{\tiny0.2}} & \cellcolor[rgb]{1.00,1.00,1.00}0.0\textsubscript{{\tiny0.0}} & \cellcolor[rgb]{0.92,0.92,1.00}3.4\textsubscript{{\tiny0.1}} & \cellcolor[rgb]{0.74,0.74,1.00}17.9\textsubscript{{\tiny0.1}} & \cellcolor[rgb]{1.00,0.89,0.89}29.3\textsubscript{{\tiny0.2}} & \cellcolor[rgb]{1.00,0.99,0.99}29.6\textsubscript{{\tiny0.2}}\\
\bottomrule
\end{tabular}
\end{center}

\label{tab:task_performance_qwen3_aux}
\end{table*}

\begin{itemize}
    \item \textbf{Safety, Chat, and Instruction-following}: Tables \ref{tab:chat_performance_qwen25}, \ref{tab:chat_performance_llama31}, and \ref{tab:chat_performance_qwen3} provide a detailed breakdown of the task performance results for Qwen2.5, Llama 3.1, and Qwen3 across safety, chat, and instruction-following tasks. Table \ref{tab:mgsm_aux} shows the results of Chat+CPT and ElChat-related ablation models in \textsc{mgsm}.
    Table \ref{tab:alpacaeval} shows the results of Chat-related models on \textsc{AlpacaEval}.

    \item \textbf{Target Language and Source (English)}: Tables \ref{tab:task_performance_qwen25}, \ref{tab:task_performance_llama31}, and \ref{tab:task_performance_qwen3} provide a detailed breakdown of the task performance results for Qwen2.5, Llama 3.1, and Qwen3 across target language and source (English) language tasks.

    \item \textbf{\textsc{sum} and \textsc{mt} Results with Auxiliary Metrics}: Tables \ref{tab:task_performance_qwen25_aux}, \ref{tab:task_performance_llama31_aux}, and \ref{tab:task_performance_qwen3_aux} provide a detailed breakdown of \textsc{sum} and \textsc{mt} performance results, measured by ROUGE-L for \textsc{sum} and chrF++ for \textsc{mt}, for Qwen2.5, Llama 3.1, and Qwen3, respectively.
    
\end{itemize}

\subsection{Inference Efficiency}
Tables \ref{tab:speed_performance_qwen25}, \ref{tab:speed_performance_llama31}, and \ref{tab:speed_performance_qwen3} provide a detailed breakdown of the inference efficiency results for Qwen2.5, Llama 3.1, and Qwen3, respectively.

\renewcommand*{\arraystretch}{1.0}
\begin{table*}[t]
\caption{Qwen2.5 7B inference speedup measured by the number of tokens generated per second.
Darker \textcolor{blue}{blue} and  \textcolor{red}{red} shades indicate higher positive and negative relative performance change over \colorbox{gray!25}{Chat} per language and task, respectively. }
\begin{center}
\resizebox{\linewidth}{!}{%
\begin{tabular}{llllllll|lllllll|lllllll}
\toprule
 & \multicolumn{7}{c|}{\textbf{Target \textsc{sum}}} & \multicolumn{7}{c|}{\textbf{English $\rightarrow$ Target \textsc{mt}}} & \multicolumn{7}{c}{\textbf{Target \textsc{mc}}}\\  
 & \multicolumn{1}{c}{\footnotesize am} & \multicolumn{1}{c}{\footnotesize bn} & \multicolumn{1}{c}{\footnotesize my} & \multicolumn{1}{c}{\footnotesize gu} & \multicolumn{1}{c}{\footnotesize si} & \multicolumn{1}{c}{\footnotesize ta} & \multicolumn{1}{c|}{\footnotesize te} 
 & \multicolumn{1}{c}{\footnotesize am} & \multicolumn{1}{c}{\footnotesize bn} & \multicolumn{1}{c}{\footnotesize my} & \multicolumn{1}{c}{\footnotesize gu} & \multicolumn{1}{c}{\footnotesize si} & \multicolumn{1}{c}{\footnotesize ta} & \multicolumn{1}{c|}{\footnotesize te} 
 & \multicolumn{1}{c}{\footnotesize am} & \multicolumn{1}{c}{\footnotesize bn} & \multicolumn{1}{c}{\footnotesize my} & \multicolumn{1}{c}{\footnotesize gu} & \multicolumn{1}{c}{\footnotesize si} & \multicolumn{1}{c}{\footnotesize ta} & \multicolumn{1}{c}{\footnotesize te}\\

\midrule
Base & \cellcolor[rgb]{1.00,0.98,0.98}29.3\textsubscript{{\tiny0.3}} & \cellcolor[rgb]{1.00,1.00,1.00}30.7\textsubscript{{\tiny0.2}} & \cellcolor[rgb]{1.00,0.97,0.97}27.8\textsubscript{{\tiny0.2}} & \cellcolor[rgb]{1.00,1.00,1.00}27.2\textsubscript{{\tiny0.1}} & \cellcolor[rgb]{0.95,0.95,1.00}32.4\textsubscript{{\tiny0.1}} & \cellcolor[rgb]{1.00,0.97,0.97}31.6\textsubscript{{\tiny0.2}} & \cellcolor[rgb]{1.00,0.97,0.97}27.6\textsubscript{{\tiny0.1}} & \cellcolor[rgb]{0.98,0.98,1.00}34.1\textsubscript{{\tiny0.3}} & \cellcolor[rgb]{1.00,0.98,0.98}33.4\textsubscript{{\tiny0.4}} & \cellcolor[rgb]{1.00,0.98,0.98}33.2\textsubscript{{\tiny0.2}} & \cellcolor[rgb]{1.00,0.98,0.98}33.9\textsubscript{{\tiny0.2}} & \cellcolor[rgb]{1.00,0.98,0.98}36.4\textsubscript{{\tiny0.3}} & \cellcolor[rgb]{1.00,0.98,0.98}33.9\textsubscript{{\tiny0.1}} & \cellcolor[rgb]{1.00,0.97,0.97}34.0\textsubscript{{\tiny0.3}} & \cellcolor[rgb]{0.98,0.98,1.00}28.9 & \cellcolor[rgb]{0.99,0.99,1.00}22.6 & \cellcolor[rgb]{0.99,0.99,1.00}15.7 & \cellcolor[rgb]{0.98,0.98,1.00}17.5 & \cellcolor[rgb]{0.99,0.99,1.00}18.2 & \cellcolor[rgb]{0.98,0.98,1.00}20.9 & \cellcolor[rgb]{1.00,1.00,1.00}17.1\\
Base+CPT & \cellcolor[rgb]{0.99,0.99,1.00}30.7\textsubscript{{\tiny1.1}} & \cellcolor[rgb]{1.00,0.93,0.93}26.6\textsubscript{{\tiny2.0}} & \cellcolor[rgb]{1.00,0.99,0.99}28.8\textsubscript{{\tiny0.2}} & \cellcolor[rgb]{0.99,0.99,1.00}28.1\textsubscript{{\tiny0.1}} & \cellcolor[rgb]{1.00,0.97,0.97}28.0\textsubscript{{\tiny1.6}} & \cellcolor[rgb]{1.00,1.00,1.00}33.1\textsubscript{{\tiny0.3}} & \cellcolor[rgb]{1.00,1.00,1.00}29.7\textsubscript{{\tiny0.2}} & \cellcolor[rgb]{0.95,0.95,1.00}35.5\textsubscript{{\tiny0.1}} & \cellcolor[rgb]{0.99,0.99,1.00}35.4\textsubscript{{\tiny0.2}} & \cellcolor[rgb]{1.00,1.00,1.00}34.6\textsubscript{{\tiny0.0}} & \cellcolor[rgb]{1.00,0.99,0.99}35.0\textsubscript{{\tiny0.1}} & \cellcolor[rgb]{1.00,0.98,0.98}36.3\textsubscript{{\tiny0.3}} & \cellcolor[rgb]{1.00,1.00,1.00}35.3\textsubscript{{\tiny0.3}} & \cellcolor[rgb]{1.00,0.99,0.99}35.5\textsubscript{{\tiny0.1}} & \cellcolor[rgb]{0.99,0.99,1.00}28.5 & \cellcolor[rgb]{0.97,0.97,1.00}23.1 & \cellcolor[rgb]{1.00,1.00,1.00}15.2 & \cellcolor[rgb]{0.99,0.99,1.00}17.1 & \cellcolor[rgb]{0.99,0.99,1.00}18.6 & \cellcolor[rgb]{0.99,0.99,1.00}20.7 & \cellcolor[rgb]{0.97,0.97,1.00}18.0\\
Base+VE & \cellcolor[rgb]{0.50,0.50,1.00}110.4\textsubscript{{\tiny1.0}} & \cellcolor[rgb]{0.50,0.50,1.00}68.9\textsubscript{{\tiny1.2}} & \cellcolor[rgb]{0.50,0.50,1.00}103.1\textsubscript{{\tiny0.1}} & \cellcolor[rgb]{0.50,0.50,1.00}106.3\textsubscript{{\tiny0.3}} & \cellcolor[rgb]{0.50,0.50,1.00}102.3\textsubscript{{\tiny0.3}} & \cellcolor[rgb]{0.50,0.50,1.00}70.8\textsubscript{{\tiny0.7}} & \cellcolor[rgb]{0.50,0.50,1.00}89.9\textsubscript{{\tiny0.8}} & \cellcolor[rgb]{0.50,0.50,1.00}86.5\textsubscript{{\tiny0.3}} & \cellcolor[rgb]{0.50,0.50,1.00}72.2\textsubscript{{\tiny0.7}} & \cellcolor[rgb]{0.50,0.50,1.00}93.5\textsubscript{{\tiny0.7}} & \cellcolor[rgb]{0.50,0.50,1.00}98.0\textsubscript{{\tiny1.4}} & \cellcolor[rgb]{0.50,0.50,1.00}102.7\textsubscript{{\tiny0.8}} & \cellcolor[rgb]{0.50,0.50,1.00}73.6\textsubscript{{\tiny0.4}} & \cellcolor[rgb]{0.50,0.50,1.00}91.3\textsubscript{{\tiny1.0}} & \cellcolor[rgb]{0.81,0.81,1.00}38.1 & \cellcolor[rgb]{0.65,0.65,1.00}37.4 & \cellcolor[rgb]{0.50,0.50,1.00}40.2 & \cellcolor[rgb]{0.50,0.50,1.00}42.6 & \cellcolor[rgb]{0.50,0.50,1.00}37.7 & \cellcolor[rgb]{0.60,0.60,1.00}36.1 & \cellcolor[rgb]{0.50,0.50,1.00}37.3\\
CV & \cellcolor[rgb]{0.50,0.50,1.00}121.8\textsubscript{{\tiny0.3}} & \cellcolor[rgb]{0.50,0.50,1.00}76.2\textsubscript{{\tiny0.2}} & \cellcolor[rgb]{0.50,0.50,1.00}98.9\textsubscript{{\tiny0.6}} & \cellcolor[rgb]{0.50,0.50,1.00}93.4\textsubscript{{\tiny0.4}} & \cellcolor[rgb]{0.50,0.50,1.00}114.7\textsubscript{{\tiny1.6}} & \cellcolor[rgb]{0.50,0.50,1.00}80.8\textsubscript{{\tiny0.7}} & \cellcolor[rgb]{0.50,0.50,1.00}89.8\textsubscript{{\tiny0.6}} & \cellcolor[rgb]{0.50,0.50,1.00}109.6\textsubscript{{\tiny0.5}} & \cellcolor[rgb]{0.50,0.50,1.00}71.6\textsubscript{{\tiny0.5}} & \cellcolor[rgb]{0.50,0.50,1.00}97.5\textsubscript{{\tiny1.1}} & \cellcolor[rgb]{0.50,0.50,1.00}94.5\textsubscript{{\tiny0.6}} & \cellcolor[rgb]{0.57,0.57,1.00}70.4\textsubscript{{\tiny0.4}} & \cellcolor[rgb]{0.54,0.54,1.00}67.2\textsubscript{{\tiny0.4}} & \cellcolor[rgb]{0.50,0.50,1.00}82.3\textsubscript{{\tiny0.3}} & \cellcolor[rgb]{0.77,0.77,1.00}40.8 & \cellcolor[rgb]{0.54,0.54,1.00}42.2 & \cellcolor[rgb]{0.50,0.50,1.00}37.1 & \cellcolor[rgb]{0.50,0.50,1.00}37.3 & \cellcolor[rgb]{0.50,0.50,1.00}41.5 & \cellcolor[rgb]{0.52,0.52,1.00}39.4 & \cellcolor[rgb]{0.50,0.50,1.00}35.4\\
\rowcolor{gray!25}
Chat & 30.4\textsubscript{{\tiny0.3}} & 30.7\textsubscript{{\tiny0.0}} & 29.6\textsubscript{{\tiny0.1}} & 27.3\textsubscript{{\tiny0.1}} & 29.8\textsubscript{{\tiny0.3}} & 33.3\textsubscript{{\tiny0.2}} & 29.6\textsubscript{{\tiny0.1}} & 32.6\textsubscript{{\tiny0.1}} & 34.7\textsubscript{{\tiny0.3}} & 34.7\textsubscript{{\tiny0.3}} & 35.3\textsubscript{{\tiny0.2}} & 37.9\textsubscript{{\tiny0.3}} & 35.2\textsubscript{{\tiny0.1}} & 35.9\textsubscript{{\tiny0.3}} & 27.9 & 22.0 & 15.3 & 16.9 & 18.0 & 20.2 & 17.0\\
Chat+CPT & \cellcolor[rgb]{0.95,0.95,1.00}33.7\textsubscript{{\tiny0.2}} & \cellcolor[rgb]{0.97,0.97,1.00}32.2\textsubscript{{\tiny0.2}} & \cellcolor[rgb]{1.00,0.99,0.99}29.4\textsubscript{{\tiny0.1}} & \cellcolor[rgb]{1.00,1.00,1.00}27.3\textsubscript{{\tiny0.0}} & \cellcolor[rgb]{0.94,0.94,1.00}33.5\textsubscript{{\tiny0.0}} & \cellcolor[rgb]{1.00,1.00,1.00}33.3\textsubscript{{\tiny0.1}} & \cellcolor[rgb]{1.00,0.97,0.97}28.2\textsubscript{{\tiny0.0}} & \cellcolor[rgb]{0.97,0.97,1.00}34.8\textsubscript{{\tiny0.4}} & \cellcolor[rgb]{1.00,1.00,1.00}34.6\textsubscript{{\tiny0.3}} & \cellcolor[rgb]{1.00,0.99,0.99}33.7\textsubscript{{\tiny0.2}} & \cellcolor[rgb]{1.00,0.99,0.99}34.2\textsubscript{{\tiny0.1}} & \cellcolor[rgb]{1.00,0.97,0.97}35.3\textsubscript{{\tiny0.1}} & \cellcolor[rgb]{1.00,0.99,0.99}34.3\textsubscript{{\tiny0.1}} & \cellcolor[rgb]{1.00,0.99,0.99}34.9\textsubscript{{\tiny0.1}} & \cellcolor[rgb]{0.98,0.98,1.00}28.8 & \cellcolor[rgb]{0.96,0.96,1.00}23.8 & \cellcolor[rgb]{0.99,0.99,1.00}15.5 & \cellcolor[rgb]{0.99,0.99,1.00}17.2 & \cellcolor[rgb]{0.95,0.95,1.00}19.7 & \cellcolor[rgb]{0.98,0.98,1.00}21.0 & \cellcolor[rgb]{0.99,0.99,1.00}17.5\\
Chat+VE & \cellcolor[rgb]{0.50,0.50,1.00}129.1\textsubscript{{\tiny0.7}} & \cellcolor[rgb]{0.50,0.50,1.00}68.3\textsubscript{{\tiny0.0}} & \cellcolor[rgb]{0.50,0.50,1.00}103.6\textsubscript{{\tiny0.6}} & \cellcolor[rgb]{0.50,0.50,1.00}94.8\textsubscript{{\tiny0.3}} & \cellcolor[rgb]{0.50,0.50,1.00}104.3\textsubscript{{\tiny0.7}} & \cellcolor[rgb]{0.50,0.50,1.00}80.6\textsubscript{{\tiny0.7}} & \cellcolor[rgb]{0.50,0.50,1.00}99.4\textsubscript{{\tiny0.4}} & \cellcolor[rgb]{0.50,0.50,1.00}102.0\textsubscript{{\tiny1.8}} & \cellcolor[rgb]{0.50,0.50,1.00}73.3\textsubscript{{\tiny0.6}} & \cellcolor[rgb]{0.50,0.50,1.00}97.6\textsubscript{{\tiny0.0}} & \cellcolor[rgb]{0.50,0.50,1.00}98.6\textsubscript{{\tiny0.8}} & \cellcolor[rgb]{0.50,0.50,1.00}81.1\textsubscript{{\tiny0.9}} & \cellcolor[rgb]{0.50,0.50,1.00}70.9\textsubscript{{\tiny0.7}} & \cellcolor[rgb]{0.50,0.50,1.00}80.2\textsubscript{{\tiny0.9}} & \cellcolor[rgb]{0.72,0.72,1.00}43.2 & \cellcolor[rgb]{0.65,0.65,1.00}37.5 & \cellcolor[rgb]{0.50,0.50,1.00}39.1 & \cellcolor[rgb]{0.50,0.50,1.00}37.6 & \cellcolor[rgb]{0.50,0.50,1.00}37.3 & \cellcolor[rgb]{0.50,0.50,1.00}40.2 & \cellcolor[rgb]{0.50,0.50,1.00}38.7\\
ElChat & \cellcolor[rgb]{0.50,0.50,1.00}117.5\textsubscript{{\tiny0.4}} & \cellcolor[rgb]{0.50,0.50,1.00}76.8\textsubscript{{\tiny0.8}} & \cellcolor[rgb]{0.50,0.50,1.00}97.6\textsubscript{{\tiny5.0}} & \cellcolor[rgb]{0.50,0.50,1.00}95.2\textsubscript{{\tiny0.4}} & \cellcolor[rgb]{0.50,0.50,1.00}104.2\textsubscript{{\tiny0.2}} & \cellcolor[rgb]{0.50,0.50,1.00}81.1\textsubscript{{\tiny0.8}} & \cellcolor[rgb]{0.50,0.50,1.00}99.7\textsubscript{{\tiny1.0}} & \cellcolor[rgb]{0.50,0.50,1.00}113.3\textsubscript{{\tiny0.5}} & \cellcolor[rgb]{0.50,0.50,1.00}70.9\textsubscript{{\tiny0.5}} & \cellcolor[rgb]{0.50,0.50,1.00}94.1\textsubscript{{\tiny0.4}} & \cellcolor[rgb]{0.50,0.50,1.00}93.0\textsubscript{{\tiny0.9}} & \cellcolor[rgb]{0.50,0.50,1.00}101.1\textsubscript{{\tiny0.2}} & \cellcolor[rgb]{0.50,0.50,1.00}74.2\textsubscript{{\tiny0.8}} & \cellcolor[rgb]{0.50,0.50,1.00}91.6\textsubscript{{\tiny0.9}} & \cellcolor[rgb]{0.82,0.82,1.00}38.0 & \cellcolor[rgb]{0.54,0.54,1.00}42.0 & \cellcolor[rgb]{0.50,0.50,1.00}37.2 & \cellcolor[rgb]{0.50,0.50,1.00}37.4 & \cellcolor[rgb]{0.50,0.50,1.00}37.7 & \cellcolor[rgb]{0.51,0.51,1.00}39.9 & \cellcolor[rgb]{0.50,0.50,1.00}38.4\\

 \bottomrule
\end{tabular}
}
\end{center}
\label{tab:speed_performance_qwen25}
\end{table*}

\renewcommand*{\arraystretch}{1.0}
\begin{table*}[t]
\caption{Llama 3.1 8B inference speedup measured by the number of tokens generated per second.
Darker \textcolor{blue}{blue} and  \textcolor{red}{red} shades indicate higher positive and negative relative performance change over \colorbox{gray!25}{Chat} per language and task, respectively.
(L) stands for linear merging.}
\begin{center}
\resizebox{\linewidth}{!}{%
\begin{tabular}{llllllll|lllllll|lllllll}
\toprule
 & \multicolumn{7}{c|}{\textbf{Target \textsc{sum}}} & \multicolumn{7}{c|}{\textbf{English $\rightarrow$ Target \textsc{mt}}} & \multicolumn{7}{c}{\textbf{Target \textsc{mc}}}\\  
 & \multicolumn{1}{c}{\footnotesize am} & \multicolumn{1}{c}{\footnotesize bn} & \multicolumn{1}{c}{\footnotesize my} & \multicolumn{1}{c}{\footnotesize gu} & \multicolumn{1}{c}{\footnotesize si} & \multicolumn{1}{c}{\footnotesize ta} & \multicolumn{1}{c|}{\footnotesize te} 
 & \multicolumn{1}{c}{\footnotesize am} & \multicolumn{1}{c}{\footnotesize bn} & \multicolumn{1}{c}{\footnotesize my} & \multicolumn{1}{c}{\footnotesize gu} & \multicolumn{1}{c}{\footnotesize si} & \multicolumn{1}{c}{\footnotesize ta} & \multicolumn{1}{c|}{\footnotesize te} 
 & \multicolumn{1}{c}{\footnotesize am} & \multicolumn{1}{c}{\footnotesize bn} & \multicolumn{1}{c}{\footnotesize my} & \multicolumn{1}{c}{\footnotesize gu} & \multicolumn{1}{c}{\footnotesize si} & \multicolumn{1}{c}{\footnotesize ta} & \multicolumn{1}{c}{\footnotesize te}\\

\midrule
Base & \cellcolor[rgb]{1.00,1.00,1.00}19.3\textsubscript{{\tiny0.0}} & \cellcolor[rgb]{1.00,0.99,0.99}21.7\textsubscript{{\tiny0.0}} & \cellcolor[rgb]{1.00,1.00,1.00}18.3\textsubscript{{\tiny0.1}} & \cellcolor[rgb]{1.00,1.00,1.00}19.7\textsubscript{{\tiny0.0}} & \cellcolor[rgb]{1.00,0.99,0.99}20.8\textsubscript{{\tiny0.0}} & \cellcolor[rgb]{1.00,0.99,0.99}22.4\textsubscript{{\tiny0.1}} & \cellcolor[rgb]{1.00,1.00,1.00}18.8\textsubscript{{\tiny0.0}} & \cellcolor[rgb]{0.99,0.99,1.00}31.9\textsubscript{{\tiny0.0}} & \cellcolor[rgb]{0.99,0.99,1.00}33.3\textsubscript{{\tiny0.1}} & \cellcolor[rgb]{0.99,0.99,1.00}28.5\textsubscript{{\tiny0.1}} & \cellcolor[rgb]{0.99,0.99,1.00}32.0\textsubscript{{\tiny0.0}} & \cellcolor[rgb]{1.00,1.00,1.00}31.9\textsubscript{{\tiny0.0}} & \cellcolor[rgb]{0.99,0.99,1.00}32.1\textsubscript{{\tiny0.0}} & \cellcolor[rgb]{0.99,0.99,1.00}31.5\textsubscript{{\tiny0.2}} & \cellcolor[rgb]{0.99,0.99,1.00}15.4 & \cellcolor[rgb]{0.99,0.99,1.00}17.7 & \cellcolor[rgb]{0.99,0.99,1.00}10.5 & \cellcolor[rgb]{0.99,0.99,1.00}13.5 & \cellcolor[rgb]{0.99,0.99,1.00}13.1 & \cellcolor[rgb]{0.99,0.99,1.00}14.6 & \cellcolor[rgb]{0.99,0.99,1.00}12.7\\
Base+CPT & \cellcolor[rgb]{1.00,1.00,1.00}19.3\textsubscript{{\tiny0.0}} & \cellcolor[rgb]{1.00,1.00,1.00}21.9\textsubscript{{\tiny0.0}} & \cellcolor[rgb]{0.99,0.99,1.00}18.5\textsubscript{{\tiny0.0}} & \cellcolor[rgb]{1.00,0.99,0.99}19.4\textsubscript{{\tiny0.1}} & \cellcolor[rgb]{1.00,1.00,1.00}21.2\textsubscript{{\tiny0.0}} & \cellcolor[rgb]{0.99,0.99,1.00}23.2\textsubscript{{\tiny0.0}} & \cellcolor[rgb]{1.00,1.00,1.00}18.9\textsubscript{{\tiny0.0}} & \cellcolor[rgb]{1.00,1.00,1.00}31.6\textsubscript{{\tiny0.0}} & \cellcolor[rgb]{1.00,1.00,1.00}33.1\textsubscript{{\tiny0.4}} & \cellcolor[rgb]{1.00,1.00,1.00}28.4\textsubscript{{\tiny0.2}} & \cellcolor[rgb]{1.00,1.00,1.00}31.6\textsubscript{{\tiny0.1}} & \cellcolor[rgb]{1.00,1.00,1.00}31.6\textsubscript{{\tiny0.1}} & \cellcolor[rgb]{1.00,1.00,1.00}31.8\textsubscript{{\tiny0.1}} & \cellcolor[rgb]{0.99,0.99,1.00}31.4\textsubscript{{\tiny0.2}} & \cellcolor[rgb]{0.99,0.99,1.00}15.3 & \cellcolor[rgb]{0.98,0.98,1.00}17.9 & \cellcolor[rgb]{0.99,0.99,1.00}10.5 & \cellcolor[rgb]{0.99,0.99,1.00}13.6 & \cellcolor[rgb]{0.99,0.99,1.00}13.1 & \cellcolor[rgb]{0.99,0.99,1.00}14.7 & \cellcolor[rgb]{0.99,0.99,1.00}12.7\\
Base+VE & \cellcolor[rgb]{0.50,0.50,1.00}206.2\textsubscript{{\tiny3.3}} & \cellcolor[rgb]{0.50,0.50,1.00}78.5\textsubscript{{\tiny0.5}} & \cellcolor[rgb]{0.50,0.50,1.00}114.7\textsubscript{{\tiny0.3}} & \cellcolor[rgb]{0.50,0.50,1.00}100.6\textsubscript{{\tiny0.5}} & \cellcolor[rgb]{0.50,0.50,1.00}121.0\textsubscript{{\tiny0.4}} & \cellcolor[rgb]{0.50,0.50,1.00}91.4\textsubscript{{\tiny0.4}} & \cellcolor[rgb]{0.50,0.50,1.00}105.4\textsubscript{{\tiny0.2}} & \cellcolor[rgb]{0.50,0.50,1.00}141.5\textsubscript{{\tiny0.8}} & \cellcolor[rgb]{0.50,0.50,1.00}78.9\textsubscript{{\tiny0.5}} & \cellcolor[rgb]{0.50,0.50,1.00}111.6\textsubscript{{\tiny0.4}} & \cellcolor[rgb]{0.50,0.50,1.00}98.2\textsubscript{{\tiny0.8}} & \cellcolor[rgb]{0.50,0.50,1.00}122.5\textsubscript{{\tiny1.2}} & \cellcolor[rgb]{0.50,0.50,1.00}84.9\textsubscript{{\tiny8.8}} & \cellcolor[rgb]{0.50,0.50,1.00}98.3\textsubscript{{\tiny0.8}} & \cellcolor[rgb]{0.50,0.50,1.00}44.7 & \cellcolor[rgb]{0.50,0.50,1.00}39.7 & \cellcolor[rgb]{0.50,0.50,1.00}36.9 & \cellcolor[rgb]{0.50,0.50,1.00}36.9 & \cellcolor[rgb]{0.50,0.50,1.00}37.0 & \cellcolor[rgb]{0.50,0.50,1.00}37.8 & \cellcolor[rgb]{0.50,0.50,1.00}38.7\\
CV & \cellcolor[rgb]{0.50,0.50,1.00}224.3\textsubscript{{\tiny2.9}} & \cellcolor[rgb]{0.50,0.50,1.00}78.2\textsubscript{{\tiny0.1}} & \cellcolor[rgb]{0.50,0.50,1.00}120.2\textsubscript{{\tiny0.2}} & \cellcolor[rgb]{0.50,0.50,1.00}107.6\textsubscript{{\tiny0.6}} & \cellcolor[rgb]{0.50,0.50,1.00}132.7\textsubscript{{\tiny0.1}} & \cellcolor[rgb]{0.50,0.50,1.00}90.1\textsubscript{{\tiny1.1}} & \cellcolor[rgb]{0.50,0.50,1.00}103.5\textsubscript{{\tiny0.2}} & \cellcolor[rgb]{0.50,0.50,1.00}185.1\textsubscript{{\tiny0.8}} & \cellcolor[rgb]{0.50,0.50,1.00}84.8\textsubscript{{\tiny0.4}} & \cellcolor[rgb]{0.50,0.50,1.00}123.5\textsubscript{{\tiny1.0}} & \cellcolor[rgb]{0.50,0.50,1.00}135.6\textsubscript{{\tiny1.4}} & \cellcolor[rgb]{0.50,0.50,1.00}123.2\textsubscript{{\tiny1.0}} & \cellcolor[rgb]{0.50,0.50,1.00}91.8\textsubscript{{\tiny1.1}} & \cellcolor[rgb]{0.50,0.50,1.00}109.5\textsubscript{{\tiny1.2}} & \cellcolor[rgb]{0.50,0.50,1.00}44.5 & \cellcolor[rgb]{0.54,0.54,1.00}33.0 & \cellcolor[rgb]{0.50,0.50,1.00}28.6 & \cellcolor[rgb]{0.50,0.50,1.00}33.4 & \cellcolor[rgb]{0.50,0.50,1.00}34.7 & \cellcolor[rgb]{0.50,0.50,1.00}31.0 & \cellcolor[rgb]{0.50,0.50,1.00}31.4\\
\rowcolor{gray!25}
Chat & 19.2\textsubscript{{\tiny0.0}} & 22.0\textsubscript{{\tiny0.0}} & 18.3\textsubscript{{\tiny0.0}} & 19.6\textsubscript{{\tiny0.0}} & 21.1\textsubscript{{\tiny0.1}} & 22.9\textsubscript{{\tiny0.0}} & 18.9\textsubscript{{\tiny0.0}} & 31.6\textsubscript{{\tiny0.1}} & 33.0\textsubscript{{\tiny0.2}} & 28.2\textsubscript{{\tiny0.0}} & 31.5\textsubscript{{\tiny0.1}} & 31.7\textsubscript{{\tiny0.0}} & 31.6\textsubscript{{\tiny0.1}} & 31.0\textsubscript{{\tiny0.1}} & 15.0 & 17.3 & 10.3 & 13.2 & 12.9 & 14.4 & 12.5\\
Chat+CPT & \cellcolor[rgb]{1.00,0.99,0.99}18.9\textsubscript{{\tiny0.0}} & \cellcolor[rgb]{1.00,1.00,1.00}21.8\textsubscript{{\tiny0.0}} & \cellcolor[rgb]{1.00,1.00,1.00}18.4\textsubscript{{\tiny0.0}} & \cellcolor[rgb]{1.00,1.00,1.00}19.5\textsubscript{{\tiny0.0}} & \cellcolor[rgb]{1.00,1.00,1.00}21.1\textsubscript{{\tiny0.0}} & \cellcolor[rgb]{0.99,0.99,1.00}23.2\textsubscript{{\tiny0.1}} & \cellcolor[rgb]{1.00,1.00,1.00}18.9\textsubscript{{\tiny0.0}} & \cellcolor[rgb]{1.00,1.00,1.00}31.8\textsubscript{{\tiny0.1}} & \cellcolor[rgb]{0.99,0.99,1.00}33.4\textsubscript{{\tiny0.1}} & \cellcolor[rgb]{1.00,1.00,1.00}28.3\textsubscript{{\tiny0.0}} & \cellcolor[rgb]{0.99,0.99,1.00}31.7\textsubscript{{\tiny0.2}} & \cellcolor[rgb]{1.00,0.99,0.99}31.3\textsubscript{{\tiny0.3}} & \cellcolor[rgb]{0.99,0.99,1.00}31.9\textsubscript{{\tiny0.1}} & \cellcolor[rgb]{1.00,1.00,1.00}31.3\textsubscript{{\tiny0.1}} & \cellcolor[rgb]{1.00,0.99,0.99}14.9 & \cellcolor[rgb]{1.00,1.00,1.00}17.2 & \cellcolor[rgb]{0.99,0.99,1.00}10.4 & \cellcolor[rgb]{1.00,1.00,1.00}13.3 & \cellcolor[rgb]{1.00,1.00,1.00}13.0 & \cellcolor[rgb]{1.00,1.00,1.00}14.4 & \cellcolor[rgb]{0.99,0.99,1.00}12.7\\
Chat+VE & \cellcolor[rgb]{0.50,0.50,1.00}218.9\textsubscript{{\tiny3.1}} & \cellcolor[rgb]{0.50,0.50,1.00}78.6\textsubscript{{\tiny0.3}} & \cellcolor[rgb]{0.50,0.50,1.00}120.2\textsubscript{{\tiny0.7}} & \cellcolor[rgb]{0.50,0.50,1.00}107.2\textsubscript{{\tiny0.5}} & \cellcolor[rgb]{0.50,0.50,1.00}135.1\textsubscript{{\tiny0.4}} & \cellcolor[rgb]{0.50,0.50,1.00}93.2\textsubscript{{\tiny0.6}} & \cellcolor[rgb]{0.50,0.50,1.00}99.3\textsubscript{{\tiny0.5}} & \cellcolor[rgb]{0.50,0.50,1.00}168.8\textsubscript{{\tiny3.7}} & \cellcolor[rgb]{0.50,0.50,1.00}79.7\textsubscript{{\tiny0.2}} & \cellcolor[rgb]{0.50,0.50,1.00}115.5\textsubscript{{\tiny1.2}} & \cellcolor[rgb]{0.50,0.50,1.00}100.4\textsubscript{{\tiny1.5}} & \cellcolor[rgb]{0.50,0.50,1.00}124.4\textsubscript{{\tiny0.9}} & \cellcolor[rgb]{0.50,0.50,1.00}89.0\textsubscript{{\tiny0.5}} & \cellcolor[rgb]{0.50,0.50,1.00}96.9\textsubscript{{\tiny0.4}} & \cellcolor[rgb]{0.50,0.50,1.00}44.3 & \cellcolor[rgb]{0.50,0.50,1.00}39.1 & \cellcolor[rgb]{0.50,0.50,1.00}34.7 & \cellcolor[rgb]{0.50,0.50,1.00}39.5 & \cellcolor[rgb]{0.50,0.50,1.00}40.6 & \cellcolor[rgb]{0.50,0.50,1.00}37.6 & \cellcolor[rgb]{0.50,0.50,1.00}36.9\\
ElChat & \cellcolor[rgb]{0.50,0.50,1.00}233.9\textsubscript{{\tiny1.9}} & \cellcolor[rgb]{0.50,0.50,1.00}75.8\textsubscript{{\tiny0.5}} & \cellcolor[rgb]{0.50,0.50,1.00}118.2\textsubscript{{\tiny2.6}} & \cellcolor[rgb]{0.50,0.50,1.00}105.5\textsubscript{{\tiny0.5}} & \cellcolor[rgb]{0.50,0.50,1.00}126.3\textsubscript{{\tiny0.6}} & \cellcolor[rgb]{0.50,0.50,1.00}90.3\textsubscript{{\tiny1.4}} & \cellcolor[rgb]{0.50,0.50,1.00}101.7\textsubscript{{\tiny0.4}} & \cellcolor[rgb]{0.50,0.50,1.00}212.0\textsubscript{{\tiny3.4}} & \cellcolor[rgb]{0.50,0.50,1.00}81.3\textsubscript{{\tiny0.8}} & \cellcolor[rgb]{0.50,0.50,1.00}113.0\textsubscript{{\tiny1.2}} & \cellcolor[rgb]{0.50,0.50,1.00}107.7\textsubscript{{\tiny0.4}} & \cellcolor[rgb]{0.50,0.50,1.00}124.1\textsubscript{{\tiny0.7}} & \cellcolor[rgb]{0.50,0.50,1.00}89.8\textsubscript{{\tiny0.8}} & \cellcolor[rgb]{0.50,0.50,1.00}107.2\textsubscript{{\tiny0.5}} & \cellcolor[rgb]{0.50,0.50,1.00}44.5 & \cellcolor[rgb]{0.50,0.50,1.00}37.9 & \cellcolor[rgb]{0.50,0.50,1.00}34.1 & \cellcolor[rgb]{0.50,0.50,1.00}38.9 & \cellcolor[rgb]{0.50,0.50,1.00}38.0 & \cellcolor[rgb]{0.50,0.50,1.00}37.6 & \cellcolor[rgb]{0.50,0.50,1.00}37.3\\
\midrule
ElChat \textbackslash Merge & \cellcolor[rgb]{0.50,0.50,1.00}201.3\textsubscript{{\tiny2.6}} & \cellcolor[rgb]{0.50,0.50,1.00}79.3\textsubscript{{\tiny0.0}} & \cellcolor[rgb]{0.50,0.50,1.00}121.6\textsubscript{{\tiny0.4}} & \cellcolor[rgb]{0.50,0.50,1.00}109.1\textsubscript{{\tiny0.2}} & \cellcolor[rgb]{0.50,0.50,1.00}132.1\textsubscript{{\tiny1.1}} & \cellcolor[rgb]{0.50,0.50,1.00}88.9\textsubscript{{\tiny0.4}} & \cellcolor[rgb]{0.50,0.50,1.00}103.8\textsubscript{{\tiny0.2}} & \cellcolor[rgb]{0.50,0.50,1.00}164.1\textsubscript{{\tiny2.2}} & \cellcolor[rgb]{0.50,0.50,1.00}76.3\textsubscript{{\tiny0.8}} & \cellcolor[rgb]{0.50,0.50,1.00}111.4\textsubscript{{\tiny1.5}} & \cellcolor[rgb]{0.50,0.50,1.00}84.2\textsubscript{{\tiny1.1}} & \cellcolor[rgb]{0.50,0.50,1.00}102.0\textsubscript{{\tiny2.7}} & \cellcolor[rgb]{0.50,0.50,1.00}84.3\textsubscript{{\tiny0.4}} & \cellcolor[rgb]{0.50,0.50,1.00}94.5\textsubscript{{\tiny0.3}} & \cellcolor[rgb]{0.50,0.50,1.00}42.2 & \cellcolor[rgb]{0.50,0.50,1.00}39.1 & \cellcolor[rgb]{0.50,0.50,1.00}34.4 & \cellcolor[rgb]{0.50,0.50,1.00}40.2 & \cellcolor[rgb]{0.50,0.50,1.00}39.9 & \cellcolor[rgb]{0.50,0.50,1.00}37.3 & \cellcolor[rgb]{0.50,0.50,1.00}37.6\\
ElChat \textbackslash Copy & \cellcolor[rgb]{0.50,0.50,1.00}235.0\textsubscript{{\tiny4.3}} & \cellcolor[rgb]{0.50,0.50,1.00}79.2\textsubscript{{\tiny0.1}} & \cellcolor[rgb]{0.50,0.50,1.00}118.6\textsubscript{{\tiny0.2}} & \cellcolor[rgb]{0.50,0.50,1.00}107.3\textsubscript{{\tiny0.2}} & \cellcolor[rgb]{0.50,0.50,1.00}123.7\textsubscript{{\tiny0.7}} & \cellcolor[rgb]{0.50,0.50,1.00}85.6\textsubscript{{\tiny1.5}} & \cellcolor[rgb]{0.50,0.50,1.00}105.0\textsubscript{{\tiny0.6}} & \cellcolor[rgb]{0.50,0.50,1.00}207.3\textsubscript{{\tiny1.6}} & \cellcolor[rgb]{0.50,0.50,1.00}83.7\textsubscript{{\tiny1.4}} & \cellcolor[rgb]{0.50,0.50,1.00}113.2\textsubscript{{\tiny0.7}} & \cellcolor[rgb]{0.50,0.50,1.00}112.1\textsubscript{{\tiny0.3}} & \cellcolor[rgb]{0.50,0.50,1.00}129.2\textsubscript{{\tiny0.5}} & \cellcolor[rgb]{0.50,0.50,1.00}92.4\textsubscript{{\tiny0.8}} & \cellcolor[rgb]{0.50,0.50,1.00}110.9\textsubscript{{\tiny2.0}} & \cellcolor[rgb]{0.50,0.50,1.00}46.7 & \cellcolor[rgb]{0.50,0.50,1.00}39.6 & \cellcolor[rgb]{0.50,0.50,1.00}34.3 & \cellcolor[rgb]{0.50,0.50,1.00}39.5 & \cellcolor[rgb]{0.50,0.50,1.00}37.5 & \cellcolor[rgb]{0.50,0.50,1.00}36.4 & \cellcolor[rgb]{0.50,0.50,1.00}37.8\\
ElChat~(L) & \cellcolor[rgb]{0.50,0.50,1.00}230.7\textsubscript{{\tiny2.4}} & \cellcolor[rgb]{0.50,0.50,1.00}78.7\textsubscript{{\tiny0.2}} & \cellcolor[rgb]{0.50,0.50,1.00}119.6\textsubscript{{\tiny0.6}} & \cellcolor[rgb]{0.50,0.50,1.00}106.3\textsubscript{{\tiny0.8}} & \cellcolor[rgb]{0.50,0.50,1.00}135.9\textsubscript{{\tiny0.2}} & \cellcolor[rgb]{0.50,0.50,1.00}92.6\textsubscript{{\tiny0.9}} & \cellcolor[rgb]{0.50,0.50,1.00}103.9\textsubscript{{\tiny0.8}} & \cellcolor[rgb]{0.50,0.50,1.00}211.4\textsubscript{{\tiny3.1}} & \cellcolor[rgb]{0.50,0.50,1.00}78.9\textsubscript{{\tiny0.7}} & \cellcolor[rgb]{0.50,0.50,1.00}111.2\textsubscript{{\tiny1.1}} & \cellcolor[rgb]{0.50,0.50,1.00}106.0\textsubscript{{\tiny0.7}} & \cellcolor[rgb]{0.50,0.50,1.00}122.2\textsubscript{{\tiny0.5}} & \cellcolor[rgb]{0.50,0.50,1.00}88.4\textsubscript{{\tiny1.9}} & \cellcolor[rgb]{0.50,0.50,1.00}105.8\textsubscript{{\tiny0.5}} & \cellcolor[rgb]{0.50,0.50,1.00}44.9 & \cellcolor[rgb]{0.50,0.50,1.00}39.2 & \cellcolor[rgb]{0.50,0.50,1.00}34.3 & \cellcolor[rgb]{0.50,0.50,1.00}39.4 & \cellcolor[rgb]{0.50,0.50,1.00}40.8 & \cellcolor[rgb]{0.50,0.50,1.00}37.7 & \cellcolor[rgb]{0.50,0.50,1.00}37.8\\

 \bottomrule
\end{tabular}
}
\end{center}
\label{tab:speed_performance_llama31}
\end{table*}

\renewcommand*{\arraystretch}{1.0}
\begin{table*}[t]
\caption{Qwen3 14B inference speedup measured by the number of tokens generated per second.
Darker \textcolor{blue}{blue} and  \textcolor{red}{red} shades indicate higher positive and negative relative performance change over \colorbox{gray!25}{Chat} per language and task, respectively. }
\begin{center}
\begin{tabular}{llll|lll|lll}
\toprule
 & \multicolumn{3}{c|}{\textbf{Target \textsc{sum}}} & \multicolumn{3}{c|}{\textbf{English $\rightarrow$ Target \textsc{mt}}} & \multicolumn{3}{c}{\textbf{Target \textsc{mc}}}\\  
 & \multicolumn{1}{c}{\footnotesize am} & \multicolumn{1}{c}{\footnotesize bn} & \multicolumn{1}{c|}{\footnotesize te} 
 & \multicolumn{1}{c}{\footnotesize am} & \multicolumn{1}{c}{\footnotesize bn} & \multicolumn{1}{c|}{\footnotesize te} 
 & \multicolumn{1}{c}{\footnotesize am} & \multicolumn{1}{c}{\footnotesize bn} & \multicolumn{1}{c}{\footnotesize te}\\

\midrule

Base & \cellcolor[rgb]{1.00,0.98,0.98}14.6\textsubscript{{\tiny0.1}} & \cellcolor[rgb]{1.00,0.96,0.96}13.5\textsubscript{{\tiny0.1}} & \cellcolor[rgb]{1.00,0.97,0.97}13.2\textsubscript{{\tiny0.1}} & \cellcolor[rgb]{1.00,0.99,0.99}15.4\textsubscript{{\tiny0.1}} & \cellcolor[rgb]{1.00,0.99,0.99}15.6\textsubscript{{\tiny0.2}} & \cellcolor[rgb]{1.00,0.99,0.99}15.6\textsubscript{{\tiny0.1}} & \cellcolor[rgb]{1.00,0.99,0.99}17.7 & \cellcolor[rgb]{1.00,1.00,1.00}16.3 & \cellcolor[rgb]{0.99,0.99,1.00}12.6\\
Base+CPT & \cellcolor[rgb]{1.00,0.98,0.98}14.6\textsubscript{{\tiny0.0}} & \cellcolor[rgb]{1.00,1.00,1.00}14.6\textsubscript{{\tiny0.1}} & \cellcolor[rgb]{1.00,0.99,0.99}13.8\textsubscript{{\tiny0.1}} & \cellcolor[rgb]{1.00,0.99,0.99}15.5\textsubscript{{\tiny0.2}} & \cellcolor[rgb]{0.99,0.99,1.00}16.1\textsubscript{{\tiny0.2}} & \cellcolor[rgb]{1.00,0.99,0.99}15.4\textsubscript{{\tiny0.2}} & \cellcolor[rgb]{1.00,0.99,0.99}17.5 & \cellcolor[rgb]{0.99,0.99,1.00}16.6 & \cellcolor[rgb]{0.99,0.99,1.00}12.5\\
Base+VE & \cellcolor[rgb]{0.50,0.50,1.00}52.3\textsubscript{{\tiny0.2}} & \cellcolor[rgb]{0.50,0.50,1.00}33.7\textsubscript{{\tiny0.5}} & \cellcolor[rgb]{0.50,0.50,1.00}43.5\textsubscript{{\tiny0.2}} & \cellcolor[rgb]{0.50,0.50,1.00}36.3\textsubscript{{\tiny0.3}} & \cellcolor[rgb]{0.57,0.57,1.00}29.6\textsubscript{{\tiny0.5}} & \cellcolor[rgb]{0.50,0.50,1.00}38.6\textsubscript{{\tiny0.3}} & \cellcolor[rgb]{1.00,1.00,1.00}17.8 & \cellcolor[rgb]{0.96,0.96,1.00}17.5 & \cellcolor[rgb]{0.77,0.77,1.00}17.8\\
CV & \cellcolor[rgb]{0.50,0.50,1.00}56.8\textsubscript{{\tiny0.2}} & \cellcolor[rgb]{0.50,0.50,1.00}34.3\textsubscript{{\tiny0.1}} & \cellcolor[rgb]{0.50,0.50,1.00}44.0\textsubscript{{\tiny0.4}} & \cellcolor[rgb]{0.61,0.61,1.00}27.9\textsubscript{{\tiny0.1}} & \cellcolor[rgb]{0.67,0.67,1.00}26.3\textsubscript{{\tiny0.2}} & \cellcolor[rgb]{0.50,0.50,1.00}35.2\textsubscript{{\tiny0.4}} & \cellcolor[rgb]{1.00,0.97,0.97}17.0 & \cellcolor[rgb]{0.96,0.96,1.00}17.6 & \cellcolor[rgb]{0.79,0.79,1.00}17.5\\
\rowcolor{gray!25}
Chat & 15.2\textsubscript{{\tiny0.2}} & 14.6\textsubscript{{\tiny0.1}} & 14.1\textsubscript{{\tiny0.1}} & 15.7\textsubscript{{\tiny0.1}} & 15.9\textsubscript{{\tiny0.2}} & 15.8\textsubscript{{\tiny0.2}} & 17.9 & 16.3 & 12.3\\
Chat+CPT & \cellcolor[rgb]{1.00,0.99,0.99}14.9\textsubscript{{\tiny0.1}} & \cellcolor[rgb]{1.00,1.00,1.00}14.5\textsubscript{{\tiny0.1}} & \cellcolor[rgb]{1.00,1.00,1.00}14.1\textsubscript{{\tiny0.0}} & \cellcolor[rgb]{0.99,0.99,1.00}16.0\textsubscript{{\tiny0.0}} & \cellcolor[rgb]{1.00,1.00,1.00}15.9\textsubscript{{\tiny0.2}} & \cellcolor[rgb]{1.00,1.00,1.00}15.8\textsubscript{{\tiny0.1}} & \cellcolor[rgb]{1.00,0.98,0.98}17.4 & \cellcolor[rgb]{1.00,0.99,0.99}15.9 & \cellcolor[rgb]{1.00,1.00,1.00}12.3\\
Chat+VE & \cellcolor[rgb]{0.50,0.50,1.00}55.8\textsubscript{{\tiny0.2}} & \cellcolor[rgb]{0.50,0.50,1.00}34.3\textsubscript{{\tiny0.3}} & \cellcolor[rgb]{0.50,0.50,1.00}44.9\textsubscript{{\tiny0.1}} & \cellcolor[rgb]{0.60,0.60,1.00}28.1\textsubscript{{\tiny0.3}} & \cellcolor[rgb]{0.50,0.50,1.00}32.3\textsubscript{{\tiny0.3}} & \cellcolor[rgb]{0.50,0.50,1.00}42.8\textsubscript{{\tiny0.7}} & \cellcolor[rgb]{1.00,0.99,0.99}17.4 & \cellcolor[rgb]{0.96,0.96,1.00}17.6 & \cellcolor[rgb]{0.78,0.78,1.00}17.6\\
ElChat & \cellcolor[rgb]{0.50,0.50,1.00}58.0\textsubscript{{\tiny0.5}} & \cellcolor[rgb]{0.50,0.50,1.00}34.5\textsubscript{{\tiny0.1}} & \cellcolor[rgb]{0.50,0.50,1.00}44.1\textsubscript{{\tiny0.2}} & \cellcolor[rgb]{0.50,0.50,1.00}43.5\textsubscript{{\tiny0.7}} & \cellcolor[rgb]{0.53,0.53,1.00}30.9\textsubscript{{\tiny0.5}} & \cellcolor[rgb]{0.50,0.50,1.00}40.3\textsubscript{{\tiny0.5}} & \cellcolor[rgb]{1.00,0.98,0.98}17.4 & \cellcolor[rgb]{0.96,0.96,1.00}17.6 & \cellcolor[rgb]{0.78,0.78,1.00}17.8\\

 \bottomrule
\end{tabular}
\end{center}
\label{tab:speed_performance_qwen3}
\end{table*}

\subsection{Ratio of Target Language Tokens} \label{appendix:ratio}

\begin{figure*}[t]
\centering
\includegraphics[width=\textwidth]{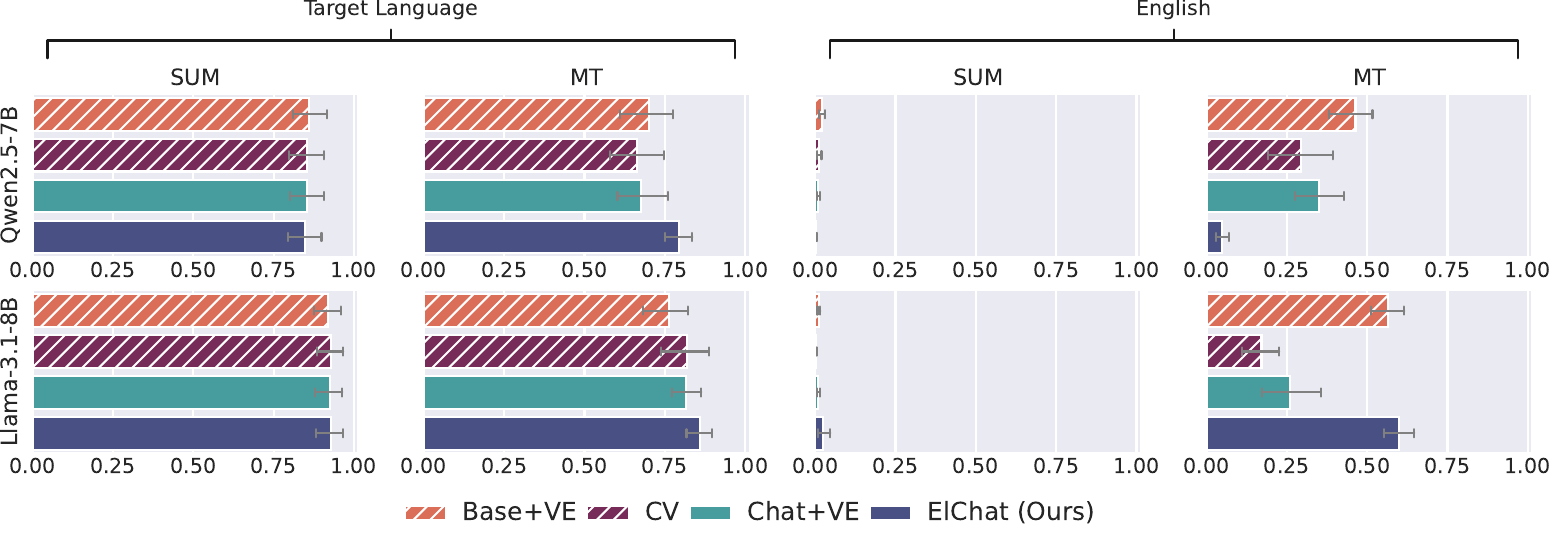}
\caption{Aggregated average ratio of target tokens in output per sample across seven target languages for each model (error bars indicate 95\% confidence interval).
}
\label{fig:ratio}
\end{figure*}

Figure \ref{fig:ratio} shows the aggregated mean ratio of target new tokens in output per sample across seven target languages for each model.

\subsection{Number of Generated Tokens} \label{appendix:tokens}
\begin{figure*}[t]
\centering
\includegraphics[width=\textwidth]{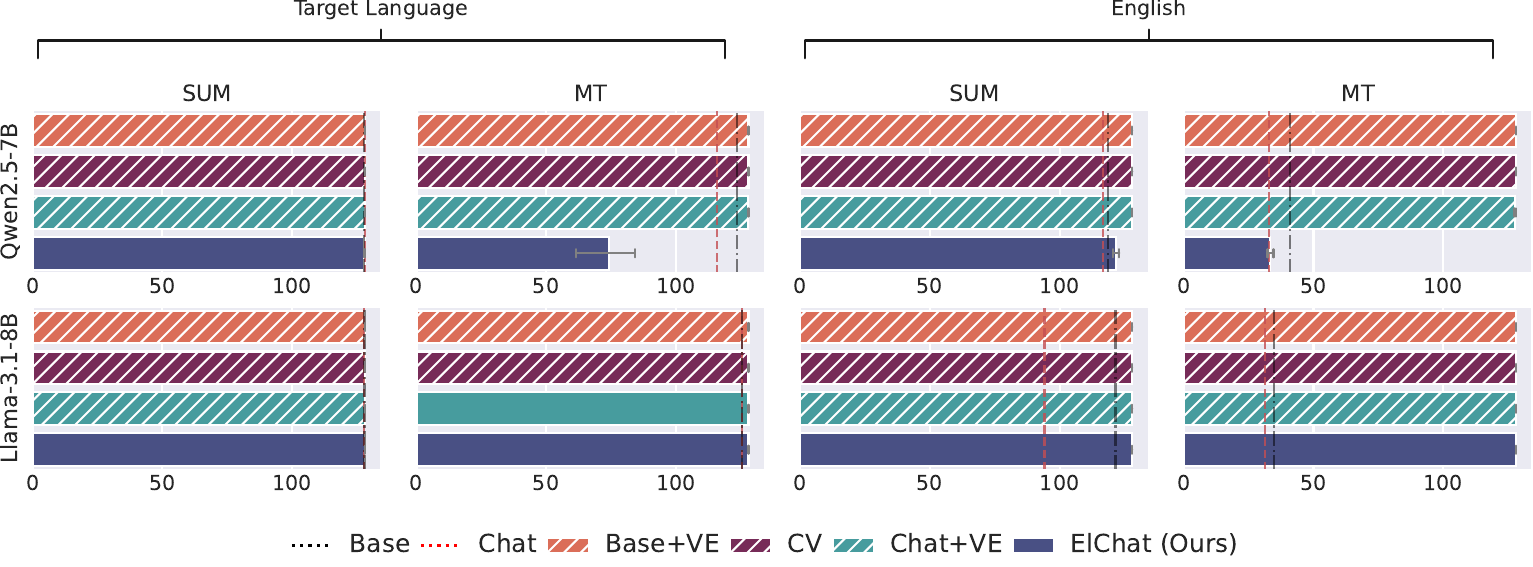}
\caption{Aggregated average number of generated tokens per sample across seven target languages for each model (error bars indicate 95\% confidence interval).
}
\label{fig:tokens}
\end{figure*}

Figure \ref{fig:tokens} shows the aggregated average number of generated tokens per sample across seven target languages for each model.

\newpage
\section{Analysis and Discussion} \label{appendix:analysis}

\subsection{CPT-only vs. VE} \label{appendix:cpt-result}

Comparing the adapted chat models (Chat+VE and ElChat) with the CPT-only model (Chat+CPT) in Tables \ref{tab:task_performance_qwen25} and \ref{tab:task_performance_llama31}, we observe three key trends: (1) Chat+VE generally performs better than Chat+CPT on target language tasks across both models; (2) Chat+CPT often outperforms Chat+VE on \textsc{mt} tasks; and (3) ElChat either matches or surpasses Chat+CPT on nearly all tasks and models, except for Target \textsc{mc} and \textsc{gmmlu} with Qwen2.5 and target-to-English \textsc{mt} with Llama 3.1.  This performance advantage of ElChat is also confirmed for chat and instruction-following tasks (Tables \ref{tab:chat_performance_qwen25}, \ref{tab:chat_performance_llama31}, and \ref{tab:mgsm_aux}), where ElChat substantially outperforms Chat+CPT in almost all cases across languages, tasks, and models.

We also observe a similar trend between Base+VE and the CPT-only adapted base model (Base+CPT) in Tables \ref{tab:task_performance_qwen25} and \ref{tab:task_performance_llama31}:
Base+VE often outperforms Base+CPT in most of the tasks, while Base+CPT excels in target-to-English \textsc{mt} across models.

These results somewhat contradict the recent observations~\citep{downey-etal-2023-embedding,yamaguchi-etal-2024-empirical,yamaguchi2024effectivelyexpandvocabularyllms} that CPT-only models often perform better than vocabulary adapted models, possibly due to the robustly aligned original embeddings.
However, \citet{fujii2024continual} reported that ``\textit{the overall impact of vocabulary expansion on performance is minor}.'' 
Further, \citet{dobler2024language} also claimed that ``\textit{we do not see a clear trend of better
performance with or without tokenizer swapping}'' for their vocabulary adaptation experiments.
We hypothesize that the superiority of CPT is greatly affected by the amount of CPT data, and it can be more apparent in low-resource settings as in \citet{yamaguchi2024effectivelyexpandvocabularyllms}, where new embeddings are likely to be not well aligned.

It is important to note that the CPT-only models (i.e., Base+CPT and Chat+CPT) have no speedups at all (Tables \ref{tab:speed_performance_qwen25} and \ref{tab:speed_performance_llama31}) as they use the same vocabulary as the source models (i.e., Base and Chat).

\begin{figure*}[t]
\centering
\includegraphics[width=\textwidth]{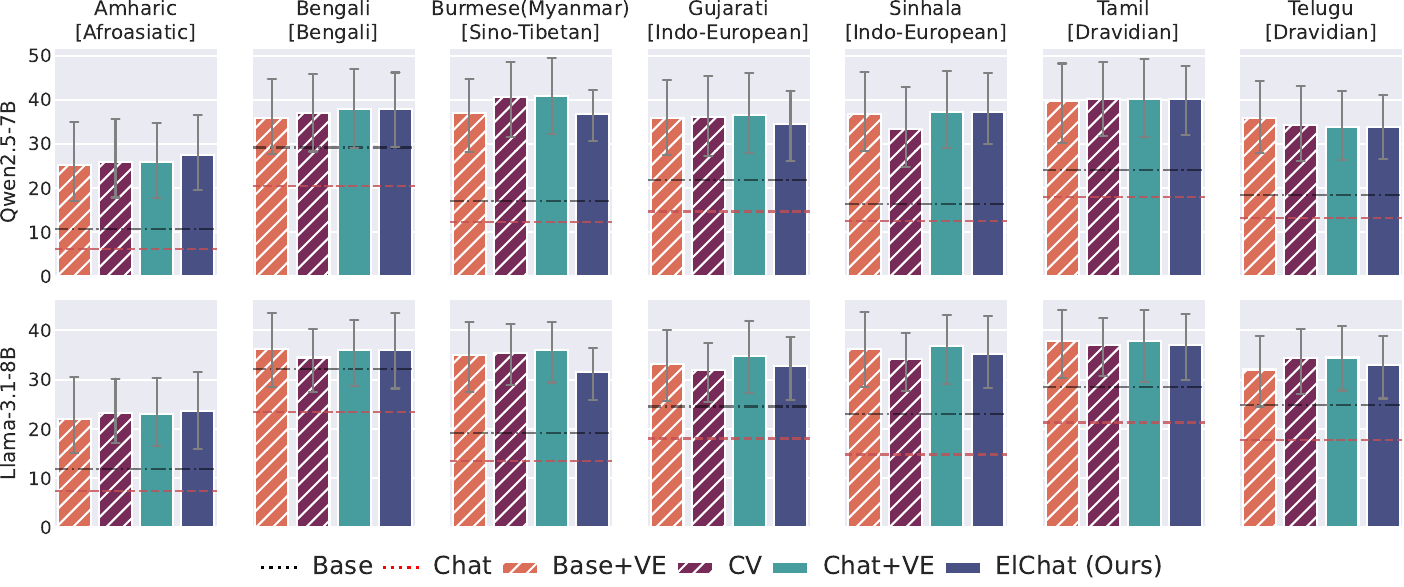}
\caption{Aggregated mean performance across target language tasks for each model by language (error bars indicate 95\% confidence interval).
}
\label{fig:performance_language}
\end{figure*}

\begin{figure*}[t]
\centering
\includegraphics[width=\textwidth]{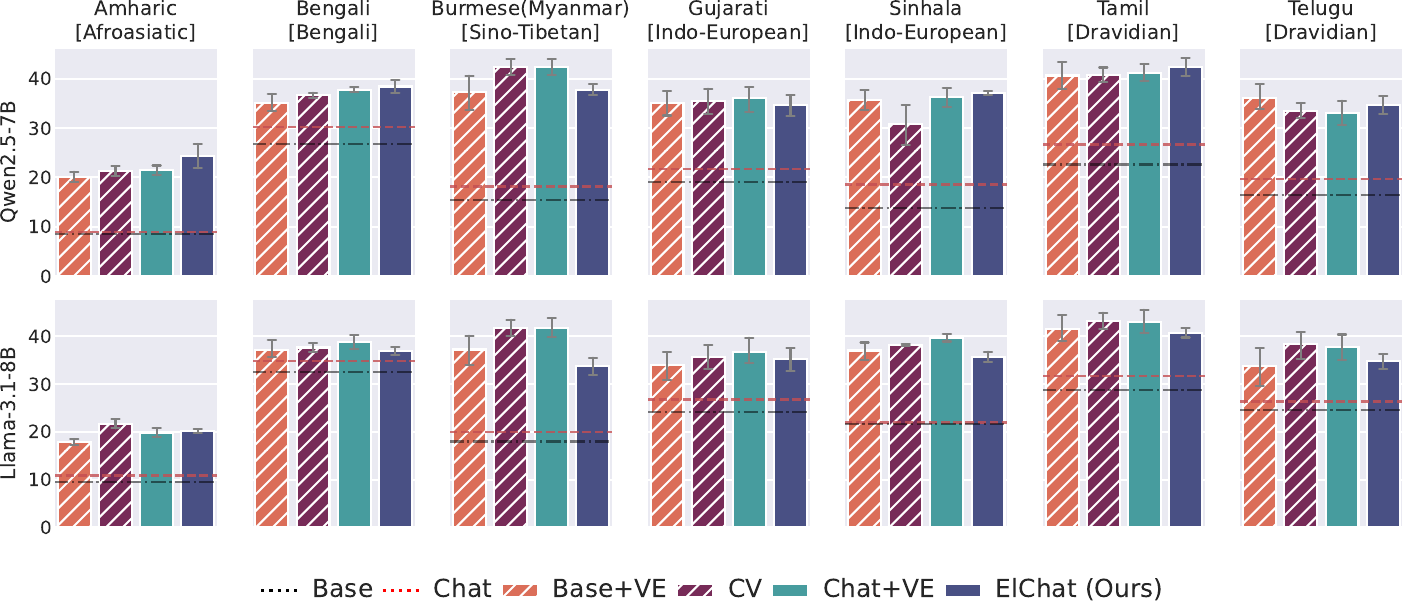}
\caption{Aggregated mean performance across target language generative tasks (\textsc{sum} and \textsc{mt}) for each model by language (error bars indicate 95\% confidence interval).
}
\label{fig:performance_language_generative}
\end{figure*}

\subsection{Additional Analysis by Language} \label{appendix:language}

We conduct additional analysis of the target language task performance of models by language.
Figure \ref{fig:performance_language} presents the aggregated mean performance across three target language tasks (i.e., Target \textsc{sum}, English-to-target \textsc{mt}, and Target \textsc{mc}).
Figure \ref{fig:performance_language_generative} shows the aggregated mean performance across \textit{generative} target language tasks (i.e., Target \textsc{sum} and English-to-target \textsc{mt}).

\paragraph{Performance improvements achieved with VE are evident across languages and models.}

Overall, we observe from Figure \ref{fig:performance_language} that ElChat consistently outperforms Base and Chat in all the target languages across models.

\paragraph{The extent to which adapted chat models improve target language task performance substantially varies by language and model.}
We observe from Figures \ref{fig:performance_language} and \ref{fig:performance_language_generative} that ElChat substantially improves their target language performance over Base and Chat across models in Amharic, Burmese (Myanmar), and Sinhala. 
However, its improvement in Bengali remains minimal for both Qwen2.5 and Llama 3.1, especially in generative tasks (Figure \ref{fig:performance_language_generative}).
Similarly, for Llama 3.1, the performance gains in Tamil, Telugu, and Gujarati are less pronounced, particularly in generative tasks (Figure \ref{fig:performance_language_generative}).
\textit{We hypothesize that this disparity strongly correlates with the amount of language-specific data used to train the source chat model.}
Note that similar findings have already been reported by \citet{yamaguchi-etal-2024-empirical} and \citet{tejaswi-etal-2024-exploring}, \textit{inter alia.}

\paragraph{How many languages do source chat LLMs support?}

The exact language coverage of the source LLMs remains unclear, as none explicitly list their supported languages.

Qwen2.5~\citep{qwen2.5,yang2024qwen2technicalreport} reportedly supports over 29 languages, including Chinese, English, French, Spanish, Portuguese, German, Italian, Russian, Japanese, Korean, Vietnamese, Thai, and Arabic. However, none of our target languages are explicitly included in this list.

Llama 3.1~\citep{dubey2024llama3herdmodels} officially supports English, German, French, Italian, Portuguese, Hindi, Spanish, and Thai. However, it employs a FastText-based language identification model to categorize documents into 176 languages during pre-processing, suggesting that some of our target languages may be included in its pre-training data.

Additionally, Llama 3.1 reportedly utilizes 15T multilingual tokens~\citep{dubey2024llama3herdmodels} for pre-training, while Qwen2.5 uses 18T tokens \textit{in total} (not exclusively multilingual)~\citep{qwen2.5,yang2024qwen2technicalreport}.
This suggests that Llama 3.1 likely benefits from exposure to a broader set of languages. The relatively modest performance improvements observed for Gujarati, Tamil, and Telugu in Llama 3.1 could indicate that these languages were already present in its training data.

\paragraph{Approximating the language coverage.}

We can roughly estimate the language coverage of the source LLMs using MADLAD-400.
The rationale is that these open-weight models are generally trained mainly on a mixture of publicly available data~\citep{gemmateam2024gemma2improvingopen,dubey2024llama3herdmodels}.
Given that MADLAD-400 is sourced from CommonCrawl as mentioned in \S\ref{sec:setup}, its data distribution can potentially approximate the relative coverage of our target languages.

Table \ref{tab:madlad} presents the data statistics of MADLAD-400 for our target languages.
We observe that Burmese, Amharic, and Sinhala have the fewest articles and total characters in MADLAD-400.
Notably, these languages also exhibit the largest performance gains in our experiments. 
This suggests a strong correlation between the size of language-specific data in MADLAD-400 and the effectiveness of VE in task performance.
On the basis of these results, we hypothesize that the two source LLMs used in our experiments might have been trained on very limited language-specific data, or possibly not at all, for Burmese, Amharic, and Sinhala.
In contrast, Tamil, Telugu, Bengali, and Gujarati each have over 1 million articles, making them at least 7.4 times larger than the Burmese dataset.
This further suggests a higher likelihood of their inclusion in the pre-training data of the source LLMs.

\begin{table}[H]
\begin{center}
\small
\caption{Data size of MADLAD-400~\cite{kudugunta2023madlad} for each language.}
\begin{tabular}{lll}
\toprule
\textbf{Language} & \textbf{Number of articles} & \textbf{Number of characters} \\ 
\midrule
  Tamil       &   5.6M                  & 10.6B                     \\
  Telugu      &  2.5M                   & 5.1B                     \\
  Bengali     & 4.3M                    & 4.3B                     \\
  Gujarati    & 1.3M                    & 2.1B                     \\
  Sinhala       &  788K                   &  1.9B                    \\
  Amharic       &  245.2K                   & 509M                     \\
  Burmese       &   176.5K                  &  1.3B                    \\ \bottomrule
\end{tabular}
\label{tab:madlad}
\end{center}
\end{table}

\subsection{Additional Analysis with Newer and Larger Model} \label{appendix:qwen3-result}

We examine the efficacy of ElChat against a newer and larger model compared to Qwen2.5 7B and Llama 3.1 8B. To this end, we employ a state-of-the-art Qwen3 14B model, applying ElChat for adaptation. 
Due to computational limitations, our evaluation focuses on Amharic, Bengali, and Telugu. Amharic is selected because it shows the most significant speedup gains in target language generative tasks with Qwen2.5 and Llama 3.1 (Tables \ref{tab:speed_performance_qwen25} and \ref{tab:speed_performance_llama31}). Bengali and Telugu are chosen as they are the only languages covered by \textsc{mgsm}.
Our analysis will focus on two key questions: (i) \textit{Does Qwen3 exhibit a similar trend in performance with ElChat?} (ii) \textit{If not, what are the potential reasons for the divergence?}

\paragraph{Safety.}
Table \ref{tab:chat_performance_qwen3} shows that the results of Qwen3 align with those of Qwen2.5 in Figure \ref{fig:chat}. 
Specifically, CV outperforms ElChat on \textsc{ToxiGen} and \textsc{ImplicitHate} by up to 5 points, while ElChat performs slightly better on \textsc{TruthfulQA} (except for Telugu). Notably, both ElChat and CV generally surpass the Chat baseline across tasks, indicating that they improve safety performance even in this newer, larger model.

\paragraph{Chat and Instruction-following.}
Table \ref{tab:chat_performance_qwen3} demonstrates that ElChat significantly better recovers chat and instruction-following abilities compared to CV across all tasks and languages. For example, ElChat shows drops of up to 6, 6, and 0.41 points from the Chat baseline in \textsc{IFEval}, \textsc{GSM8K}, and \textsc{MT-Bench}, respectively. In contrast, CV experiences considerably larger drops of up to 41, 56, and 0.66 points for the same tasks. A similar trend is observed in \textsc{AlpacaEval} (Table \ref{tab:alpacaeval}). On \textsc{mgsm} (Table \ref{tab:mgsm_aux}), both ElChat and CV outperform the Chat baseline; they are competitive in Bengali with a 1-point difference, though CV outperforms ElChat in Telugu by 6 points. Overall, these trends are consistent with our previous observation in \S\ref{subsec:task_chat} that ElChat is more effective than CV in enhancing both chat and instruction-following capabilities.

\paragraph{Target Language.}
Contrary to our observations with Qwen2.5 and Llama 3.1 (\S\ref{subsec:task_target}; Figure \ref{fig:performance}, left), ElChat does not consistently outperform Chat in Qwen3 for target language tasks, as shown in Table \ref{tab:task_performance_qwen3}. While we see consistent gains for Amharic, Bengali and Telugu show performance drops in \textsc{mt}, \textsc{mc}, and \textsc{gmmlu} with decreases up to 18 points (e.g., Bengali \textsc{gmmlu}).
Similar drops are also observed in Chat+VE and CV, whereas Chat+CPT generally maintains performance comparable to Chat.

We hypothesize that Bengali and Telugu may already be well-represented within Qwen3 due to its significantly enhanced multilingual capabilities.
Indeed, Qwen3 was pre-trained on 36T tokens, covering up to 119 languages and dialects.
This represents a substantial increase from Qwen2.5, specifically, 29 supported languages and 18T tokens, suggesting a much broader and deeper understanding of various languages.
This hypothesis is further supported by Table 36 in \citet{yang2025qwen3technicalreport}, which lists both Bengali and Telugu as officially supported by Qwen3, while Amharic is not.

Finally, when comparing ElChat and CV, we observe they perform competitively across tasks and languages, with a maximum difference of 5.3 points in Amharic \textsc{mt}. This generally aligns with our observations in Qwen2.5 and Llama 3.1.

\paragraph{Source Language (English).}
Consistent with observations in \S\ref{subsec:task_english} for Qwen2.5 and Llama 3.1, we observe from Table \ref{tab:task_performance_qwen3} that ElChat generally improves source language (English) performance compared to Chat+VE across tasks and languages. However, these improvements are marginal or modest, with a maximum gain of 6.3 points (Telugu \textsc{mt}). This smaller improvement in English indicates less performance degradation during the VE-process (i.e., less catastrophic forgetting of general English capabilities), suggesting enhanced robustness in Qwen3.

Comparing ElChat with CV, ElChat consistently demonstrates superior or equivalent performance in source language tasks across all languages. This aligns with our observation in \S\ref{subsec:task_english}. Notably, the gains tend to be more pronounced in the two generative tasks, \textsc{sum} and \textsc{mt}, with improvements of up to 19.3 points (Bengali \textsc{mt}).

\paragraph{Inference Efficiency.}
Finally, Table \ref{tab:speed_performance_qwen3} shows that both ElChat and CV consistently provide speedup gains of up to 3.8x (Amharic \textsc{sum} with ElChat) in generative tasks within Qwen3. While CV exhibits some reductions in speedup ratios compared to Base+VE (e.g., in Amharic and Bengali), ElChat maintains inference speedups similar to Chat+VE across all tasks and languages. These findings are generally consistent with the results presented in \S\ref{sec:efficiency-result} for Qwen2.5 and Llama 3.1.

\section{License}
This study uses various publicly available models and datasets with different licenses, as detailed below, all of which permit their use for academic research.

\subsection{Models}
Qwen2.5 and Qwen3 are distributed under Apache License 2.0.
Llama 3.1 is licensed under the Llama 3 Community License Agreement.\footnote{\url{https://llama.meta.com/llama3/license/}}

\subsection{Datasets}
XL-Sum is licensed under CC BY-NC-SA 4.0.
Belebele and FLORES-200 are licensed under CC BY-SA 4.0.
BBH, MMLU, GSM8K, and ImplicitHate are distributed under the MIT License.
MGSM is distributed under CC BY 4.0.
ToxiGen is licensed under Community Data License Agreement - Permissive - Version 2.0.
AlpacaEval, IFEval, MT-Bench, and TruthfulQA are distributed under Apache License 2.0.

\end{document}